\documentclass[10pt,twocolumn,letterpaper]{article}

\usepackage[pagenumbers]{cvpr}

\usepackage{xspace}
\usepackage{soul}
\usepackage{booktabs}
\usepackage{graphicx}
\usepackage{pifont}
\usepackage{relsize}
\usepackage{multirow}

\definecolor{badgray}{HTML}{999999}
\definecolor{pastelblue}{HTML}{D1E8FF}
\definecolor{pastelgreen}{HTML}{D1FFD7}

\usepackage{tikz}
\usepackage{tcolorbox}

\usepackage{tikz}
\usetikzlibrary{arrows,shapes,positioning,shadows,trees,mindmap}
\usepackage[edges]{forest}
\usetikzlibrary{arrows.meta}
\colorlet{linecol}{black!75}
\usepackage{xkcdcolors} %

\usepackage[normalem]{ulem}

\usepackage{tikz}
\usetikzlibrary{backgrounds}
\usetikzlibrary{arrows,shapes}
\usetikzlibrary{tikzmark}
\usetikzlibrary{calc}
\newcommand{\highlight}[2]{\colorbox{#1!17}{$\displaystyle #2$}}

\colorlet{mhpurple}{Plum!80}

\renewcommand{\highlight}[2]{\colorbox{#1!17}{#2}}

\usetikzlibrary{
    backgrounds,
    arrows,
    shapes,
    calc,
    tikzmark,      %
    backgrounds,   %
    fit,           %
    positioning,   %
}

\usepackage{microtype}

\newcommand{\cmark}{\textcolor{ForestGreen}{\ding{51}}}
\newcommand{\xmark}{\textcolor{red}{\ding{55}}}
\newcommand{\supmat}{Supp. Mat.\xspace}

\newcommand{\mytilde}{\raise.17ex\hbox{$\scriptstyle\sim$}}

\usepackage{listings}
\usepackage[misc]{ifsym}
\usepackage{tcolorbox}
\tcbuselibrary{most}
\tcbset{
  aibox/.style={
    width=\textwidth,
    top=10pt,
    colback=white,
    colframe=black,
    colbacktitle=black,
    enhanced,
    center,
    attach boxed title to top left={yshift=-0.1in,xshift=0.15in},
    boxed title style={boxrule=0pt,colframe=white,},
  }
}
\newtcolorbox{AIbox}[2][]{aibox,title=#2,#1}

\newlength{\DepthReference}
\newlength{\HeightReference}
\newlength{\Width}%

\newcommand{\MyColorBox}[3][white]%
{%
    \settowidth{\Width}{#3}%
    \fcolorbox{#2}{#1}%
    {%
        \raisebox{-\DepthReference}%
        {%
                \parbox[b][\HeightReference+\DepthReference][c]{\Width}{\centering\textcolor{#2}{#3}}%
        }%
    }%
}

\usepackage{titletoc}
\usepackage{multicol}
\usepackage{censor}

\newcommand{\roneName}{p4zA}
\newcommand{\rtwoName}{Fi2U}
\newcommand{\rthreeName}{YSxJ}

\newcommand{\rone}[1][]{%
\ifx&#1&%
{\textbf{\color{Cerulean}{\roneName}}}%
\else%
{\textbf{\color{Cerulean}{{\smaller{R1}}#1}}}%
\fi%
}

\newcommand{\rtwo}[1][]{%
\ifx&#1&%
{\textbf{\color{ForestGreen}{\rtwoName}}}%
\else%
{\textbf{\color{ForestGreen}{{\smaller{R2}}#1}}}%
\fi%
}

\newcommand{\rthree}[1][]{%
\ifx&#1&%
{\textbf{\color{YellowOrange}{\rthreeName}}}%
\else%
{\textbf{\color{YellowOrange}{{\smaller{R3}}#1}}}%
\fi%
}

\newcommand{\issue}[2][]{%
\ifx&#1&%
\noindent\textbf{#2.}%
\else%
\noindent\textbf{[#1]}~\textbf{#2.}%
\fi%
}

\newcommand{\issuenocolon}[2][]{%
\ifx&#1&%
\noindent\textbf{#2}%
\else%
\noindent\textbf{[#1]}~\textbf{#2}%
\fi%
}

\newcommand{\method}{SITH\xspace}
\newcommand{\methodacro}{\textbf{S}emantic \textbf{I}nspection of \textbf{T}ransformer \textbf{H}eads\xspace}

\newcommand{\comp}{COMP\xspace}
\newcommand{\compacro}{\textbf{C}oherent \textbf{O}rthogonal \textbf{M}atching \textbf{P}ursuit\xspace}

\newcommand{\plusours}{ \ w/ \textbf{\method (Ours)}}

\newcommand{\image}[0]{{I}}
\newcommand{\imageencoder}[0]{\mathcal{E}_{I}}
\newcommand{\textencoder}[0]{\mathcal{E}_{T}}
\newcommand{\conceptpool}[0]{{\Gamma}}
\newcommand{\concept}[0]{{\gamma}}

\newcommand{\matr}[1]{\bm{#1}}
\newcommand{\vect}[1]{\bm{#1}}

\usepackage{scalerel,graphicx,xparse}

\NewDocumentCommand\emojisnow{}{%
\raisebox{-0.2em}{\includegraphics[height=1.2em]{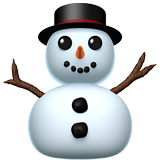}}%
}

\newcommand{\icon}[2][]{\texorpdfstring{\protect{\raisebox{-0.2em}{\includegraphics[width=1.2em]{#2}}}}{#1}}

\definecolor{cvprblue}{rgb}{0.21,0.49,0.74}
\usepackage[pagebackref,breaklinks,colorlinks,allcolors=cvprblue]{hyperref}
\usepackage{algorithm}
\usepackage{algpseudocode}
\usepackage{makecell}
\usepackage{bm}
\usepackage{textcomp}

\def\paperID{*****}
\def\confName{CVPR}
\def\confYear{2026}

\title{From Weights to Concepts: Data-Free Interpretability of CLIP \texorpdfstring{\protect \\}{} via Singular Vector Decomposition}

\author{
Francesco Gentile\textsuperscript{1}$^*$ \quad
Nicola Dall'Asen\textsuperscript{1,2} \quad
Francesco Tonini\textsuperscript{1,3} \quad
Massimiliano Mancini\textsuperscript{1}  \\[-0.05em]
Lorenzo Vaquero\textsuperscript{3} \quad
Elisa Ricci\textsuperscript{1,3} \\[0.4em]
\textsuperscript{1}University of Trento \quad \textsuperscript{2}University of Pisa \quad
\textsuperscript{3}Fondazione Bruno Kessler \\[-0.05em]
{\tt\small \url{https://frangente.github.io/SITH}}
{}\texorpdfstring{\protect\vspace{-0.9em}}{}
}

\begin{document}
\maketitle
\begin{abstract}
As vision-language models 
are deployed at scale, understanding their internal mechanisms becomes increasingly critical.
Existing interpretability methods predominantly rely on activations, making them dataset-dependent, vulnerable to data bias, and often restricted to coarse head-level explanations.
We introduce \method (\methodacro), a fully data-free, training-free framework that directly analyzes CLIP's vision transformer in \textbf{weight space}.
For each attention head, we decompose its value-output matrix into singular vectors and interpret each one via \comp (\compacro), a new algorithm that explains them as sparse, semantically coherent combinations of human-interpretable concepts.
We show that \method yields coherent, faithful intra-head explanations, validated through reconstruction fidelity and interpretability experiments.
This allows us to use \method for precise, interpretable weight-space model edits that amplify or suppress specific concepts, improving downstream performance without retraining.
Furthermore, we use \method to study model adaptation, showing how fine-tuning primarily reweights a stable semantic basis rather than learning entirely new features.
\end{abstract}
    
\section{Introduction}
\label{sec:intro}

Vision-Language Models (VLMs), such as CLIP~\cite{radford2021learning}, have demonstrated impressive capabilities in multimodal understanding and have become central to numerous practical applications~\cite{xu2022simple,wu2023cora,jia2022visual}. 
The standard approach to VLMs has been to treat them as black boxes, focusing on their outputs rather than the underlying computational processes that produce them. However, as these models grow in scale and see broader real-world deployment, understanding how they represent and integrate concepts has become increasingly important. Mechanistic interpretability \cite{kastner2024explaining} seeks to shed some light on understanding neural networks by mapping internal mechanisms to human-meaningful explanations, allowing us to identify which sub-components are responsible for the outputs of downstream tasks.

\begin{figure}[!t]
    \centering
    \includegraphics[width=\linewidth]{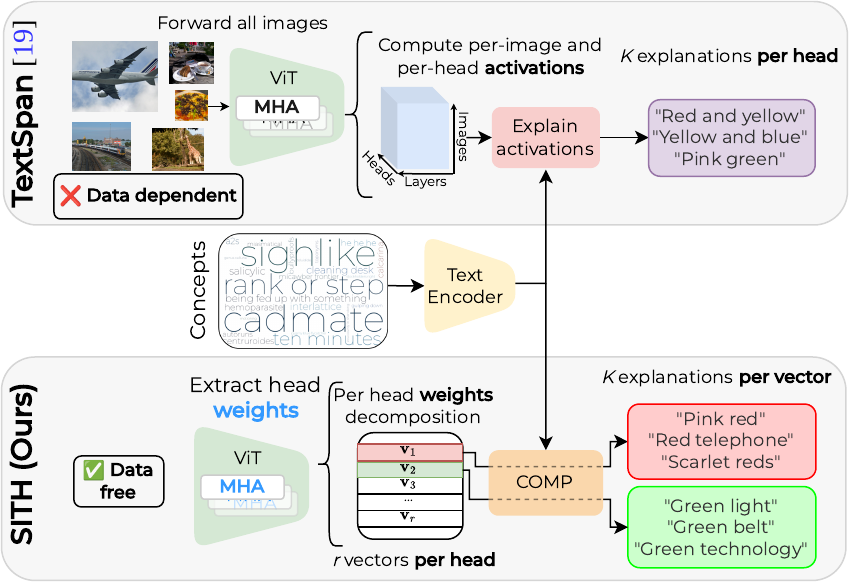}
    \caption{
    \textbf{Comparison between TextSpan~\cite{gandelsman2023interpreting} and our proposed \method applied to CLIP's vision transformer.}
    TextSpan computes attention-head \emph{activations} on a large image dataset (\eg, ImageNet~\cite{deng2009imagenet}) and aligns them with textual concepts, yielding \emph{head-level} interpretations (\eg, head $h_1$ focuses on colors).
    In contrast, \method is \emph{data-free}: it directly decomposes each head's \emph{weights} into singular vectors and interprets them via \comp, providing fine-grained, semantically coherent explanations at the \emph{vector level} (\eg, vector $\vect{v}_2$ within $h_1$ specializes in color green).
    }
    \label{fig:teaser}
\end{figure}

Despite significant advances in model interpretability, mechanistic understanding of VLMs
remains an open problem. %
Previous works proposed to investigate the output representation of CLIP vision transformer %
layers~\cite{rao2024discover,pach2025sparse,bhalla2024interpreting} or of its attention heads~\cite{gandelsman2023interpreting} by associating intermediate model \textit{activations} with human-interpretable concepts, as those features lie in the shared vision-language space.
While such approaches have provided valuable insights into CLIP, they either explain how single samples are represented, or extract only coarse-grained descriptions of entire attention heads.
Moreover, as these methods depend on large-scale datasets to derive activations, their interpretation of model components is inherently influenced by dataset biases (\cref{fig:teaser}, top).

\noindent Motivated by these challenges, this work introduces \method (\methodacro), a novel approach to mechanistic interpretability that does not depend on model activations and
directly analyzes the
weights of CLIP's attention heads (\cref{fig:teaser}, bottom).

\method builds on the seminal work of~\citet{elhage2021mathematical}, which shows that the computation performed by an attention head can be expressed as a weighted combination of its input patches transformed by the head's value-output (VO) weight matrix. Leveraging this insight, \method decomposes the attention head's VO matrix using Singular Value Decomposition (SVD)~\cite{eckart1936approximation}, revealing its dominant computational directions.
Then, to interpret these directions semantically, we propose \compacro (\comp), a novel decomposition algorithm representing each singular vector as a sparse, positive combination of textual concepts, optimizing both reconstruction fidelity and semantic coherence.

\method's weight-based perspective offers several key advantages.
As shown in~\cref{tab:works_comparison}, our approach is entirely training-free, requiring no additional optimization to obtain concept interpretations.
Second, \method is data-free, as it does not depend on any dataset or model activations to derive the interpretations, thus avoiding dataset-induced biases. 
Third, by analyzing singular vectors rather than entire heads, it enables a more fine-grained dissection of the model's knowledge,
allowing us to isolate and intervene on individual semantic subspaces encoded within a head.

\begin{table}[!t]
\centering
\caption{\textbf{Comparison between prior works and our \method}. %
}
\resizebox{0.95\columnwidth}{!}{%
\begin{tabular}{@{}l@{\hskip2pt}c@{\hskip5pt}c@{\hskip5pt}c@{}}
\toprule
            & \textbf{Training-free} & \textbf{Data-free} & \textbf{Explanation Granularity} \\ \midrule
SAE~\cite{rao2024discover,pach2025sparse,zaigrajew2025interpreting} & \xmark & \xmark & Sample-level \\
TextSpan~\cite{gandelsman2023interpreting} & \cmark & \xmark & Head-level \\
\textbf{\method (Ours)}     & \cmark        & \cmark  & Intra-head (singular vector)  \\ \bottomrule
\end{tabular}%
}
\label{tab:works_comparison}
\end{table}

By applying \method to the visual encoder of CLIP, %
we find that individual singular vectors map to distinct, human-interpretable concepts, such as textures, locations, backgrounds, and colors (\cref{sec:evaluation}). 
These findings enable precise, concept-level interventions: by amplifying or suppressing specific singular vectors, we can reduce the model's sensitivity to spurious correlations (\cref{sec:editing:spurious}), suppress undesired concepts (\cref{sec:editing:safe}), and improve performance on downstream tasks (\cref{sec:editing:classification}), all without the need for training or data.
Moreover, \method provides a compelling lens into model adaptation (\cref{sec:finetuning}).
It exposes how fine-tuning subtly reorients the model's feature basis toward the semantic space of the target task, with changes that align with the fine-tuning objective. 
In summary, our contributions are:
\begin{itemize}
    \item We propose \method, a data-free, training-free, and weight-based interpretability framework that provides input-independent explanations of CLIP's attention heads.
    \item \method achieves fine-grained explanations by decomposing model weights into singular vectors and associating them with human-interpretable concepts via \comp, our novel sparse decomposition technique.
    \item We demonstrate how \method can benefit downstream applications via data-free interventions that suppress spurious correlations, reduce sensitivity to unsafe content, and improve classification performance.
    \item We use \method to analyze model adaptation, showing that fine-tuning introduces meaningful, task-specific directional changes without altering the overall structure of the learned feature basis.
\end{itemize}

\section{Related works}
\label{sec:related}

\noindent \textbf{Mechanistic Interpretability} aims to uncover how neural networks operate internally by examining their core components and the computational pathways they form. 
This can be done either by studying model responses to input features (\textit{activation-based} interpretability) or by inspecting its weights directly (\textit{weight-based} interpretability).

\noindent \textbf{Activation-based interpretability}
studies initially focused on individual neurons~\cite{olah2018building,cammarata2020curve,goh2021multimodal,olah2017feature,nostalgebraist2020logitlens}.
However, interpreting neuron representations is difficult, especially in large models, as neurons often exist in a state of superposition~\cite{elhage2022toy}, simultaneously encoding multiple unrelated concepts.
Sparse Autoencoders (SAEs)
project hidden representations into a sparse set of features that ideally align with human-interpretable concepts~\cite{bricken2023monosemanticity,rajamanoharan2024jumping,rajamanoharan2024improving,gao2025scaling,bussmann2024batchtopk,bussmann2025learning,costa2025flat}, and have been used to analyze both language~\cite{yun2021transformer,cunningham2023sparse} and 
multimodal
models~\cite{rao2024discover,stevens2025sparse,surkov2024one,kim2024interpreting,lim2024sparse,zaigrajew2025interpreting,pach2025sparse}.
However, SAEs exhibit severe instability, as models trained on similar datasets can produce entirely different dictionaries~\cite{fel2025archetypal}. 
Moreover, the encoding of hierarchical concepts within SAEs is often fragile: seemingly monosemantic features fail to activate when expected, with their activations instead being absorbed by child features~\cite{chanin2024absorption}.
\citet{gandelsman2023interpreting} proposed interpreting the attention heads of CLIP's vision transformer by aligning their output activations across images with a set of textual concepts.
This approach revealed that several attention heads specialize in specific semantic roles, such as colors, numbers, and geographic locations. 
However, like all activation-based methods, it requires large-scale datasets (\eg ImageNet~\cite{deng2009imagenet}) to collect activations from all model components for analysis (see~\cref{tab:works_comparison}).
Moreover the resulting explanations are coarse-grained, as they identify which concepts are represented by a head but do not specify which internal features encode those concepts. 

\begin{figure*}[!t]
\centering
\includegraphics[width=1.0\linewidth]{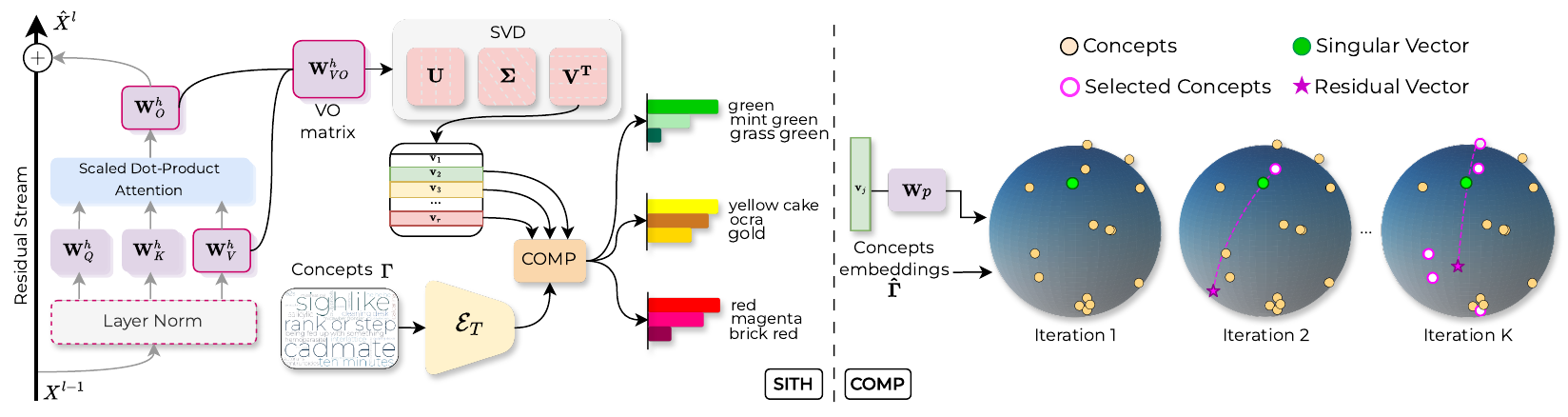}
\caption{\textbf{Overview of \method and \comp}.
\textbf{(left)} \method isolates the \MyColorBox[Plum!17]{RedViolet}{Value-Output} (VO) matrix of a CLIP ViT attention head and factorizes it via Singular Value Decomposition (SVD), yielding singular vectors $\{\mathbf v_j\}$ that capture the head's dominant writing directions.
Given a textual concept pool $\conceptpool$ processed by CLIP's text encoder $\textencoder$, each vector is then explained using \comp, yielding sparse and coherent per-vector interpretations (e.g., ``\textit{green}'', ``\textit{mint green}'', and ``\textit{grass green}'').
\textbf{(right)} \comp iteratively decomposes a \textcolor{ForestGreen}{target singular vector} $\vect{\hat{v}}_j$ as a sparse, non-negative combination of \textcolor{YellowOrange}{concept embeddings $\matr{\hat{\conceptpool}}$}.
In Iteration 1 it selects the \MyColorBox[white]{Rhodamine}{concept} with the highest similarity to the \textcolor{ForestGreen}{target}.
The \textcolor{RoyalPurple}{residual vector} \raisebox{-0.2mm}{\includegraphics[height=3.0mm]{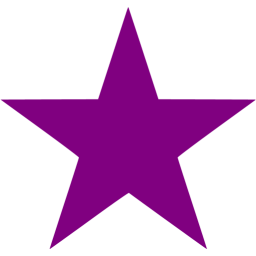}} captures the remaining, unexplained part of the original \textcolor{ForestGreen}{vector}.
From Iteration 2 to $K$, \comp repeatedly chooses additional concepts that are both close to the current \textcolor{RoyalPurple}{residual} and \emph{semantically consistent} with the accumulated \MyColorBox[white]{Rhodamine}{concept set} (see \cref{eq:epsilon}).
This process continues until the sparsity budget $K$ is reached.
}
\label{fig:method}
\end{figure*}

\noindent \textbf{Weight-based interpretability} addresses these limitations by directly analyzing the model's weights to reliably uncover learned features~\cite{voss2021visualizing}.
While applying parameter decomposition or spectral analysis to Transformer weights has prior precedent, existing works do so under different assumptions and goals. For instance, \citet{braun2025interpretability} and \citet{bushnaq2025stochastic} learn data-dependent factorizations, whereas existing SVD-based analyses focus on understanding inter-layer communication~\cite{merullo2024talking}, the geometric properties of weights~\cite{song2023uncovering}, or the local/global behavior of attention~\cite{pan2024dissecting}. 
Furthermore, works analyzing the features encoded by weights~\cite{geva2021transformer,dar2022analyzing} or singular vectors~\cite{millidge2022singular} via token-space projections remain limited to unimodal language models and rely on simple nearest-neighbor searches, which fail to capture the full semantic content, leading to incomplete interpretations.
In contrast, we propose a coherence-regularized sparse coding algorithm to map singular vectors into complete and human-interpretable explanations. We then use these decompositions within an end-to-end, data-free interpretability framework for VLMs, enabling fine-grained intra-head concept attribution and direct singular-value editing at scale.

\section{\method: Interpreting CLIP Weights}
\label{sec:method}
\method 
associates features from the heads of the CLIP Vision Transformer with human-interpretable concepts drawn from an overcomplete pool $\conceptpool = \{\concept_1, \dots, \concept_C\}$.
Unlike prior works~\cite{gandelsman2023interpreting}, we extract such features from the model \textit{weights} directly, without requiring additional data.
We begin by reviewing the CLIP architecture, the residual stream formulation, and how to isolate the contribution of each attention head (\cref{sec:method:preliminaries}).
Then, we apply 
SVD~\cite{eckart1936approximation} to the attention head's value-output matrix, revealing the dominant directions of information flow within each head (\cref{sec:method:svd}).
Finally, we propose \comp, a sparse concept attribution method that maps these directions to interpretable natural language concepts $\concept$, enabling a compact and human-aligned interpretation of the attention heads (\cref{sec:method:comp}). We provide a visualization of our method in \cref{fig:method}.

\subsection{Preliminaries}
\label{sec:method:preliminaries}
\textbf{CLIP} (Contrastive Language-Image Pretraining)~\cite{radford2021learning} maps images and text into a shared embedding space $\subseteq \mathbb{R}^d$ using an image encoder $\imageencoder$ and a text encoder $\textencoder$, trained in a contrastive learning fashion.
Following recent works~\cite{gandelsman2023interpreting,zhang2024vision}, we focus on CLIP models whose vision backbone is based on the Vision Transformer (ViT)~\cite{dosovitskiy2020image} architecture.
Given an image $\image$, $\imageencoder$ first splits it into $P$ non-overlapping patches, which are projected into $D$-dimensional tokens and summed to positional embeddings.
Then, a learnable \texttt{[CLS]} token is prepended to the image patches, resulting in a sequence $\mathbf{X}^0 \in \mathbb{R}^{(P+1) \times D}$ with $\vect{x}_{\texttt{CLS}}^0, \vect{x}_1^0, \vect{x}_2^0, \ldots, \vect{x}_P^0$ as rows, that is processed by $L$ transformer layers. Each layer $l$ computes its output $\matr{X}^{l}$ via two residual updates:
\begin{align}  \label{eq:residual_full}
\matr{\hat{X}}^{l} &= \text{MHA}^l(\text{LN}(\matr{X}^{l-1})) + \matr{X}^{l-1} \\
\matr{X}^{l} &= \text{FFN}^l(\text{LN}(\matr{\hat{X}}^{l})) + \matr{\hat{X}}^{l},
\end{align}
where $\text{MHA}$ and $\text{FFN}$ are the multi-head attention module and feed-forward network components, while $\text{LN}$ is a layer normalization operation. Finally, the CLIP representation $\imageencoder(\image) \in \mathbb{R}^d$ of the image $\image$ is obtained by projecting the final \texttt{[CLS]} token output $\vect{x}^L_{\texttt{CLS}}$ through a learned matrix $\matr{W}_p \in \mathbb{R}^{D \times d}$.

\noindent \textbf{Residual stream decomposition.}
We can express the output of the \texttt{[CLS]} token after $L$ transformer layers as the sum of its initial embedding and the direct contributions from each layer through the residual stream~\cite{gandelsman2023interpreting}:
\begin{equation}
\resizebox{0.9\columnwidth}{!}{%
$\vect{x}^L_{\texttt{CLS}} = \vect{x}^0_{\texttt{CLS}} + \sum_{l=1}^{L} \left( \text{MHA}^l(\matr{X}^{l-1})_{\texttt{CLS}} + \text{FFN}^l(\matr{\hat{X}}^{l})_{\texttt{CLS}} \right)$,%
}
\end{equation}
where the affine parameters of the LN are folded into the adjacent projection matrices~\cite{elhage2021mathematical} (see \supmat, \cref{sec:supp:implementation_details} for details).
This enables a clean separation of contributions from the \texttt{[CLS]} token, the MHA, and the FFN modules.

\noindent\textbf{Decomposing Attention Heads.}
Following~\citet{elhage2021mathematical}, the multi-head attention can be expressed as the sum of the outputs of $H$ independent attention heads:
\begin{equation}
\label{eq:mha}
\text{MHA}(\matr{X}) = \sum_{h=1}^{H} \matr{A}^h \matr{X} \matr{W}_{VO}^h ,
\end{equation}
where $\matr{A}^h$ is the attention matrix for head $h$, and $\matr{W}_{VO}^h \in \mathbb{R}^{D \times D}$ is the value-output (VO) weight matrix for head $h$ obtained by combining its value $\matr{W}_V^h$ and output $\matr{W}_O^h$ projection matrices as $\matr{W}_{VO}^h := \matr{W}_V^h \matr{W}_O^h$ (refer to~\cref{sec:supp:implementation_details} of the \supmat for additional details).

While both the attention pattern $\matr{A}^h$ and VO matrix $\matr{W}_{VO}^h$ drive a head's computation, they play distinct roles. $\matr{A}^h$ acts as a router deciding how information flows between tokens (e.g., from an ``apple'' patch to the \texttt{[CLS]} token), while $\matr{W}_{VO}^h$ determines which information is moved (e.g., reading ``redness'' from the source and writing it to the destination's residual stream). Hence, by isolating $\matr{W}_{VO}^h$, we gain an input-independent understanding of the specific semantic features a head can extract and write.

\subsection{Finding Principal Directions with SVD}
\label{sec:method:svd}

As stated above, analyzing the $\matr{W}_{VO}^h$ (hereafter $\matr{W}_{VO}$ for ease of notation) of an attention head provides insight into how the head transforms information within the residual stream. Because the VO matrix is a linear transformation, its behavior can be characterized in terms of the directions along which it most strongly amplifies or suppresses information. To capture these dominant axes, we propose to factorize the VO matrix using %
Singular Value Decomposition (SVD)~\cite{eckart1936approximation}, yielding:
\begin{equation}
\matr{W}_{VO} = \matr{U} \matr{\Sigma} \matr{V}^T
\end{equation}
where, $\matr{U}$ and $\matr{V}$ are $D \times r$ semi-orthogonal matrices, with $r = \text{rank}(\matr{W}_{VO})$, whose columns are the left singular vectors $\{\vect{u}_i\}$ and right singular vectors $\{\vect{v}_i\}$, respectively. $\matr{\Sigma}$ is a diagonal matrix of non-negative singular values $\sigma_i \ge 0$, sorted in descending order.

Intuitively, each left singular vector $\mathbf{u}_i$ defines an input direction in the residual stream that the head reads from, while the corresponding right singular vector $\mathbf{v}_i$ defines an output direction in the residual stream that the head writes to. The associated singular value $\sigma_i$ quantifies how strongly the head maps the input direction $\mathbf{u}_i$ to the output direction $\mathbf{v}_i$. Thus, by analyzing the singular vectors, we can identify the most significant features that the attention head is capable of extracting and writing back to the residual stream\footnote{Note that the effective contribution of a writing direction $\mathbf{v}_i$ to the residual stream is also influenced by how well input tokens align with the corresponding reading direction $\mathbf{u}_i$, and by the attention scores.}.

\subsection{Semantic Interpretation via COMP}
\label{sec:method:comp}

Since the singular vectors lie in the same semantic space of the image embeddings, we can leverage the embedding space of CLIP to semantically interpret them. More formally, given the right singular matrix $\matr{V}^T$ (and analogously for the left singular matrix $\matr{U}$), let $\matr{\hat{V}}^T = \matr{V}^T \matr{W}_p$ be the singular vectors projected into the CLIP multimodal space\footnote{To mitigate the multimodality gap of CLIP, we follow the approach of~\citet{bhalla2024interpreting}; see \cref{sec:supp:implementation_details} of the \supmat}. We can identify those concepts in the concept pool $\conceptpool$ which better correspond to the semantic content of each singular vector by computing the cosine similarity between $\vect{\hat{v}}$ and the concept embeddings $\matr{\hat{\conceptpool}} = \textencoder{(\conceptpool)} \in \mathbb{R}^{C \times d}$, and choosing the \textit{top-k} most similar concepts.

While this approach provides a basic interpretation, we found it often fails to capture the full semantic content of the singular vectors.
This is particularly true when the concept pool lacks an exact match for the singular vector, leading to incomplete or misleading interpretations.
For instance, a singular vector representing ``a red apple'' may be ambiguously mapped to either ``apple'' or ``red'' if those are the only available concepts, thus missing the combined meaning.

To address this problem, we propose to express each singular vector $\vect{\hat{v}}$ as a sparse, non-negative linear combination of concept embeddings $\matr{\hat{\conceptpool}}$,
by finding a sparse coefficient vector $\vect{c} \in \mathbb{R}^C$ that solves the following $L_0$-minimization problem:
\begin{equation}
\min_{\vect{c}} \|\vect{c}\|_0 ~~ \text{subject to} ~~ \|\vect{\hat{v}} - \matr{\hat{\Gamma}}^T\vect{c}\|_2^2 \le \epsilon ~~ \text{and} ~~ \vect{c} \ge 0
\label{eq:nno}
\end{equation}
with $\vect{c} \ge 0$ ensuring that each singular vector is expressed in terms of its constituent concepts~\cite{bhalla2024interpreting}.

A standard greedy algorithm for this problem is Non-Negative Orthogonal Matching Pursuit (NNOMP)~\cite{pati1993orthogonal}. 
Let $S_k$ denote the indices of the support set of concepts selected up to step $k$, and let $\vect{r}_k$ be the residual vector, \ie the reconstruction error 
of the target vector $\vect{\hat{v}}$ at step $k$ using $\matr{\hat{\Gamma}}_{S_k}$ as the reconstruction basis.
Starting with an empty support set $S_0$ and initial residual $\vect{r}_0 = \vect{\hat{v}}$, NNOMP proceeds iteratively as follows: 
(1) select the concept $\vect{\hat{\concept}}_i$ from the vocabulary with the highest correlation score $\langle \vect{r}_{k-1}, \vect{\hat{\concept}}_i \rangle$, 
(2) add $i$ to the support set $S_k$, 
(3) update the coefficients $\vect{c}$ by solving a non-negative least squares problem over the selected concepts $\matr{\hat{\Gamma}}_{S_k}$, and 
(4) update the residual $\vect{r}_k$ using $\vect{r}_k = \vect{\hat{v}} - \matr{\hat{\Gamma}}^T_{S_k}\vect{c}$.
This process is iteratively repeated until a stopping criterion is met, such as reaching a target sparsity level $K$ or achieving a sufficiently small residual norm.

A key limitation of this selection strategy (step 1) is that it solely optimizes for reconstruction error, which can lead to a semantically incoherent set of concepts, thus making the final explanation harder to interpret. 

To address this, we introduce \compacro (\comp), an algorithm that modifies the scoring function in the concept selection step to explicitly balance reconstruction quality with semantic coherence.
In \comp, given residual $\vect{r}_{k-1}$ and support set $S_{k-1}$, the concept selection criterion at iteration $k$ is replaced by the following scoring function:
\begin{equation}
    \vspace{\baselineskip}
            \label{eq:epsilon}
                \text{score}(\vect{\hat{\concept}}_i) = \tikzmarknode{x}{\highlight{red}{$\langle \vect{r}_{k-1}, \vect{\hat{\concept}}_i \rangle$}} + \tikzmarknode{s}{\highlight{blue}{$\frac{\lambda}{|S_{k-1}|} \sum_{j \in S_{k-1}} \langle \vect{\hat{\concept}}_i, \vect{\hat{\concept}}_j \rangle$}}, 
\end{equation}
\begin{tikzpicture}[overlay,remember picture,>=stealth,nodes={align=left,inner ysep=1pt},<-]
    \path (x.south) ++ (0,-1.5em) node[anchor=south east,color=red] (scalep){\footnotesize Reconstruction Term};
    \draw [color=red!67](x.south) |- ([xshift=-0.3ex,color=green]scalep.south west);
    \path (s.south) ++ (0,-1.5em) node[anchor=south west,color=blue!67] (mean){\footnotesize Coherence Term};
    \draw [color=blue!57](s.south) |- ([xshift=-0.3ex,color=blue]mean.south east);
\end{tikzpicture}\hspace{-3pt}
where \textcolor{red!47}{\textit{Reconstruction Term}} is the standard NNOMP criterion, \textit{i.e.}, selecting concepts that align with the unexplained portion of the target singular vector, and the \textcolor{blue!67}{\textit{Coherence Term}} is a novel regularizer that measures the average semantic similarity
between a candidate concept $\vect{\hat{\concept}}_i$ and all concepts $\vect{\hat{\concept}}_j$ already chosen. 
This term encourages the selection of new concepts that are semantically related to the concepts already in the support set.
The hyperparameter $\lambda \ge 0$ controls this trade-off. When $\lambda=0$, COMP recovers standard NNOMP. As $\lambda$ increases, the algorithm increasingly favors semantic coherence, producing a sparse set of concepts that is not only reconstructive but also forms a more interpretable and meaningful explanation. We provide the pseudocode of COMP in \cref{sec:supp:pseudocode} of the \supmat

\section{Evaluating \method}
\label{sec:evaluation}
For a given attention head $h \in [1, H]$ within transformer layer $l \in [1, L]$, \method analyzes 
it by decomposing its VO matrix $\matr{W}^{l,h}_{VO}$ into $r$ singular vectors, and linking each singular vector to $K$ concepts.
Here we show that such concept sets provide highly coherent and human-interpretable explanations, while remaining faithful to the original semantic information encoded within the vectors.

\begin{figure}[!t]
\centering
\includegraphics[width=0.95\linewidth]{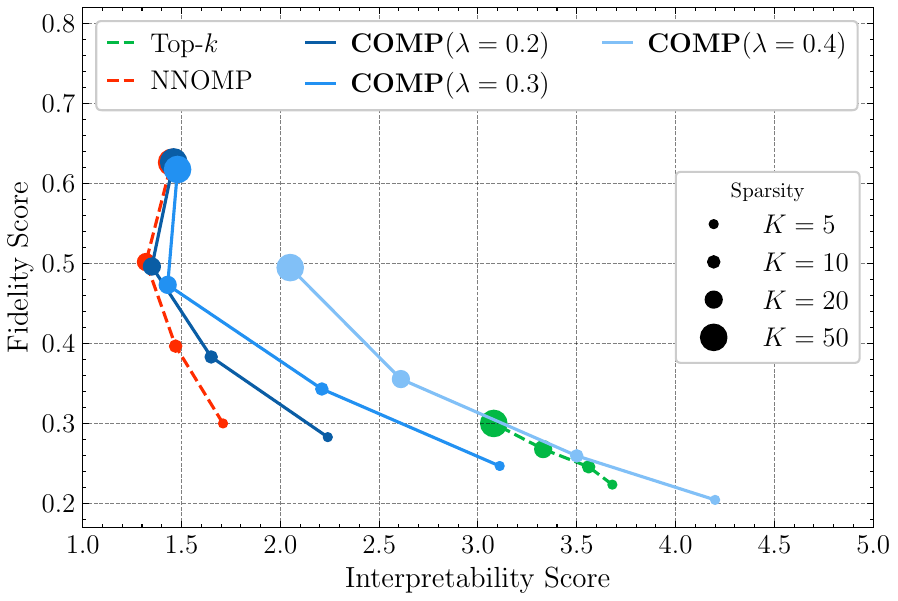}
\label{fig:reconstruction-similarity}
\caption{\textbf{Interpretability and fidelity scores of \method evaluated under varying sparsity levels} using different decomposition strategies: our proposed \comp, NNOMP~\cite{pati1993orthogonal}, and top-k.
}
\label{fig:reconstruction_vs_monosemanticity}
\end{figure}

\noindent \textbf{Experimental setting.}
Following~\citet{gandelsman2023interpreting}, we focus our analysis on the last 4 layers of the OpenCLIP ViT-L/14 model~\cite{ilharco2021openclip} ($L = 24$, $H=16$, $r = 64$) pretrained on the LAION-2B dataset~\cite{schuhmann2022laion}.
We use ConceptNet 5.5~\cite{speer2017conceptnet} as our concept dictionary (see \cref{secc:supp:concept_ablation} of the \supmat for additional dictionaries) and use GPT-5-mini~\cite{openai_gpt5_systemcard} for those experiments requiring an LLM (see \cref{sec:supp:prompts} of the \supmat for the exact prompts).
Our analysis focuses on the right singular vectors $\vect{v}_i \in \matr{V}$, as they correspond to the output directions of each head and thus have a direct effect on the final image representation.
Refer to \cref{sec:supp:vit_l14} of the \supmat for analysis on the left singular vectors $\vect{u}_i \in \matr{U}$.

\begin{table}[!t]
\centering
\caption{Examples of singular vectors from the last four layers of ViT-L/14 along with their reconstructed concepts using \method with \comp ($\lambda = 0.3$, $K=5$). The numbers in parentheses indicate the coefficients assigned to each concept in the reconstruction.}
\resizebox{0.95\linewidth}{!}{%
\begin{tabular}{l l}
\toprule
\textbf{Layer 23, Head 8, SV 0~~\icon{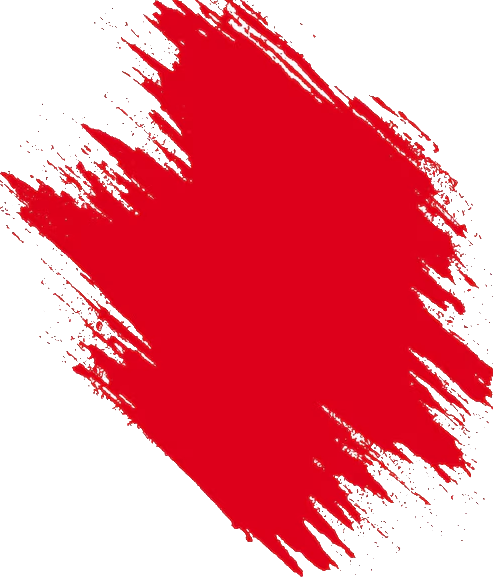}} & \textbf{Layer 22, Head 7, SV 0~~\emojisnow} \\
\midrule
\makecell[l]{pink red (0.3144) \\ red telephone (0.1571) \\ red and white factors (0.1503) \\ scarlet reds (0.1473) \\ red background (0.1369)} & 
\makecell[l]{late december (0.1900) \\ winterwear (0.1833) \\ barren trees in winter (0.1729) \\ winter buds (0.1554) \\ winter in valley (0.1383)} \\
\toprule
\textbf{Layer 21, Head 11, SV 3~~\icon{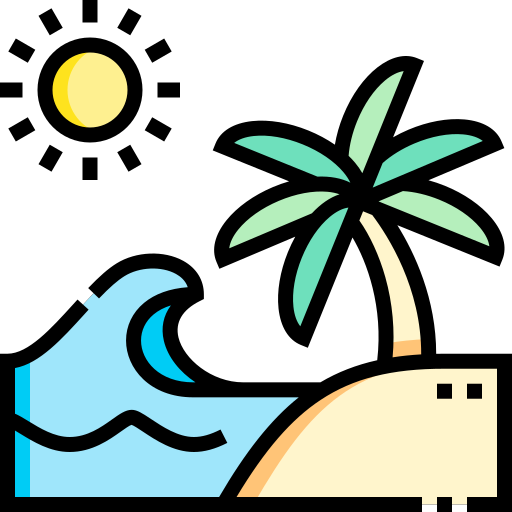}} & \textbf{Layer 20, Head 4, SV 0~~\icon{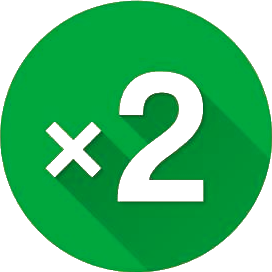}} \\
\midrule
\makecell[l]{ocean beach (0.1334) \\ surf culture (0.1170) \\ sandiego (0.1116) \\ catalina island (0.1069) \\ southern california (0.0213)} &
\makecell[l]{two girls (0.2121) \\ doublepack (0.2065) \\ two combatants (0.1902) \\ comedy duos (0.1666) \\ two credit cards (0.1612)} \\
\midrule
\end{tabular}}
\label{tab:explanations}
\end{table}

\begin{figure*}[!t]
\centering
\includegraphics[width=1.0\linewidth]{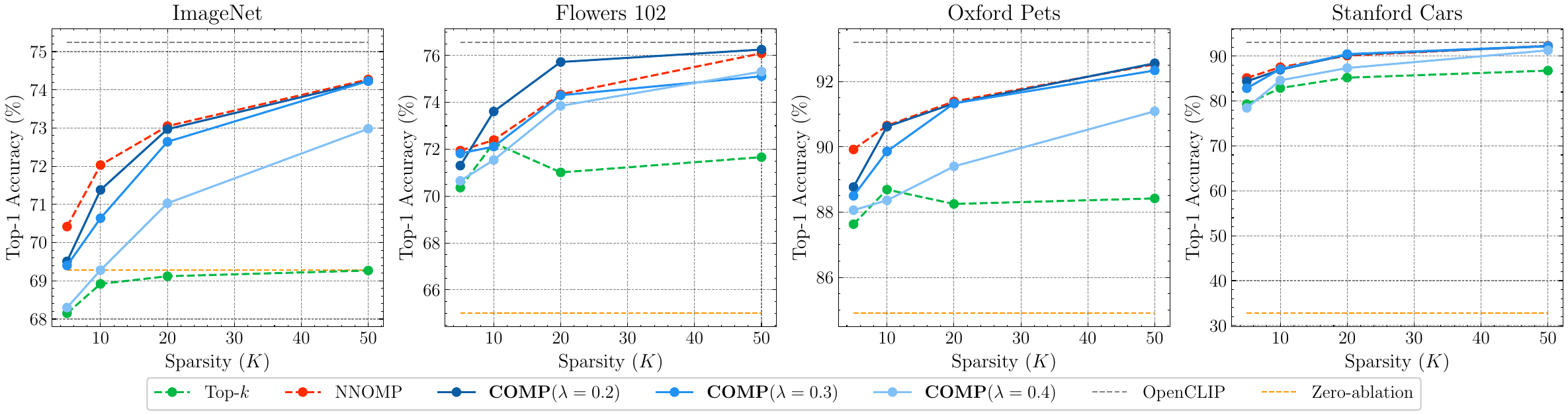}
\caption{\textbf{Zero-shot classification accuracy when replacing the original singular vectors of layer $l = 23$ with their reconstructions} using different numbers of concepts and across different datasets~\cite{deng2009imagenet,nilsback2008automated,parkhi2012cats,krause20133d}. We also report the accuracy of the original OpenCLIP model, and its accuracy when all heads of the layer are zeroed-out (\ie, Zero-ablation).}
\label{fig:reconstruction-accuracy}
\vspace{-1em}
\end{figure*}

\subsection{Interpretability-Fidelity Analysis}
\label{sec:eval_sparsity}

For a decomposition to be useful, the set of concepts associated with a singular vector must satisfy two competing criteria: (i) \emph{fidelity}, the degree to which the concepts faithfully capture the semantic information embedded in the vector; and (ii) \emph{interpretability}, the extent to which the concepts can be understood as a coherent %
explanation.
Fidelity is quantified via cosine similarity between the original vector and its reconstruction using our extracted concepts (see \supmat, \cref{sec:supp:implementation_details} for computation of the reconstructed vector), while
interpretability is assessed using an LLM-as-a-judge rating the coherence of the concept set on a 5-point Likert scale.
A rating of 1 indicates semantically unrelated or incoherent concepts (\eg, ``\textit{vintage toaster}'' and ``\textit{hiking}''), whereas a rating of 5 denotes a highly coherent and 
conceptually aligned 
grouping (\eg, ``\textit{cat}'' and ``\textit{feline fur}'').

In \cref{fig:reconstruction_vs_monosemanticity}, we report results for \method using \textbf{\comp} across varying values of the regularization parameter $\lambda \in \{0.2, 0.3, 0.4\}$, and the number of selected concepts $K \in \{5, 10, 20, 50\}$ on the last layer ($l = 23$; see \supmat, \cref{sec:supp:vit_l14} for results on all other tested layers).
For comparison, we also consider two alternative decomposition strategies, replacing \comp with non-negative orthogonal matching pursuit (\textbf{NNOMP})~\cite{pati1993orthogonal} and a naive \textbf{top-$\bm{k}$} selection of the most similar concepts.
The top-$k$ strategy naturally retrieves highly coherent concept sets, but it struggles with completeness: it yields poor reconstructions that do not improve significantly even when increasing $K$, as it only captures a narrow slice of the singular vector's semantic content. 
In contrast, NNOMP achieves strong reconstruction performance but sacrifices interpretability, often selecting disjointed and less coherent concept sets even at low sparsity levels $K = 5$. 
Our proposed \comp achieves an optimal balance between completeness and coherence, offering reconstructions that closely match original singular vectors while maintaining semantically cohesive, human-interpretable explanations.

Qualitative results of the resulting concept sets along with the weight attributed to each concept can be found in \cref{tab:explanations} (see \cref{sec:supp:qualitative} of the \supmat for further qualitative results).
It is important to note that none of the methods perfectly reconstruct all singular vectors, even at high values of $K=50$.
This is likely due to certain singular vectors encoding not only semantic concepts, but also non-semantic or task-irrelevant ``noisy'' information that cannot be linked to human-interpretable concepts, as suggested in prior work~\cite{bhalla2024interpreting}.
However, this phenomenon does not adversely affect downstream performance.
As shown in \cref{fig:reconstruction-accuracy}, substituting the original singular vectors with their \method-based reconstructions results in negligible degradation in zero-shot classification accuracy across multiple benchmarks~\cite{deng2009imagenet,nilsback2008automated,parkhi2012cats,krause20133d}.

\subsection{Grounding Singular Vectors to Images}
\label{sec:eval_image_matching}

\begin{figure}[!t]
\centering
\includegraphics[width=\linewidth,height=87pt]{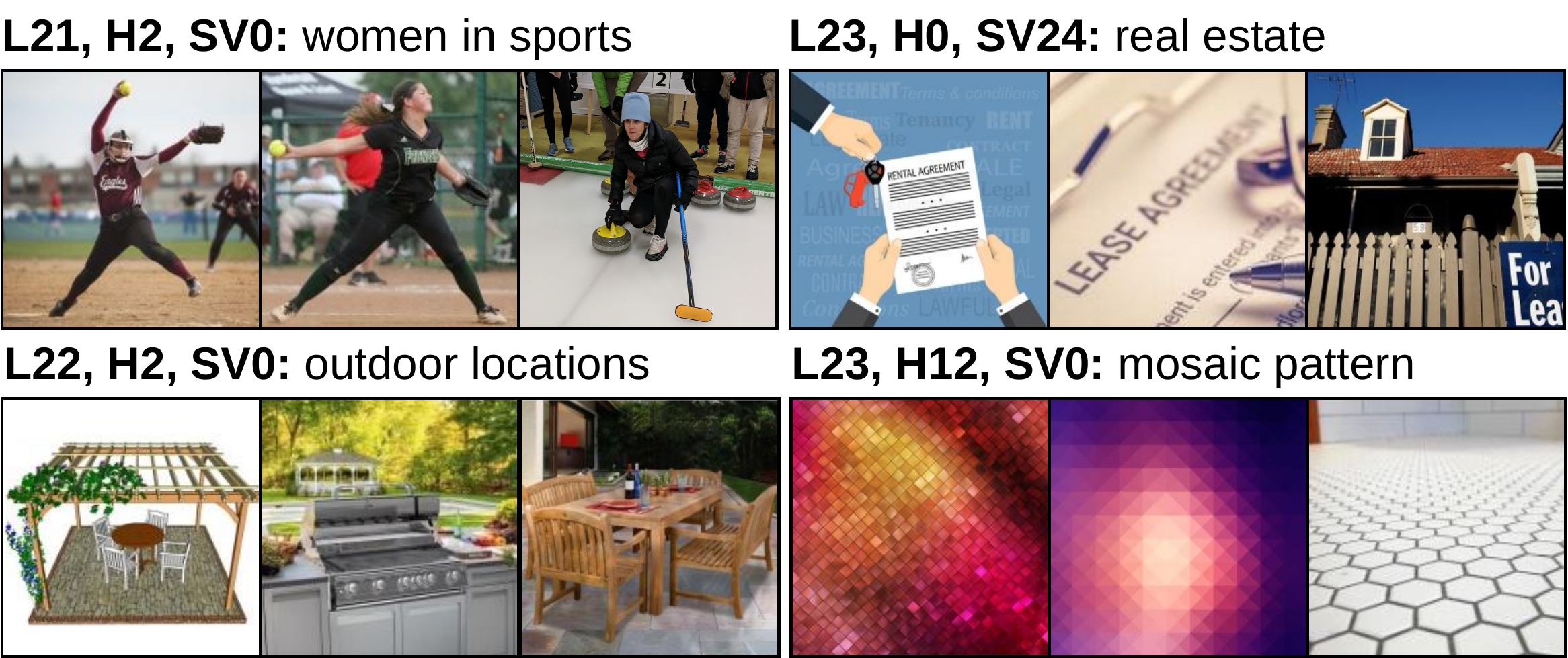}
\caption{\textbf{Top-3 images matched to a singular vector.} Each cell shows the top-3 images from CC12M~\cite{changpinyo2021cc12m} that are most similar to a specific singular vector, along with the interpretation of that singular vector. See~\cref{sec:supp:qualitative} of \supmat for additional examples.}
\label{fig:image_matching}
\vspace{-1.5em}
\end{figure}

\begin{figure}[!t]
\centering
\includegraphics[width=0.9\linewidth]{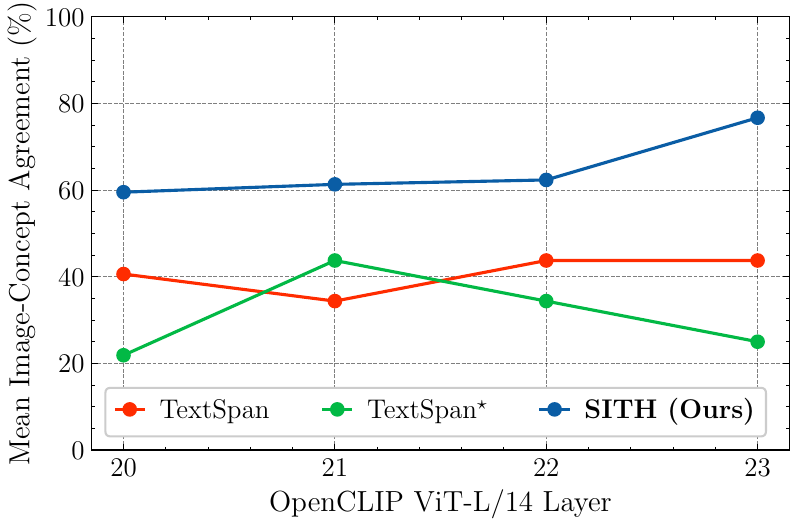}
\caption{\textbf{Mean image-concept agreement scores} across the last four layers of OpenCLIP~\cite{ilharco2021openclip}. $^\star$ replaces the original concept pool of TextSpan~\cite{gandelsman2023interpreting} with ConceptNet 5.5~\cite{speer2017conceptnet}.}
\label{fig:matching-score}
\end{figure}

To further verify that \method not only successfully reconstructs singular vectors, but also captures their actual visual focus, we conduct an image matching experiment on the large-scale CC12M dataset~\cite{changpinyo2021cc12m}.
For each image, we compute the similarity between the image embedding at layer $l$ and each singular vector.
We then rank images by similarity to each vector and retrieve the top matches.
As illustrated in \cref{fig:image_matching}, the retrieved images exhibit strong conceptual alignment with the semantic explanations assigned by \method, suggesting that the attributed meanings are indeed grounded in visual evidence.
To further quantify this alignment, we run an experiment where we ask a vision-LLM-as-a-judge to rate the correspondence between every set of retrieved images and their singular vector interpretation (see \cref{sec:supp:prompts} of the \supmat for the complete prompt).
As shown in \cref{fig:matching-score}, \method achieves consistently high matching scores across the final four layers of the model, particularly the last one.
We also evaluate TextSpan~\cite{gandelsman2023interpreting} under the same setting with both its original concept dictionary and ConceptNet 5.5~\cite{speer2017conceptnet}, showing that it yields coarser explanations that fail to capture the full function of the attention head.
These results demonstrate that \method produces fine-grained, coherent explanations for individual vectors, capturing distinct, interpretable visual functions.

\section{\method Enables Interpretable Model Editing}
\label{sec:editing}

One of the many strengths of \method is its ability to enable \emph{controlled editing of a model's behavior} through a fine-grained decomposition of internal representations. By adjusting the singular values associated with specific singular vectors, we can directly modulate how much the model relies on particular features or concepts during inference.
To determine which components to edit, we use an LLM~\cite{openai_gpt5_systemcard} to evaluate the relevance of each concept set for a given downstream task, guiding the amplification or suppression of singular values. This provides a lightweight, interpretable, and data-free alternative to \emph{singular value fine-tuning}~\cite{sun2022singular}, requiring no training data or gradient updates.
In the following, we show how \method can be applied to mitigate spurious correlations (\cref{sec:editing:spurious}), suppress unsafe concepts (\cref{sec:editing:safe}), and to improve zero-shot classification accuracy (\cref{sec:editing:classification}).

\subsection{Suppressing Spurious Correlations}
\label{sec:editing:spurious}
We can use the concepts provided by \method to identify and suppress features corresponding to confounding factors that should not influence predictions.
We follow~\citet{gandelsman2023interpreting} and test \method on the Waterbirds classification dataset~\cite{sagawa2019distributionally}, which contains birds on backgrounds that do not correspond to their natural habitats.
As shown in \cref{tab:spurious}, by removing the singular vectors whose concepts are related to background information (\ie, setting their corresponding singular values to $0$, see~\cref{sec:supp:implementation_details} of \supmat for additional details), we obtain a significant improvement, both in terms of overall accuracy and worst-group accuracy.
Notably, our method outperforms TextSpan~\cite{gandelsman2023interpreting}, highlighting the advantage of surgically editing only the subcomponents of an attention head, rather than removing entire heads.

\begin{table}[!t]
\centering
\caption{\textbf{Zero-shot classification accuracy on Waterbirds~\cite{sagawa2019distributionally}} before and after suppressing spurious features. 
$^\star$ indicates recomputed with zero-ablation of heads for zero-shot comparison.
``Random'' refers to average performance obtained by randomly ablating same number of singular values as our method, averaged over 5 runs.
}
\resizebox{0.7\columnwidth}{!}{%
\begin{tabular}{lcc}
\toprule
 & \textbf{Overall} & \textbf{Worst-group} \\
\midrule
OpenCLIP~\cite{ilharco2021openclip} & 73.5 & 47.9 \\
\ w/ Random & 72.0 & 45.1 \\
\ w/ TextSpan$^{\star}$~\cite{gandelsman2023interpreting} & 81.8 & 68.0 \\
\plusours & \textbf{82.7} & \textbf{70.6} \\
\bottomrule
\end{tabular}%
}
\label{tab:spurious}
\end{table}

\begin{table}[!t]
\centering
\caption{\textbf{Retrieval R@10 results on the ViSU~\cite{poppi2023removing} test set}.
Left columns show text-to-image and image-to-text performance when using textual and visual safe queries (\ie, \textbf{T} and \textbf{V}, respectively) on a pool of both safe and unsafe data. Right columns use unsafe textual and visual queries (\ie, \textbf{T*} and \textbf{V*}, respectively).
$^{\dagger}$ Safe-CLIP~\cite{ilharco2021openclip} is a training-based method specific for NSFW removal that uses OpenAI-CLIP~\cite{radford2021learning} pretraining.
}
\resizebox{0.85\linewidth}{!}{%
\begin{tabular}{l@{\hskip2pt}c@{\hskip6pt}c@{\hskip8pt}c@{\hskip6pt}c}
\toprule
& \multicolumn{2}{c}{\textbf{Safe Query}} & \multicolumn{2}{c}{\textbf{Unsafe Query}}\\
& \textbf{\smaller{T$\rightarrow$(V $\cup$ V$^{\star}$)}} & \textbf{\smaller{V$\rightarrow$(T $\cup$ T$^{\star}$)}} & \textbf{\smaller{T$^{\star}$$\rightarrow$(V $\cup$ V$^{\star}$)}} & \textbf{\smaller{V$^{\star}$$\rightarrow$(T $\cup$ T$^{\star}$)}} \\
\midrule
{\color{badgray}{Safe-CLIP}}$^{\dagger}$~\cite{poppi2023removing} & {\color{badgray}69.2} & {\color{badgray}73.9} & {\color{badgray}46.3} & {\color{badgray}62.3} \\
OpenCLIP~\cite{ilharco2021openclip} & \textbf{75.1} & 77.0 & 29.3 & 38.9 \\
\plusours & 74.5 & \textbf{77.3} & \textbf{29.5} & \textbf{40.4}  \\
\bottomrule
\end{tabular}}
\label{tab:visu_retrieval}
\end{table}

\begin{table}[!t]
\centering
\caption{\textbf{Zero-shot classification accuracy on various datasets} before and after editing the model by amplifying/suppressing specific singular values. 
}
\resizebox{0.85\columnwidth}{!}{%
\begin{tabular}{@{}lccc@{}}
\toprule
\multicolumn{1}{c}{} & Flowers 102~\cite{nilsback2008automated} & FGVC-Aircraft~\cite{maji13fine-grained}  & DTD~\cite{cimpoi2014describing}           \\ \midrule
OpenCLIP~\cite{ilharco2021openclip}             & 76.5              & 36.6          & 50.1          \\
 \ w/ Random               & 76.4              & 36.3              & 49.9          \\
\plusours                 & \textbf{77.5}     & \textbf{36.9} & \textbf{50.9} \\ \bottomrule
\end{tabular}%
}
\label{tab:improving_classification}
\end{table}

\subsection{Removing NSFW Concepts}
\label{sec:editing:safe}
\method can also enhance the safety of CLIP by suppressing its ability to process NSFW (Not Safe For Work) concepts such as nudity or violence.
This is achieved by identifying and suppressing the specific singular vectors associated with these undesired terms (see \cref{sec:supp:implementation_details} of the \supmat for additional details).
To evaluate the effectiveness of our intervention, we replicate the image-text retrieval experiment conducted by \citet{poppi2023removing} on the ViSU dataset,
which includes paired safe and unsafe image-caption samples.
The objective is to ensure that, regardless of the query term (\ie, image or text) being safe or unsafe, the model preferentially retrieves the corresponding safe content only.
As shown in~\cref{tab:visu_retrieval}, our modified CLIP demonstrates improved retrieval accuracy compared to the unmodified baseline, particularly when handling unsafe queries.
Notably, it even outperforms Safe-CLIP~\cite{poppi2023removing}, a model specifically trained for safety, when the input queries are safe.

\begin{figure*}[!ht]
    \centering
    \includegraphics[width=0.95\textwidth]{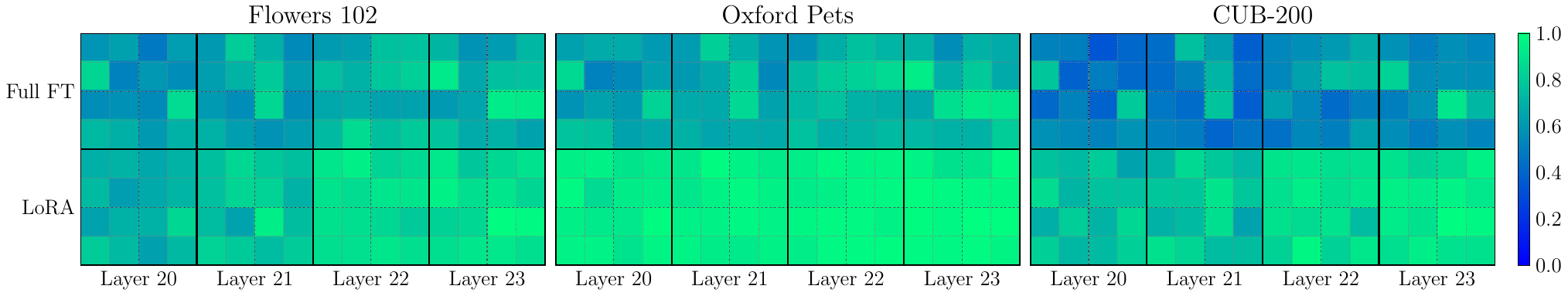}
    \caption{\textbf{Changes in the attention heads after model adaptation.} For each attention head (\ie square in the plot) of the last four layers of CLIP ViT-L/14, we report the normalized spectral cosine similarity~\cite{basile2025residual} between the right singular vectors of the pre-trained ($\matr{W}_{VO}^{pre}$) and finetuned ($\matr{W}_{VO}^{ft}$) matrices of that head, across different fine-tuning datasets and adaptation methods. High values correspond to subtle shifts in the head after fine-tuning.}
    \label{fig:spectral_sim}
\end{figure*}

\subsection{Improving Classification Performance}
\label{sec:editing:classification}
While~\cref{sec:editing:spurious,sec:editing:safe} address the suppression of undesirable concepts, here we aim to amplify features that are beneficial for a specific downstream classification task.
Given a task and its class labels, we compute a similarity score between the concept set of each singular vector and the class names.
We then rescale each singular value by a factor proportional to its similarity with the task, thus amplifying the contribution of relevant concepts and suppressing that of irrelevant ones (see \cref{sec:supp:implementation_details} of the \supmat for details). %
As shown in \cref{tab:improving_classification}, this simple intervention leads to consistent improvements across three different datasets~\cite{nilsback2008automated,maji13fine-grained,cimpoi2014describing}, achieving gains of up to $1.0$ points.
In contrast, randomly selecting the scaling factors leads to consistent performance degradation.
These results underscore the effectiveness of \method for
data-free singular value fine-tuning~\cite{sun2022singular}, enabling task-aware model adaptation without requiring labeled data.

\begin{figure}[!t]
    \centering
    \includegraphics[width=0.88\linewidth]{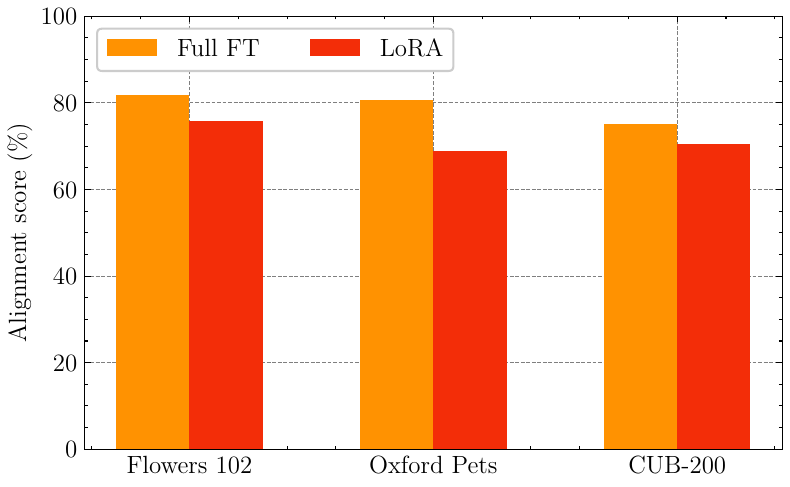}
    \caption{\textbf{Analysis of $\Delta \matr{W}$ with \method after model adaptation}.
    Using an LLM-as-a-judge we compute the percentage of task singular vectors~\cite{gargiulo2025task} aligned with the fine-tuning domain across different benchmarks and adaptation methods.}
    \label{fig:task_relevance}
    \vspace{-1em}
\end{figure}

\section{Interpreting Model Adaptation Techniques}
\label{sec:finetuning}

A key open question in mechanistic interpretability is understanding how pre-trained models adapt to new tasks through fine-tuning~\cite{sharkey2025open}. 
Here we show how \method can be used as a
tool to dissect the semantic changes induced by fine-tuning.
Specifically, we finetune CLIP ViT-L/14 on three fine-grained classification datasets (Flowers 102~\cite{nilsback2008automated}, Oxford Pets~\cite{parkhi2012cats}, and CUB-200~\cite{wah2011caltech}) and compare the original and finetuned attention head weights ($\matr{W}_{VO}^{pre}$ and $\matr{W}_{VO}^{ft}$, respectively). 
To ensure
our findings are not specific to full fine-tuning, we repeat our analysis also using LoRA \cite{hu2022lora} (see \cref{sec:supp:implementation_details} of the \supmat for further details).

\noindent \textbf{Fine-tuning Induces a Subtle Subspace Shift.}
To measure the geometric change to the value-output matrix induced by fine-tuning, we compute the SVD of both $\matr{W}_{VO}^{pre}$ and $\matr{W}_{VO}^{ft}$ and compare their right singular vector subspaces using normalized spectral cosine similarity~\cite{basile2025residual}.
As seen in~\cref{fig:spectral_sim}, the singular vectors remain remarkably stable across all adaptation methods, indicating that fine-tuning does not drastically alter the learned semantic basis, but rather applies a slight modification to the pre-trained weights.

\noindent \textbf{The Fine-tuning Delta is Semantically Aligned.}
To verify whether the subtle changes induced by fine-tuning have coherent semantic directions,
we analyze the task singular vectors~\cite{gargiulo2025task} (\ie, the singular vectors of the difference matrix $\Delta \matr{W} = \matr{W}_{VO}^{ft} - \matr{W}_{VO}^{pre}$) using our \comp method.
We asked an LLM to classify whether the explanations obtained by \comp for the task singular vectors are relevant to the fine-tuning domain. As shown in \cref{fig:task_relevance}, the majority of the task singular vectors are indeed aligned with the fine-tuning task across all datasets and adaptation methods.
For instance, when fine-tuning on Flowers 102, the top task singular vectors are associated to concepts from the plant domain, such as ``alpine flowers'', ``oriental photinia'', and ``camelia flower''. We observe similar patterns also for
the other two tasks, with 
Oxford Pets yielding concepts like ``english bulldog'', ``bordeaux mastiff'', and ``egyptian mau'', and CUB-200 producing bird-related concepts, including ``black catbird'', ``red legged tinamous'', and ``chiffchaff''.

\section{Conclusion}
We introduced \method, a novel data-free and training-free framework for interpreting the internal features used by CLIP's vision transformer. By applying Singular Value Decomposition directly to the Value-Output (VO) weight matrices, we isolated the computational directions within each attention head. We proposed \comp, a decomposition algorithm that translates these singular vectors into sparse, semantically coherent combinations of human-interpretable concepts. 
We demonstrated that this intra-head level of analysis is not only faithful and interpretable but also enables precise, interpretable model edits to suppress spurious correlations, remove undesirable content, and enhance downstream task performance without retraining. Additionally, it enables understanding that weights' updates are task-aligned during fine-tuning on downstream tasks.

\noindent \textbf{Future work} will extend the analysis to other components of VLMs, such as FFN layers, and to the Query-Key matrices to better understand how attention patterns arise from semantic features.

\section*{Acknowledgments}

The authors acknowledge the CINECA award under the ISCRA initiative for the availability of high-performance computing resources and support. This work was sponsored by the Italian Ministerial grants PRIN 2022: “BFAIR: Bias-Free Artificial Intelligence methods for automated visual Recognition” (CUP E53D23008010006), and the EU Horizon projects ELIAS (No. 101120237), ELLIOT (No. 101214398), TURING (No. 101215032), and IAMI (No. 101168272).

{
    \small
    \bibliographystyle{ieeenat_fullname}
    \bibliography{main}
}

\onecolumn
\clearpage
\setcounter{page}{1}

\appendix

{
\centering
        \Large
        \textbf{\thetitle}\\
        \vspace{0.5em}Supplementary Material \\
        \vspace{1.0em}
}

\startcontents[chapters]
\printcontents[chapters]{}{1}{}

\twocolumn

\clearpage

\section{Additional implementation details}
\label{sec:supp:implementation_details}

This section provides the theoretical foundations and additional implementation details of \method. We begin by detailing the mechanistic view of Multi-Head Attention (MHA) to show how we isolate the Value-Output (VO) weight matrix from each attention head (\cref{sec:supp:mha}). We also describe how we fold the Layer Normalization (LN) into the MHA weights to ensure that each attention head directly reads from the residual stream (\cref{sec:supp:folding_ln}). 

Next, we provide more details on how we project the singular vectors into the multimodal embedding space to minimize the modality gap and interpret them via the concept pool (\cref{sec:supp:projection}). We also detail how we project the reconstructed vectors back into the residual stream for evaluation purposes (\cref{sec:supp:reconstruction}).

Finally, we describe the steps for our model editing and model adaptation analysis experiments. This covers the methodology for pruning singular vectors to remove spurious features (\cref{sec:supp:spurious}) and NSFW concepts (\cref{sec:supp:safe}), the identification and amplification of task-relevant singular vectors to improve classification performance (\cref{sec:supp:classification}), and the specific settings used to analyze model adaptation under Full Fine-tuning and LoRA (\cref{sec:supp:finetuning}).

\subsection{Multi-head Attention}
\label{sec:supp:mha}

In the standard implementation of CLIP-ViT~\cite{dosovitskiy2020image}, the multi-head attention (MHA) mechanism is computed as the concatenation of the outputs of each attention head, followed by a linear transformation:
\begin{equation}
\text{MHA}(\matr{X}) = \text{Concat}(\text{H}_1(\matr{X}), \ldots, \text{H}_H(\matr{X}))\matr{W}_O ,
\end{equation}
where $\matr{X} \in \mathbb{R}^{(P+1) \times D}$ are the \texttt{[CLS]} token and input patches, $H$ is the number of attention heads, and $\matr{W}_O \in \mathbb{R}^{H \cdot d_h \times D}$ is the output weight matrix, with $d_h = D / H$ being the dimension of each head. Each attention head is computed as:
\begin{equation}
\text{H}_h(\matr{X}) = \text{softmax}\left(\frac{\matr{X}\matr{W}_Q^{h}(\matr{X}\matr{W}_K^{h})^T}{\sqrt{d_h}}\right)\matr{X}\matr{W}_V^{h},
\end{equation}
where $\matr{W}_Q^{h}, \matr{W}_K^{h}, \matr{W}_V^{h} \in \mathbb{R}^{D \times d_h}$ are the query, key, and value weight matrices for head $h$, respectively.

While standard, this implementation can make it challenging to apply mechanistic interpretability methods that aim to analyze the contributions of individual attention heads. To address this, we can express the MHA mechanism in a mathematically equivalent form that disentangles the contributions of each head and ensures that each attention head directly reads from and writes to the residual stream.

First, note that concatenation followed by a linear transformation can be expressed as a sum of linear transformations applied to each head's output. Indeed, given the output matrix $\matr{W}_O$, we can partition it into $H$ sub-matrices $\matr{W}_O^{h} \in \mathbb{R}^{d_h \times D}$, such that $\matr{W}_O = [\matr{W}_O^{1}; \ldots; \matr{W}_O^{H}]$. Then, we can rewrite the concatenation followed by the linear transformation as:
\begin{equation}
    [\text{H}_1(\matr{X}), \dots, \text{H}_H(\matr{X})] \begin{bmatrix}
        \matr{W}_O^1 \\
        \vdots \\
        \matr{W}_O^H
    \end{bmatrix} \\
    = \sum_{h=1}^H \text{H}_h(\matr{X}) \matr{W}_O^h .
\end{equation}
This allows us to express the MHA mechanism as the sum of $H$ independent attention heads:
\begin{equation}
\text{MHA}(\matr{X}) = \sum_{h=1}^H \text{H}'_h(\matr{X}),
\end{equation}
where we merge the output linear transformation into each head's computation as:
\begin{equation}
\text{H}'_h(\matr{X}) = \text{H}_h(\matr{X}) \matr{W}_O^{h} .
\end{equation}

Following this, we split the computation of each attention head into the Query-Key (QK) and Value-Output (VO) circuits.
Formally, given the formulation of each attention head as:
\begin{equation}
\text{H}'_h(\matr{X}) = \text{softmax}\left(\frac{\matr{X}\matr{W}_Q^{h}\matr{W}_K^{hT}\matr{X}^T}{\sqrt{d_h}}\right)\matr{X}\matr{W}_V^{h}\matr{W}_O^{h},
\end{equation}
we can merge the Query and Key weight matrices into a single matrix $\matr{W}_{QK}^{h} = \matr{W}_Q^{h}\matr{W}_K^{hT} \in \mathbb{R}^{D \times D}$, and the Value and Output weight matrices into another matrix $\matr{W}_{VO}^{h} = \matr{W}_V^{h}\matr{W}_O^{h} \in \mathbb{R}^{D \times D}$.
Consequently, each attention head can be expressed as:
\begin{equation}
\text{H}'_h(\matr{X}) = \text{softmax}\left(\frac{\matr{X}\matr{W}_{QK}^{h}\matr{X}^T}{\sqrt{d_h}}\right)\matr{X}\matr{W}_{VO}^{h},
\end{equation}
where the QK matrix governs how the attention weights are computed from the input patches, and the VO matrix determines how the attended patches are projected back into the residual stream.

\subsection{Folding LN into MHA}
\label{sec:supp:folding_ln}

The CLIP-ViT architecture utilizes a Pre-LN formulation~\cite{xiong2019on}, where the Layer Normalization (LN) operation is applied to the input of every Multi-Head Attention (MHA) and Feed-Forward Network (FFN) block, rather than to their outputs. Consequently, the attention heads do not directly read from the residual stream $\matr{X}$, but rather a normalized version $\text{LN}(\matr{X})$ of it. To ensure that each attention head reads directly from the residual stream, we fold the linear components of the LN layer into the MHA weights.

\noindent \textbf{Folding Affine Parameters.} Let $\vect{w}$ and $\vect{b}$ denote the learnable weight and bias parameters of the LN layer, respectively. We absorb these parameters into the query, key, and value projection matrices ($\matr{W}_Q, \matr{W}_K, \matr{W}_V$) and their respective biases ($\matr{b}_Q, \matr{b}_K, \matr{b}_V$) as follows:
\begin{align}
\matr{W}_{\{Q,K,V\}}' &= \text{diag}(\vect{w}) \matr{W}_{\{Q,K,V\}} \\ 
\vect{b}_{\{Q,K,V\}}' &= \vect{b}_{\{Q,K,V\}} + \matr{W}_{\{Q,K,V\}}^T \vect{b} 
\end{align}
where $\text{diag}(\vect{w})$ is a diagonal matrix with the elements of $\vect{w}$ on the diagonal.
This transformation ensures that the analysis of the folded weights $\matr{W}'$ accounts for the component-wise scaling and shifting applied by the LN.

\noindent \textbf{Handling Centering and Normalization.} Beyond the affine parameters, the core operation of Layer Normalization involves centering the input vector and scaling it to have unit variance. Since normalization does not affect the direction of the input vectors, we can safely omit it when analyzing the reading and writing directions of the attention heads. 

On the other hand, centering the input is equivalent to projecting the input vectors onto a hyperplane orthogonal to the all-ones vector $\vect{1}$, thus it changes the direction of the input vectors. However, in the CLIP ViT architecture, every transformer block and the final projection to the multimodal embedding space are preceded by a LayerNorm. This implies that any information encoded in the direction of the all-ones vector is systematically removed before it can be processed by subsequent layers or the final projection. 
Consequently, we can posit a theoretically equivalent model where every block is constrained to operate in (\ie, read from and write to) the subspace orthogonal to $\vect{1}$, thus making the explicit centering operation of the LN redundant.

To align our analysis with this effective computational model, we project the weight matrices onto the orthogonal complement of $\vect{1}$. This ensures that we only analyze the active subspace of the residual stream. Practically, this is implemented via mean subtraction~\cite{nanda2022transformerlens}.
For input-reading weights (\ie, $\matr{W}_Q, \matr{W}_K, \matr{W}_V$), we subtract the mean from each column. This ensures that the dot product with any vector parallel to $\vect{1}$ is zero, making the weights invariant to the mean of the input, which is removed nevertheless by the LN. For output-writing weights (\ie, $\matr{W}_O$), we subtract the mean from each row. This ensures that the output of the head sums to zero, guaranteeing that the head never writes into the $\vect{1}$ direction of the residual stream.

\noindent \textbf{Final Folded Weights.} After applying the affine folding and mean subtraction steps, we obtain the final folded weight matrices $\matr{W}_Q', \matr{W}_K', \matr{W}_V', \matr{W}_O'$ 
that are used in place of the original weights to derive the QK and VO weight matrices (see \cref{sec:supp:mha}).

\subsection{Projecting Singular Vectors into the Multimodal Space}
\label{sec:supp:projection}

In \cref{sec:method:comp} of the main paper, we showed that, as the left ($\vect{u} \in \matr{U}$) and right ($\vect{v} \in \matr{V}$) singular vectors of the VO matrix lie in the same residual stream space as the image patches, we can project them into the multimodal embedding space to interpret them via the concept pool $\conceptpool$.

Here we provide a more detailed explanation of this projection step.

\noindent \textbf{Layer Normalization.} 
At the end of the CLIP-ViT architecture,
the representation of the \texttt{[CLS]} token is passed through a final Layer Normalization (LN) layer before being projected by the vision projection matrix $\matr{W}_p$. To ensure that the singular vectors are correctly projected into the multimodal space, we similarly apply the LN transformation to them before the projection. Specifically, given a right singular vector $\vect{v} \in \mathbb{R}^D$ (the same applies to left singular vectors), we compute its unaligned multimodal representation as follows:
\begin{equation}
    \vect{\tilde{v}} = \text{norm}(\matr{W}_p^T\text{LN}(\vect{v}))
\end{equation}
where $\text{norm}(\cdot)$ indicates normalization to unit length.

\noindent \textbf{Mitigating the Multimodality Gap.} A well-documented phenomenon in contrastive vision-language models is the modality gap~\cite{liang2022mind}, where image and text embeddings tend to cluster in distinct, cone-shaped regions of the unit hypersphere. This gap can hinder the interpretability of singular vectors when projected into the multimodal space, as they may not align with the text embeddings of their corresponding concepts.

To address this gap, we adopt the re-centering approach proposed by \citet{bhalla2024interpreting}. Specifically, we geometrically align the two modalities by mean-centering both the projected singular vectors and the concept embeddings using the estimated means of the image and text embedding distributions:
\begin{align}
    \vect{\hat{v}} &= \text{norm}(\vect{\tilde{v}} - \vect{\mu}_{img}) \\
    \vect{\hat{\concept}}_i &= \text{norm}(\text{norm}(\textencoder(\concept_i)) - \vect{\mu}_{txt}) \quad \forall i=1, \ldots, C
\end{align}
where $\vect{\mu}_{img}$ is the mean image embedding computed over the CC12M dataset~\cite{changpinyo2021cc12m}, and $\vect{\mu}_{txt}$ is the mean text embedding computed over the text concepts in the concept pool $\conceptpool$.

\subsection{Projecting Reconstructions back into the Residual Stream}
\label{sec:supp:reconstruction}

To evaluate the fidelity of the \comp reconstructions, as well as to measure the effect on downstream performance when replacing singular vectors with their reconstructions, we need to project the decompositions made in the multimodal space back into the residual stream space.

Given the sparse coefficient vector $\vect{c} \in \mathbb{R}^C$ obtained via \comp and the matrix of aligned concept embeddings $\matr{\hat{\conceptpool}} \in \mathbb{R}^{C \times d}$, we first compute the reconstructed singular vector in the centered multimodal space as follows:
\begin{equation}
    \vect{\hat{v}}_{rec} = \text{norm}(\matr{\hat{\conceptpool}}^T \vect{c}) ,
\end{equation}
where the normalization ensures that the reconstructed vector lies on the unit hypersphere surface of the centered multimodal space. Then, to reverse the modality gap mitigation step, we move the reconstructed vector back into the image cone by adding the image mean and re-normalizing:
\begin{equation}
    \vect{\tilde{v}}_{rec} = \text{norm}(\vect{\hat{v}}_{rec} + \vect{\mu}_{img}) .
\end{equation}
To project the reconstructed vector back into the residual stream space, we utilize the Moore-Penrose pseudo-inverse of the projection matrix $\matr{W}_p$~\cite{moore1920reciprocal,bjerhammar1951application,penrose1955generalized}:
\begin{equation}
    \vect{v}'_{rec} = \matr{W}_p^{\dagger T} \vect{\tilde{v}}_{rec} ,
\end{equation}
where $\matr{W}_p^{\dagger} \in \mathbb{R}^{d \times D}$ is the pseudo-inverse matrix.

Theoretically, a complete inversion of the forward pass would require reversing the LayerNorm operation applied before the projection. 
However, we found that attempting to invert the affine transformation of the LN yielded suboptimal results.
Consequently, we omit the LN inversion entirely when projecting back into the residual stream.

Finally, as the singular vectors are defined to have unit norm, we re-normalize the reconstructed vector in the residual stream space:
\begin{equation}
    \vect{v}_{rec} = \text{norm}(\vect{v}'_{rec})
\end{equation}

\subsection{Spurious Feature Removal}
\label{sec:supp:spurious}

In \cref{sec:editing:spurious} of the main paper, we demonstrate how \method can be employed to identify which singular vectors encode spurious ``background'' or ``location'' features, and subsequently remove them from the model to enhance robustness against background and location biases on the Waterbirds classification dataset~\cite{sagawa2019distributionally}.
Given the high dimensionality of the search space (totaling 4,096 singular vectors across the last four layers of CLIP ViT-L/14), manually inspecting the semantic explanations generated by \comp for every vector is intractable. To automate this inspection process, we leverage an LLM as a semantic judge. For this specific intervention, we utilize \comp ($\lambda = 0.3$) with a sparsity level of $K=5$, ensuring that the explanations capture the dominant semantic concept encoded by each vector.

We employ GPT-5-mini~\cite{openai_gpt5_systemcard} to classify the degree to which the provided concept set relates to ``background'' or ``location'' features. The model assigns a relevance score on a Likert scale from 1 (``not related at all'') to 5 (``strongly related''); the exact prompt is provided in \cref{tab:eval_spurious_prompt_gpt5}. Finally, we apply a hard thresholding operation: any singular vector receiving a score $\geq 3$ is classified as spurious, and its corresponding singular value $\sigma_i$ is set to zero, effectively nullifying its contribution to the residual stream.

\subsection{Removing NSFW Concepts}
\label{sec:supp:safe}

In \cref{sec:editing:safe} of the main paper, we illustrated how \method can be utilized to identify and eliminate singular vectors that encode inappropriate or unsafe content, thereby enhancing the safety of the CLIP ViT-L/14 model for retrieval tasks. Here, we provide additional implementation details regarding this experiment.

Following the same automated discovery pipeline described in \cref{sec:supp:spurious}, we employ an LLM-as-a-judge to identify singular vectors encoding inappropriate content. Specifically, we prompt the LLM to evaluate singular vectors against the seven categories of inappropriate content defined by SafeCLIP~\cite{poppi2023removing}: \textit{hate}, \textit{harassment}, \textit{violence}, \textit{self-harm}, \textit{sexual content}, \textit{shocking images}, and \textit{illegal activity} (see \cref{tab:eval_safe_prompt_gpt5} for the exact prompt).

Different from the spurious feature removal task, here we apply a dual-threshold strategy. For vectors strongly related to unsafe categories (score $\geq 4$), we set the singular value $\sigma_i = 0$, thus removing the feature entirely. For vectors with a moderate relation to unsafe categories (score $= 3$), we set the singular value $\sigma_i = -1$. This effectively inverts the vector's contribution to the residual stream, thus pushing the representation away from the unsafe subspace.

\subsection{Improving Classification Performance}
\label{sec:supp:classification}

In \cref{sec:editing:classification} of the main paper, we showed that it is possible to use \method to enhance the classification accuracy of CLIP ViT-L/14 by amplifying and suppressing specific singular vectors. Here, we provide additional implementation details regarding this experiment.

\noindent \textbf{Identifying Task-Relevant Concepts.} Given a downstream classification task with a set of $M$ class labels $\mathcal{Y} = \{y_1, y_2, \dots, y_M\}$, we first aim to identify which concepts in our concept pool $\conceptpool$ are most relevant to the task. 
To do so, we decompose the embeddings of each class name ($\textencoder(y_m)$) into a set of constituent concepts using \comp.
To ensure the decomposition yields fundamental semantic attributes rather than trivial matches, we use a filtered concept dictionary $\conceptpool' = \conceptpool \setminus \mathcal{Y}$, where the class names themselves are removed from the candidate pool.

Given the set of concepts extracted for each class $y_m$, we define the union of these sets as the global task concept pool $\conceptpool_{task}$:
\begin{equation}
    \conceptpool_{task} = \bigcup_{m=1}^{M} \text{COMP}(\textencoder(y_m); \textencoder(\conceptpool')) .
\end{equation}
This pool represents the collection of semantic attributes (e.g., colors, shapes, textures, habitats) that are relevant to the classification task.

\noindent \textbf{Scoring Singular Vectors Relevance.} Given a right singular vector $\vect{v}_i$, \comp decomposes it into a set of $K$ pairs of coefficients and concepts $\{(w_{i,k}, \concept_{i,k})\}_{k=1}^K$, where $w_{i,k}$ is the importance weight and $\concept_{k} \in \conceptpool$ is the corresponding concept.

To quantify the relevance of the singular vector $\vect{v}_i$ to the classification task, we compute the weighted similarity between its constituent concepts and the task concept pool $\conceptpool_{task}$. Specifically, for each concept $\concept_{i,k}$ in the vector's explanation, we find its maximum cosine similarity with any concept in the task pool $\conceptpool_{task}$. We then weight this similarity by the corresponding coefficient $w_{i,k}$ and sum over all $K$ concepts to obtain the relevance score:
\begin{equation}
    \text{R}(\vect{v}_i) = \sum_{k=1}^{K} w_{i,k} \max_{\concept_{j} \in \conceptpool_{task}} \langle \textencoder(\concept_{i,k}), \textencoder(\concept_{j}) \rangle ,
\end{equation}
where $\langle \cdot, \cdot \rangle$ indicates the cosine similarity between two embeddings. This formulation ensures that a singular vector is considered relevant if its constituent concepts are semantically close to any concept required by the downstream task.

\noindent \textbf{Editing Singular Values.} To convert the relevance score $\text{R}(\vect{v}_i)$ into a scaling factor $\alpha_i$, we introduce a base threshold $\tau$. The purpose of $\tau$ is to shift the distribution of scores such that only highly relevant vectors are amplified ($>1.0$) while irrelevant ones are suppressed ($<1.0$).

To prevent the complete elimination of any singular vector, we also apply a clamping operation to ensure the scaling factor never drops below a minimum value of $0.8$. The final scaling factor $\alpha_i$ is computed as:
\begin{equation}
    \alpha_i = \max(0.8, \text{R}(\vect{v}_i) + \tau) .
\end{equation}

Finally, the original singular value $\sigma_i$ associated with the singular vector $\vect{v}_i$ is updated as follows:
\begin{equation}
    \sigma_i' = \alpha_i \cdot \sigma_i .
\end{equation}
This effectively performs a ``soft'' feature selection: vectors encoding semantic concepts unrelated to the task are dampened, while vectors aligned with the task's semantics are preserved or amplified.

\subsection{Model Adaptation}
\label{sec:supp:finetuning}

In this section, we provide additional implementation details regarding the fine-tuning analysis described in \cref{sec:finetuning} of the main paper, including training hyperparameters, the mathematical definition of the similarity metric used, and the LLM evaluation protocol.

\noindent \textbf{Training Details.} In our analysis of model adaptation, we examine how the value-output weight matrices evolve during fine-tuning. While \method analyzes the collapsed $\matr{W}_{VO}$ matrices, standard fine-tuning updates the parameter matrices $\matr{W}_V$ and $\matr{W}_O$ separately. Consistent with this, we fine-tune the pretrained value $\matr{W}_V^{pre}$ and output $\matr{W}_O^{pre}$ matrices of the last four layers of the OpenCLIP ViT-L/14 vision encoder, resulting in $\matr{W}_V^{ft}$ and $\matr{W}_O^{ft}$, respectively. Then, for a given attention head, we construct the post-adaptation VO matrix as $\matr{W}_{VO}^{ft} = \matr{W}_V^{ft} \matr{W}_O^{ft}$, and compare it to the pre-trained VO matrix $\matr{W}_{VO}^{pre} = \matr{W}_V^{pre} \matr{W}_O^{pre}$.

We perform fine-tuning on three fine-grained classification datasets: Flowers 102~\cite{nilsback2008automated}, Oxford Pets~\cite{parkhi2012cats}, and CUB-200~\cite{wah2011caltech}. For each dataset, we fine-tune the model for 10 epochs using a batch size of 64 and a learning rate of $1\times 10^{-4}$. For LoRA fine-tuning, we set the rank equal to 8 and the $\alpha$ to 16.

\noindent \textbf{Normalized Spectral Cosine Similarity.} To quantify the geometric shift in the semantic basis of the attention heads, we utilize the normalized spectral cosine similarity. This metric, adapted from \citet{basile2025residual}, measures the alignment between two sets of vectors in a way that is weighted by their importance.

To compute this metric, we iteratively match the singular vectors from the pre-trained and fine-tuned VO matrices based on their weighted cosine similarity, ensuring that each vector is only matched once. Formally, let $\mathbb{S}_{pre} = \{(\vect{v}_i^{pre}, \sigma_i^{pre})\}_{i=1}^r$ and $\mathbb{S}_{ft} = \{(\vect{v}_j^{ft}, \sigma_j^{ft})\}_{j=1}^r$ be the sets of right singular vectors and their associated singular values for a pre-trained and fine-tuned head, respectively. Furthermore, let $\mathcal{I}_n$ and $\mathcal{J}_n$ be the sets of indices of the singular vectors that have already been matched in the previous $n$ iterations (so that $\mathcal{I}_0 = \mathcal{J}_0 = \emptyset$). Then, we define the spectral cosine similarity for the $n$-th matched pair as follows:
\begin{equation}
s_n = \left[ \max_{i \notin \mathcal{I}_{n-1}, j \notin \mathcal{J}_{n-1}} | \langle \vect{v}_i^{pre}, \vect{v}_j^{ft} \rangle | \right] \sigma_i^{pre} \sigma_j^{ft} ,
\end{equation}
where we consider the absolute value of the cosine similarity as singular vectors are defined up to the sign (\ie, $\vect{v}$ and $-\vect{v}$ are both valid singular vectors).
The final Normalized Spectral Cosine Similarity is then computed as:
\begin{equation}
\text{Sim}(\mathbb{S}_{pre}, \mathbb{S}_{ft}) = \sqrt{\frac{\sum_{n=1}^r s_n^2}{\sum_{n=1}^r (\sigma_n^{pre} \sigma_n^{ft})^2}} .
\end{equation}
This metric ranges from 0 to 1, where 1 indicates that the two sets are perfectly aligned.

\noindent \textbf{LLM Evaluation Protocol.} To assess the semantic alignment of the task singular vectors with the fine-tuning domain, we employ an LLM-based evaluation protocol. Specifically, for each task singular vector we use \comp (with $\lambda = 0.3$ and sparsity budget $K=5$) to generate textual explanations that describe the concepts encoded by the vector. We then prompt GPT-5-mini~\cite{openai_gpt5_systemcard} to classify whether the concepts in each explanation are relevant to the fine-tuning domain on a binary scale (Yes/No). The prompts used for this evaluation are reported in \cref{tab:finetune_flowers_prompt_gpt5,tab:finetune_pets_prompt_gpt5,tab:finetune_cub200_prompt_gpt5}.
Finally, we compute the percentage of task singular vectors that are classified as relevant to the fine-tuning task for each dataset and adaptation method (see \cref{fig:task_relevance} of the main paper).

\begin{table*}[!th]
    \centering
    \caption{\textbf{Comparison of different concept pools} along four critical axes: scale (\ie, number of concepts), granularity, safety alignment, and language coverage. ConceptNet 5.5 outperforms the alternatives across all dimensions, making it the most suitable choice for our interpretability framework.}
    \label{tab:concept_pools}
    \begin{tabular}{lcccc}
        \toprule
        \textbf{Concept Pool} & \textbf{Scale} & \textbf{Granularity} & \textbf{Safety Alignment} & \textbf{Language Coverage} \\
        \midrule
        TextSpan~\cite{gandelsman2023interpreting} & 3498 & Low & High & English-only \\
        SpLiCE~\cite{bhalla2024interpreting} & 15K & Medium & High & English-only \\
        WordNet~\cite{miller-1994-wordnet} & 153K & High & Medium & English-only \\
        ConceptNet 5.5~\cite{speer2017conceptnet} & 1.35M & High & Low & Multilingual \\
        \bottomrule
    \end{tabular}
\end{table*}

\begin{figure*}[!th]
    \centering
    \includegraphics[width=0.9\linewidth]{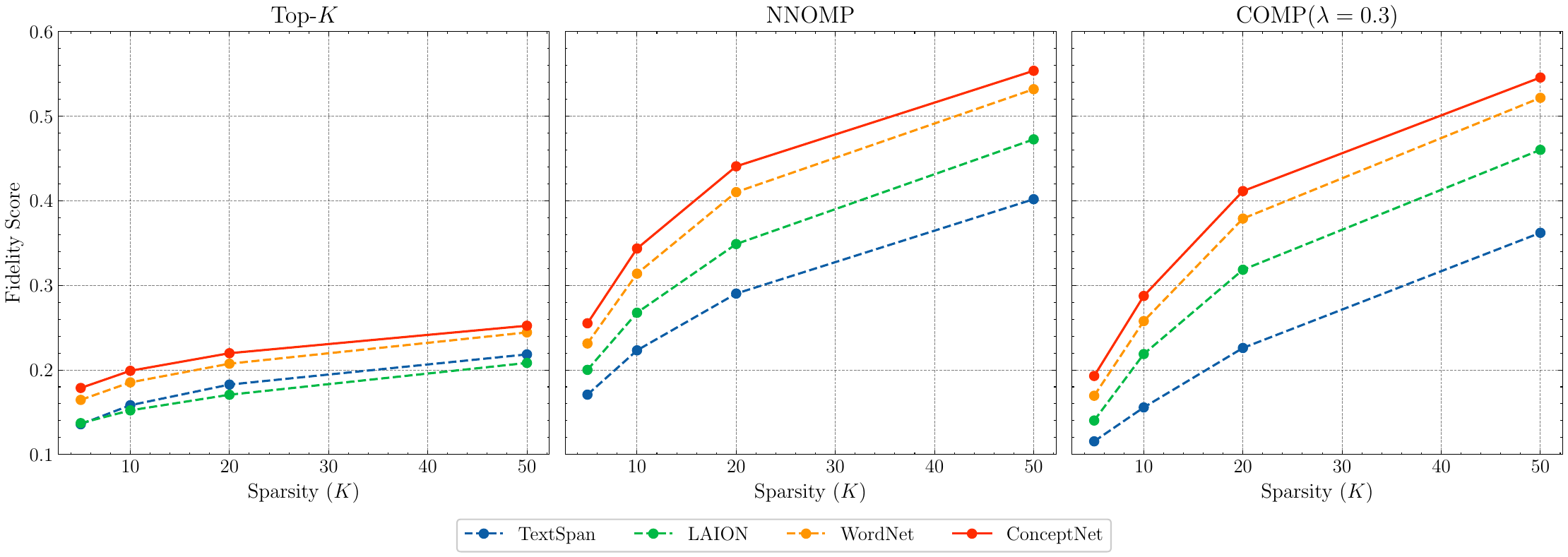}
    \caption{\textbf{Fidelity Score for different Concept Pools.} For each reconstruction method, we report the cosine-similarity (averaged across the last four layers of ViT-L-14) between the original singular vectors and their reconstructed versions at different sparsity levels and for different concept pools. We can see that for all methods and sparsity levels, ConceptNet guarantees superior reconstruction capabilities.}
    \label{fig:supp:concept_ablation}
\end{figure*}

\section{Ablating the Concept Pool}
\label{secc:supp:concept_ablation}

In this section, we evaluate \method against multiple concept dictionaries commonly used in the interpretability literature: TextSpan~\cite{gandelsman2023interpreting}, a highly curated set of 3,498 image descriptions generated by ChatGPT, SpLiCE~\cite{bhalla2024interpreting}, a frequency-based pool derived from LAION-400m~\cite{schuhmann2021laion}, containing the top-10k single-word and top-5k two-word concepts, and WordNet~\cite{miller-1994-wordnet}, a large lexical database of English.

\noindent \textbf{Dictionary Comparison}.
We evaluate these dictionaries based on four critical axes: scale (total number of concepts), granularity (ability to capture nuances), safety alignment (presence of NSFW concepts), and language coverage. A summary of this comparison is provided in \cref{tab:concept_pools}.

Existing dictionaries often fall short in one or more of these aspects. For instance, TextSpan, while highly curated, is limited in scale and thus struggles to cover the vast semantic space of CLIP. Furthermore, because it consists of short image descriptions, it tends to capture broad, scene-level summaries rather than the specific, fine-grained attributes that are often encoded by individual singular vectors.

Similarly, the SpLiCE pool presents challenges regarding granularity. To reduce redundancy, this pool aggressively removes concepts with high cosine similarity ($>0.9$). While this ensures diversity, it inadvertently eliminates semantic nuances, such as the distinction between ``cherry red'' and ``scarlet red'', which can be crucial for accurately interpreting singular vectors. Consequently, our sparse decomposition method (\comp) would be forced to select a more generic concept or combine multiple less relevant concepts, leading to less precise and interpretable explanations.

Another significant limitation shared by both the TextSpan and SpLiCE pools is their explicit filtering of unsafe content. Although removing NSFW terms is standard practice for generative applications, it is a severe limitation for mechanistic interpretability. As demonstrated in our experiments on NSFW removal (see \cref{sec:editing:safe}), CLIP natively encodes concepts related to nudity and violence within specific singular vectors. To successfully identify and suppress these concepts, the concept pool must first contain them.

\begin{figure*}[!th]
    \centering
    \includegraphics[width=\linewidth]{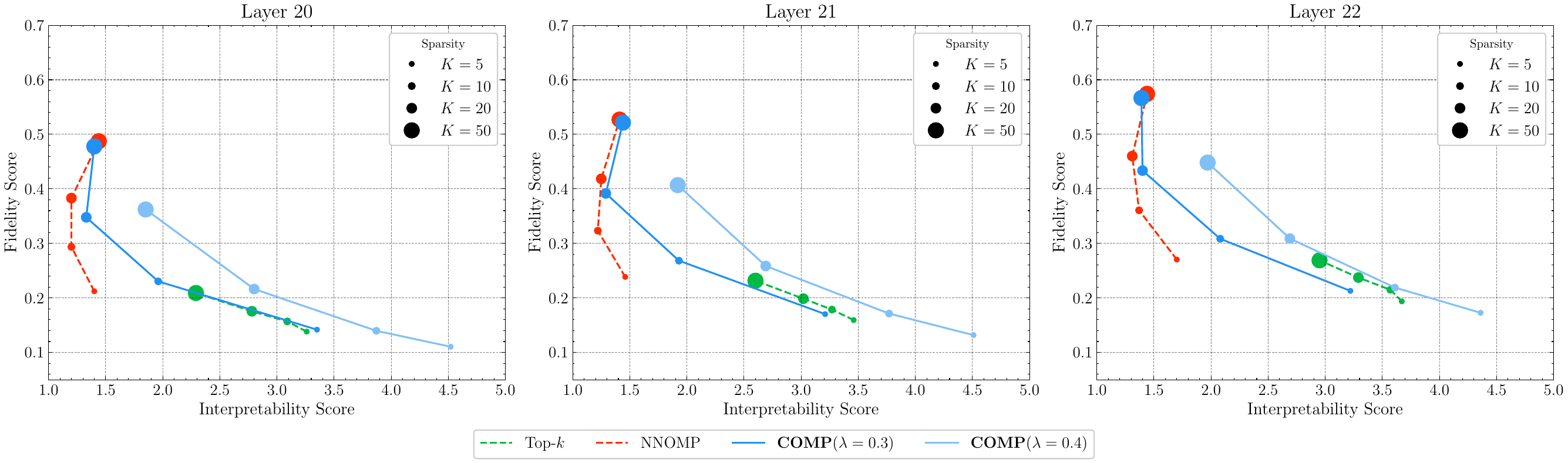}
    \caption{\textbf{Interpretability vs. Fidelity trade-off across Layers 20, 21, and 22 of CLIP ViT-L/14.} For each tested layer, we compare the performance of our proposed \comp method against two baselines: Top-$k$ selection and NNOMP~\cite{pati1993orthogonal}. Each point represents a different sparsity level ($K \in \{5, 10, 20, 50\}$). Across all layers, \comp consistently achieves a superior balance between interpretability and fidelity compared to the baselines.}
    \label{fig:supp:fidelity_interpretability_layers_20_21_22}
\end{figure*}

ConceptNet 5.5 addresses these limitations comprehensively. It offers a massive scale (more than 1.3 million concepts) that captures the long tail of semantic concepts, including synonyms and variations that allow for high-fidelity sparse approximations. Crucially, it retains NSFW concepts, which enables the safety interventions proposed in our main paper. Finally, unlike TextSpan, SpLiCE, and WordNet, which are strictly English-only, ConceptNet is multilingual. 

\noindent \textbf{Quantitative Ablation.} To 
verify that ConceptNet 5.5 better captures the semantic content of CLIP's weights, we evaluate the reconstruction fidelity of the singular vectors across the last four layers of the ViT-L/14 model using each of the four concept pools.
To ensure that our findings are not specific to our sparse decomposition method (\comp), we also evaluate two alternative decomposition techniques: Non-Negative Orthogonal Matching Pursuit (NNOMP) and Top-k selection. We also test varying the number of concepts selected ($k \in \{5, 10, 20, 50\}$) to assess the robustness of each dictionary across different sparsity levels.

As illustrated in \cref{fig:supp:concept_ablation}, ConceptNet 5.5 consistently outperforms the other concept pools across all reconstruction methods and sparsity levels. This confirms that the larger, more diverse search space of ConceptNet allows \method to find semantic combinations that more accurately approximate the singular vectors of CLIP.

\section{Extended Quantitative Analysis for CLIP ViT-L/14}
\label{sec:supp:vit_l14}

In this section, we extend the quantitative evaluation of \method presented in the main paper. We first analyze the interpretability-fidelity trade-off across earlier layers of the model (\cref{sec:supp:layers_20_22}) and then demonstrate the robustness of our approach by applying it to the left singular vectors (\cref{sec:supp:left_singular_vectors}).

\subsection{Interpretability-Fidelity Analysis on Additional Layers}
\label{sec:supp:layers_20_22}

In \cref{sec:eval_sparsity} of the main paper, we present the interpretability-fidelity trade-off for the last layer of CLIP ViT-L/14 ($l=23$). Here, we extend this analysis to the preceding layers $l \in \{20, 21, 22\}$ to verify the consistency of our findings across different depths of the network.

\noindent \textbf{Robustness of \comp.} As observed in the main paper (see \cref{fig:reconstruction_vs_monosemanticity}), the results for layers 20, 21, and 22 (see \cref{fig:supp:fidelity_interpretability_layers_20_21_22}) confirm that \comp consistently identifies the most favorable trade-off between reconstruction fidelity and semantic interpretability. While the baseline Top-$k$ approach yields high interpretability but poor fidelity, and NNOMP achieves high fidelity but produces polysemantic (and thus less interpretable) explanations, \comp successfully bridges this gap across all analyzed layers.

\noindent \textbf{Layer-wise Trends.} Beyond the relative performance of the methods, comparing the plots across layers reveals a clear trend: both fidelity and interpretability scores progressively improve as we move towards the last layer. We hypothesize that this phenomenon is driven by two primary factors: the semantic abstraction level of the features and the geometric alignment with the output space.

\begin{itemize}
    \item \textbf{Semantic Abstraction.} It is well-established in deep learning literature that shallower layers tend to encode lower-level features, while deeper layers capture higher-level, more abstract representations~\cite{zeiler2014visualizing,olah2017feature,dorszewski2025colors}. However, ConceptNet 5.5~\cite{speer2017conceptnet} predominantly consists of high-level semantic concepts. Consequently, reconstructing the lower-level singular vectors of earlier layers using a dictionary of high-level concepts is inherently more difficult, leading to lower fidelity scores.
    \item \textbf{Geometric Alignment.} \method relies on the model's final projection matrix $\matr{W}_p$ to map singular vectors from the residual stream to the multimodal space where the decomposition is then performed. However, in the standard forward pass of CLIP, $\matr{W}_p$ operates on the residual stream of the final layer $L$. While the singular vectors of layer $l$ reside in the residual stream space, the residual stream evolves as it passes through subsequent layers. Therefore, applying $\matr{W}_p$ to the weights of earlier layers introduces an approximation error due to the possible misalignment between the residual stream at layer $l$ and layer $L$. Furthermore, to evaluate the fidelity of a decomposition, we need to project the reconstruction from the multimodal space back into the residual stream; for shallower layers, this inversion likely incurs a greater approximation error, further degrading the fidelity score.
\end{itemize}

Despite these challenges, we note that \method with \comp maintains a superior Pareto frontier compared to baselines even in these earlier layers, demonstrating the method's robustness.

\subsection{Analysis of Left Singular Vectors}
\label{sec:supp:left_singular_vectors}

While the main text focuses on the right singular vectors $\matr{V}$, which define the directions the attention heads write to in the residual stream, the left singular vectors $\matr{U}$ play an equally critical role by defining the directions the heads read from the input. To demonstrate the generality of our approach, we replicate the interpretability-fidelity analysis from \cref{sec:eval_sparsity} on the left singular vectors of the last layer ($l=23$) of CLIP ViT-L/14.

\begin{figure}
    \centering
    \includegraphics[width=\linewidth]{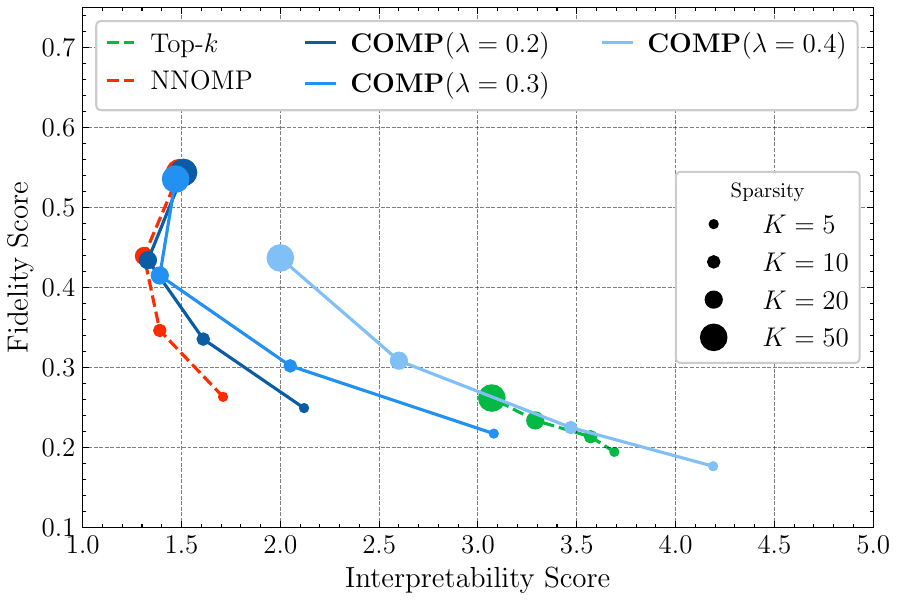}
    \caption{\textbf{Interpretability vs. Fidelity trade-off for the left singular vectors of the last layer of CLIP ViT-L/14}. We compare the performance of our proposed \comp against two baselines: Top-$k$ selection and NNOMP~\cite{pati1993orthogonal}. Each point represents a different sparsity level ($K \in \{5, 10, 20, 50\}$).}
    \label{fig:supp:fidelity_interpretability_left}
\end{figure}

\noindent \textbf{Effectiveness of \comp.} Consistent with our findings for the right singular vectors, the results shown in \cref{fig:supp:fidelity_interpretability_left} confirm that \comp achieves the most favorable trade-off between fidelity and interpretability when applied to left singular vectors, being able to faithfully explain them. The baseline methods exhibit the same limitations observed previously: Top-$k$ provides coherent but low-fidelity explanations, while NNOMP yields high fidelity at the cost of semantic coherence. \comp effectively balances these objectives, demonstrating that our decomposition algorithm is robust regardless of whether it is applied to the input or output space of the attention mechanism.

\noindent \textbf{Comparison with Right Singular Vectors.} When comparing the absolute performance scores between the two sets of vectors, we observe that the left singular vectors consistently exhibit slightly lower reconstruction fidelity scores than their right singular vector counterparts. We attribute this discrepancy to the geometric alignment issue discussed earlier. Indeed, right singular vectors represent updates being written forward into the stream, effectively aligning them closer to the final output representation. In contrast, left singular vectors represent features read from the incoming residual stream (i.e., the output of the preceding layer). Therefore, there is likely a greater misalignment between the left singular vectors and the final projection space. Consequently, projecting $\matr{U}$ into the semantic space incurs a higher approximation error, which slightly degrades the fidelity of the subsequent reconstruction.

\section{Qualitative Results}
\label{sec:supp:qualitative}

In this section, we provide a more extensive qualitative analysis of the interpretations produced by \method. 
We focus on the top-5 singular vector pairs (i.e., the pairs $(\vect{u}_i, \vect{v}_i)$ associated with the 5 largest singular values $\sigma_i$) for various attention heads. Indeed, the Singular Value Decomposition $\matr{W}_{VO} = \matr{U} \Sigma \matr{V}^T$ allows us to express the Value-Output matrix as a sum of rank-1 matrices: $\matr{W}_{VO} = \sum_{i=1}^{r} \sigma_i \vect{u}_i \vect{v}_i^T$. According to the Eckart-Young theorem~\cite{eckart1936approximation}, the sum of the first $k$ terms of this expansion provides the best rank-$k$ approximation of the original matrix in terms of the Frobenius norm. Therefore, the singular vectors associated with the largest singular values encode the most dominant reading ($\vect{u}_i$) and writing ($\vect{v}_i$) directions of the attention head, effectively defining its primary functional roles.

For each analyzed singular vector, we show: (1) the sparse concept set returned by \comp with $\lambda=0.3$ and $K=5$, (2) and the top-4 images from the CC12M dataset whose \texttt{[CLS]} token at layer $l$ (\ie, the layer of the analyzed attention head) has the highest cosine similarity with the singular vector. We report the results of the analyzed heads in \cref{tab:supp:vit-large14_l22_h02,tab:supp:vit-large14_l22_h03,tab:supp:vit-large14_l23_h00,tab:supp:vit-large14_l23_h04,tab:supp:vit-large14_l23_h08,tab:supp:vit-large14_l23_h11}.

\noindent \textbf{Left-Right Semantic Alignment.} We find that for many attention heads, the top singular vector pairs exhibit a strong semantic alignment between their reading and writing directions. For instance, in \cref{tab:supp:vit-large14_l23_h08}, all top-5 singular vector pairs of Head 8 in Layer 23 of ViT-L/14 are dedicated to colors, with each pair corresponding to a pair of colors, such as ``orange'' (reading) to ``purple'' (writing) and ``yellow'' (reading) to ``blue'' (writing). Similar patterns are also observed in other heads, such as Head 2 in Layer 22 (\cref{tab:supp:vit-large14_l22_h02}) and Head 4 in Layer 23 (\cref{tab:supp:vit-large14_l23_h04}).

\noindent \textbf{Intra-Head Semantic Alignment.} We observe that within many attention heads, the dominant directions are often semantically correlated, effectively grouped under a broader ``theme''. For instance, the Head 0 of Layer 23 in ViT-L/14 (see \cref{tab:supp:vit-large14_l23_h00}) is focused on materials, with each singular vector encoding a specific material type such as ``steel'', ``paper'', and ``glass''. Similarly, Head 11 of the same layer (see \cref{tab:supp:vit-large14_l23_h11}) captures letters, with singular vectors representing different characters like ``C'', ``M'' and ``S''. This head is particularly interesting as it also shows how CLIP is effectively able to read text in images. 
This intra-head semantic coherence suggests that certain attention heads are specialized in processing specific categories of information.

\noindent \textbf{Comparison with TextSpan.} Our findings strongly align with the head-level classifications provided by the activation-based method TextSpan~\cite{gandelsman2023interpreting}, as the function of many attention heads identified by TextSpan matches the concepts assigned to the top singular vectors.
However, \method offers a clear advantage in granularity: where TextSpan might broadly label a head as encoding certain colors, \method decomposes this behavior, identifying exactly which singular vector is responsible for ``red'', which for ``green'', etc. Furthermore, because \method is data-free, it identifies these functionalities solely from weights, avoiding the potential bias where a head might be mislabeled simply because the probing dataset lacks specific concept classes.

\noindent \textbf{Generalization across architectures, scales, and training regimes.} To demonstrate that the findings of \method are not limited to a specific model or training paradigm, we extend our qualitative analysis to a broader suite of vision-language models. Specifically, we evaluate OpenCLIP ViT-B/32 and ViT-H/14 to assess generalizability across different network capacities. Furthermore, to investigate the impact of architectural and data variations, we apply \method to MobileCLIP ViT-L/14~\cite{vasu2024mobileclip}, which builds upon the FastViT architecture~\cite{vasu2023fastvit} utilizing a highly optimized training regime.
The results, presented in \cref{tab:supp:vit-base32_l11_h02,tab:supp:vit-huge14_l31_h07,tab:supp:vit-huge14_l31_h11,tab:supp:vit-huge14_l31_h12,tab:supp:vit-huge14_l31_h13,tab:supp:mobile-large14_l22_h00,tab:supp:mobile-large14_l22_h10,tab:supp:mobile-large14_l23_h13}, show that the same semantic patterns identified in ViT-L/14 are consistently found across these diverse models. For instance, the color-related head identified in ViT-L/14 Layer 23 Head 8 (see \cref{tab:supp:vit-large14_l23_h08}) is also present in ViT-B/32 Layer 11 Head 2 (see \cref{tab:supp:vit-base32_l11_h02}) and ViT-H/14 Layer 31 Head 13 (see \cref{tab:supp:vit-huge14_l31_h13}), while the location-related head of ViT-L/14 Layer 23 Head 2 (see \cref{tab:supp:vit-large14_l22_h02}) is also found in ViT-H/14 Layer 31 Head 12 (see \cref{tab:supp:vit-huge14_l31_h12}) and MobileCLIP ViT-L/14 Layer 22 Head 10 (see \cref{tab:supp:mobile-large14_l22_h10}).

These findings further reinforce the \textit{Universality Hypothesis} in mechanistic interpretability, which posits that different neural networks converge on similar features and circuits when trained on similar data distributions~\cite{olah2020zoom}. This has been extensively explored in activation space, showing that distinct models converge toward shared representational spaces or ``Platonic'' concepts~\cite{dravid2023rosetta,huh2024position,thasarathan2025universal}. In contrast, our data-free weight-space analysis via \method reveals universality at the level of functional components (i.e., attention heads). Rather than just learning the same latent concepts in their activation spaces, models of vastly different capacities, architectures, and training paradigms allocate attention heads to perform identical, specialized semantic operations (e.g., colors, materials, or locations). This aligns with findings in the Large Language Model (LLM) literature, where specific attention head mechanisms, such as ``induction heads'' for in-context learning~\cite{olsson2022context} or ``successor heads'' for ordinal sequences~\cite{gould2024successor}, have been shown to universally emerge across diverse models. Our results extend this universality to vision-language models, demonstrating that the emergence of functionally specialized attention heads is a fundamental property of these models, transcending specific design choices, model scales, and training methodologies.

\begin{table*}
\centering

\caption{\textbf{Layer 22, Head 2 of ViT-L/14 encodes \textit{locations}}. The first 5 pairs of left/right singular vectors from Layer 22, Head 2 of ViT-L/14. For each singular vector, we display the top-4 images from CC12M~\cite{changpinyo2021cc12m} most similar to it, along with the explanation generated by \comp with $\lambda = 0.3$ and $K = 5$. \textcolor{red}{Unsafe text/images have been redacted/blurred.}}

\newcolumntype{I}{ >{\centering\arraybackslash} m{3cm} }

\begin{tabular}{ I m{0.25\textwidth} | m{0.25\textwidth} I }
\toprule

\textbf{Left}: house rooms & \multicolumn{2}{c}{\textbf{1st Singular Vector}} & \textbf{Right}: outdoor locations \\

\midrule

\includegraphics[width=3cm]{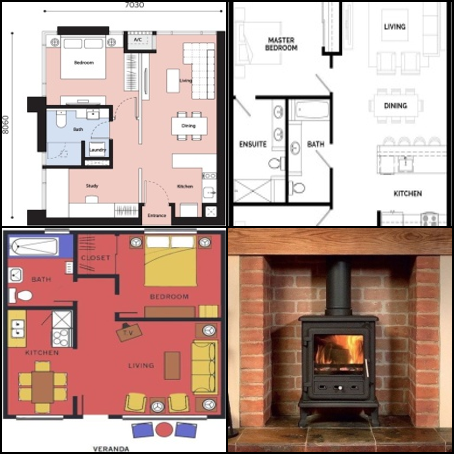} 
&
\begin{itemize}
	\item gogglebox (0.1730)
	\item cleaning living room (0.1713)
	\item hotel suite (0.1609)
	\item kitchen bedroom bathroom (0.1470)
	\item room of house (0.1068)
\end{itemize}
& 
\begin{itemize}
	\item outdoor concert (0.1937)
	\item park near mall entrance (0.1920)
	\item outside budget (0.1890)
	\item having s\censor{ex} outdoors (0.1883)
	\item outdoor stalls (0.1638)
\end{itemize}
&
\includegraphics[width=3cm]{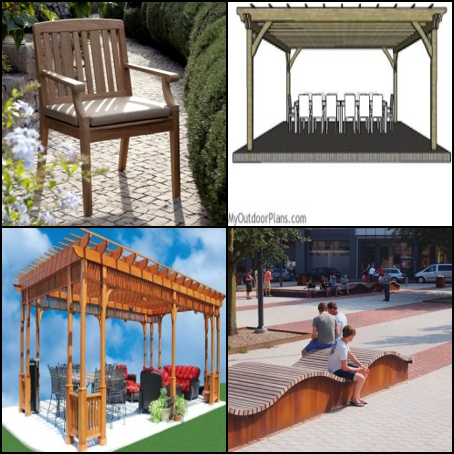} 
\\

\midrule

\textbf{Left}: street & \multicolumn{2}{c}{\textbf{2nd Singular Vector}} & \textbf{Right}: event rooms \\

\midrule

\includegraphics[width=3cm]{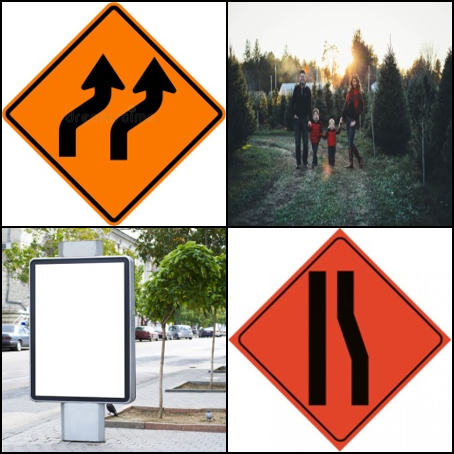} 
& 
\begin{itemize}
	\item roadside trees (0.1622)
	\item street price (0.1557)
	\item getting outdoors (0.1546)
	\item product for car owners (0.1370)
	\item street mobile (0.0820)
\end{itemize}
&
\begin{itemize}
	\item locker room (0.2331)
	\item conference center (0.1971)
	\item banquetting (0.1832)
	\item building institution club etc (0.1538)
	\item wedding hall (0.1244)
\end{itemize}
& 
\includegraphics[width=3cm]{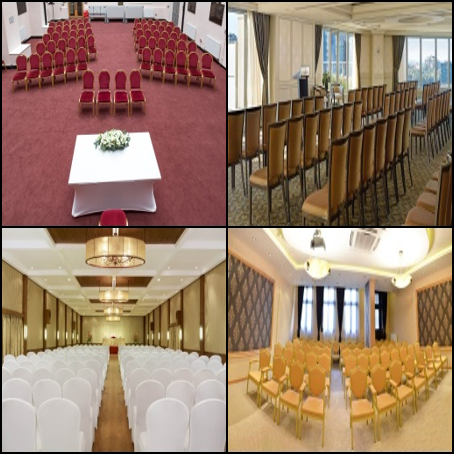} 
\\

\midrule 

\textbf{Left}: backyard & \multicolumn{2}{c}{\textbf{3rd Singular Vector}} & \textbf{Right}: shops \\

\midrule

\includegraphics[width=3cm]{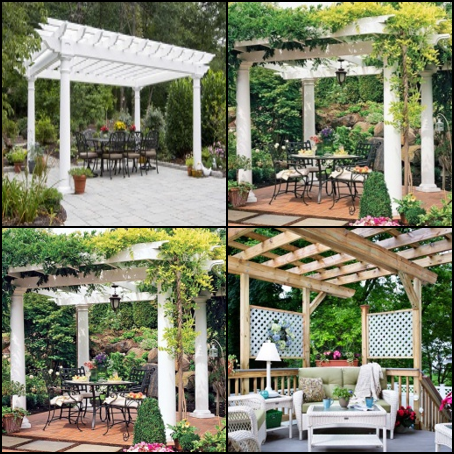} 
& 
\begin{itemize}
	\item balcony bra (0.2164)
	\item accommodation set aside for guests (0.2002)
	\item outdoor home (0.1697)
	\item lawn parties (0.1653)
	\item relaxing on porch (0.1228)
\end{itemize}
& 
\begin{itemize}
	\item checkout clerk at supermarket (0.1383)
	\item optical store (0.1296)
	\item store interior (0.1100)
	\item in bookstore (0.1051)
	\item computer store owner (0.0852)
\end{itemize}
& 
\includegraphics[width=3cm]{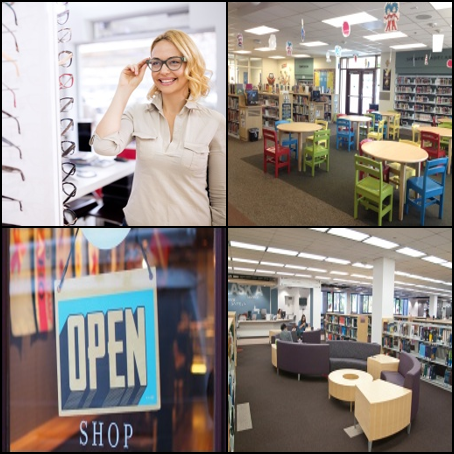} 
\\

\midrule 

\textbf{Left}: home areas & \multicolumn{2}{c}{\textbf{4th Singular Vector}} & \textbf{Right}: road, automobile \\

\midrule

\includegraphics[width=3cm]{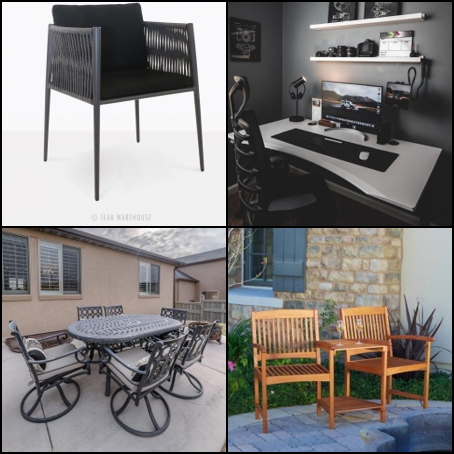} 
& 
\begin{itemize}
	\item play in back yard (0.1716)
	\item basement style foundation (0.1527)
	\item garden office (0.1522)
	\item work productive watch (0.1393)
	\item home office (0.1022)
\end{itemize}
& 
\begin{itemize}
	\item travel on roads (0.1859)
	\item nuptial procession (0.1784)
	\item escort bride (0.1742)
	\item automobile expo (0.1658)
	\item road maps (0.1534)
\end{itemize}
&
\includegraphics[width=3cm]{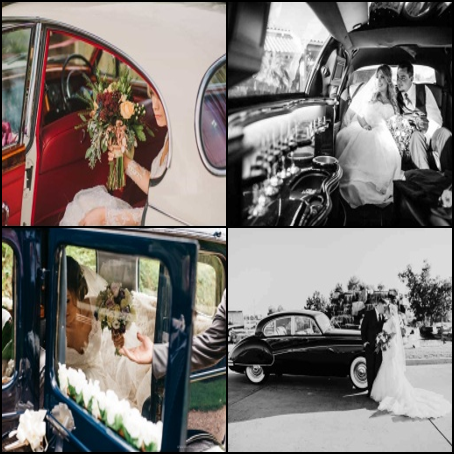} 
\\

\midrule 

\textbf{Right}: celebration events & \multicolumn{2}{c}{\textbf{5th Singular Vector}} & \textbf{Right}: traveling areas \\

\midrule

\includegraphics[width=3cm]{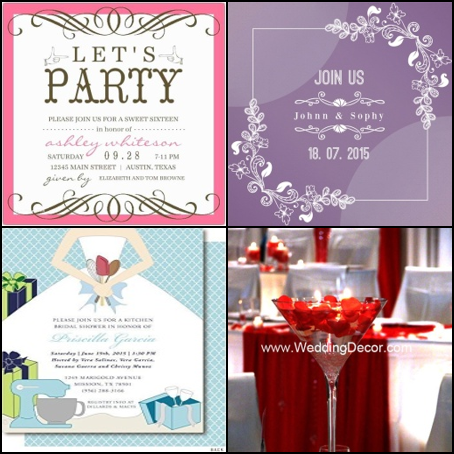} 
& 
\begin{itemize}
	\item wedding food (0.1940)
	\item sales event (0.1519)
	\item car festooned for celebration (0.1513)
	\item birthday banquet (0.1222)
	\item outdoor party (0.1138)
\end{itemize}
& 
\begin{itemize}
	\item travel from city to city (0.1948)
	\item most stadiums (0.1629)
	\item boarding bridge (0.1555)
	\item metro stations (0.1076)
	\item located between flights of stairs (0.1046)
\end{itemize}
& 
\includegraphics[width=3cm]{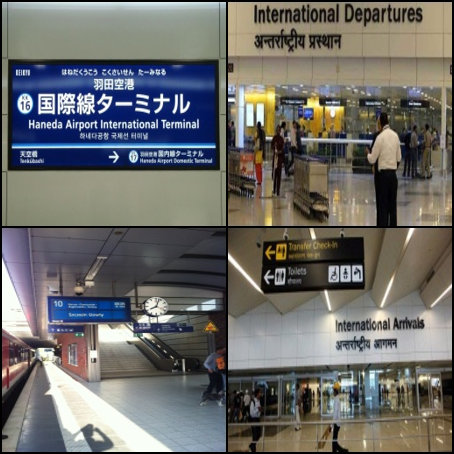} 
\\

\bottomrule
\end{tabular}
\label{tab:supp:vit-large14_l22_h02}
\end{table*}

\begin{table*}
\centering

\caption{\textbf{Layer 22, Head 3 of ViT-L/14 encodes \textit{objects}/\textit{body parts}}. The first 5 pairs of left/right singular vectors from Layer 22, Head 3 of ViT-L/14. For each singular vector, we display the top-4 images from CC12M~\cite{changpinyo2021cc12m} most similar to it, along with the explanation generated by \comp with $\lambda = 0.3$ and $K = 5$. \textcolor{red}{Unsafe text/images have been redacted/blurred.}}

\newcolumntype{I}{ >{\centering\arraybackslash} m{3cm} }

\begin{tabular}{ I m{0.25\textwidth} | m{0.25\textwidth} I }
\toprule

\textbf{Left}: objects for the upper torso & \multicolumn{2}{c}{\textbf{1st Singular Vector}} & \textbf{Right}: fabric items \\

\midrule

\includegraphics[width=3cm]{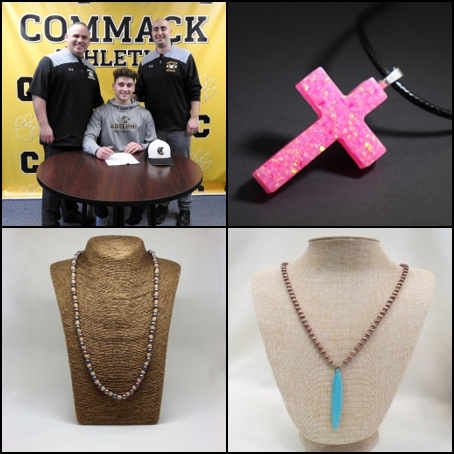} 
&
\begin{itemize}
	\item pullover sweater (0.1905)
	\item necklacings (0.1726)
	\item museum bust (0.1720)
	\item water on chest (0.1484)
	\item on chest (0.1204)
\end{itemize}
& 
\begin{itemize}
	\item frieze pants (0.1977)
	\item pair of short pantlegs (0.1654)
	\item curtains (0.1580)
	\item classic pants (0.1186)
	\item summer trousers (0.1148)
\end{itemize}
&
\includegraphics[width=3cm]{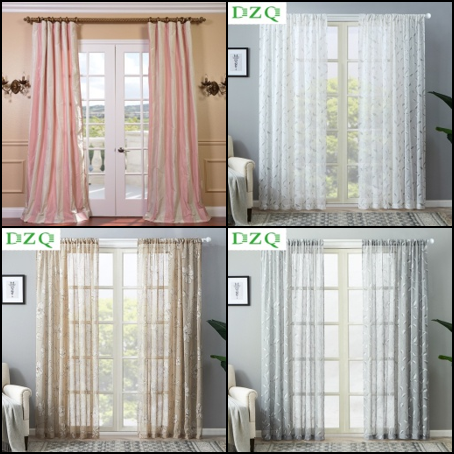} 
\\

\midrule

\textbf{Left}: cards & \multicolumn{2}{c}{\textbf{2nd Singular Vector}} & \textbf{Right}: shirt, table \\

\midrule

\includegraphics[width=3cm]{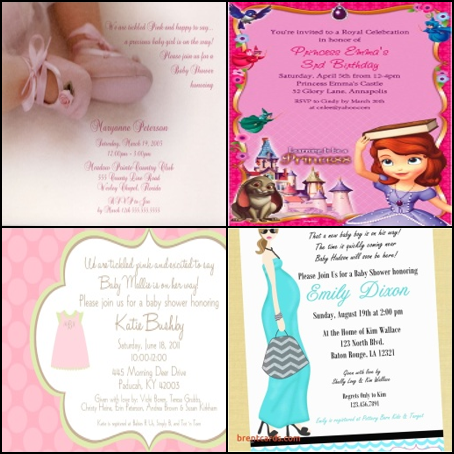} 
& 
\begin{itemize}
	\item long jacket (0.1569)
	\item invite job applications (0.1460)
	\item card case (0.1157)
	\item invitation card (0.0733)
	\item pc card (0.0571)
\end{itemize}
&
\begin{itemize}
	\item shirt arm (0.3040)
	\item nest table (0.1871)
	\item tablebases (0.1627)
	\item tabletopped (0.1433)
	\item table (0.0630)
\end{itemize}
& 
\includegraphics[width=3cm]{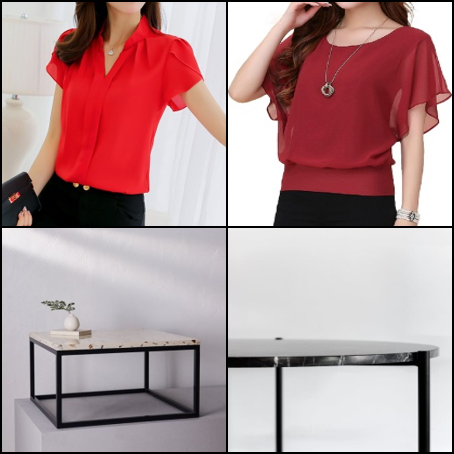} 
\\

\midrule 

\textbf{Left}: head parts & \multicolumn{2}{c}{\textbf{3rd Singular Vector}} & \textbf{Right}: dress \\

\midrule

\includegraphics[width=3cm]{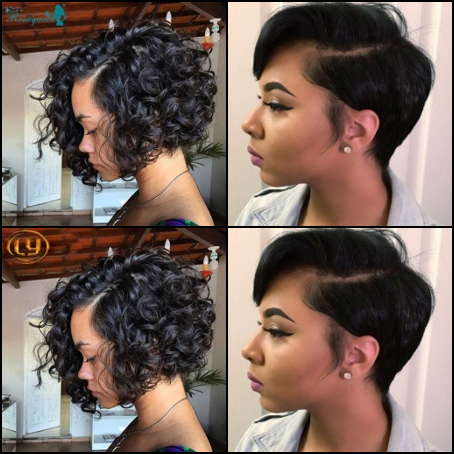} 
& 
\begin{itemize}
	\item open fence (0.1835)
	\item wig head (0.1754)
	\item hearing (0.1661)
	\item earsies (0.1175)
	\item human ears (0.0388)
\end{itemize}
& 
\begin{itemize}
	\item iliotibial band (0.2007)
	\item dress and skirts (0.1926)
	\item shirtdress (0.1881)
	\item drawer under telephone (0.1648)
	\item under dress (0.1175)
\end{itemize}
& 
\includegraphics[width=3cm]{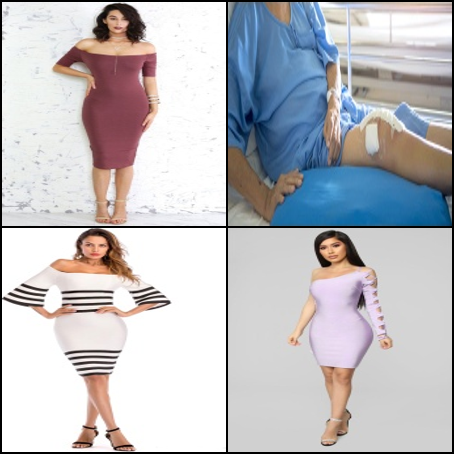} 
\\

\midrule 

\textbf{Left}: dorsal region & \multicolumn{2}{c}{\textbf{4th Singular Vector}} & \textbf{Right}: body parts \\

\midrule

\includegraphics[width=3cm]{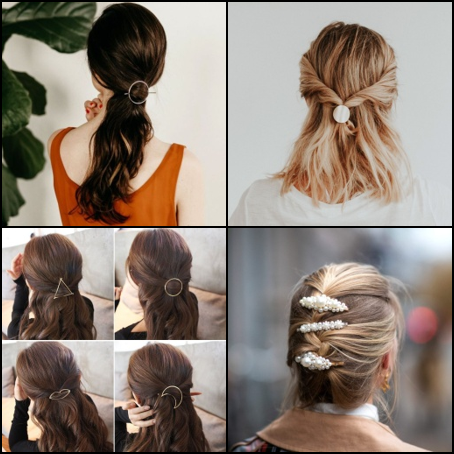} 
& 
\begin{itemize}
	\item backpack (0.1874)
	\item hairpin for bun (0.1719)
	\item shoulder blade (0.1514)
	\item back tee (0.1467)
	\item pin back hair (0.1119)
\end{itemize}
& 
\begin{itemize}
	\item facial \censor{cum shot} (0.1993)
	\item inregisters (0.1785)
	\item polydactylies (0.1489)
	\item feet touch cold floor (0.1425)
	\item foot \censor{jobs} (0.0561)
\end{itemize}
&
\includegraphics[width=3cm]{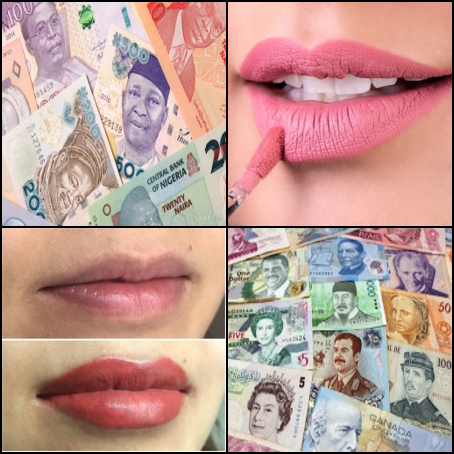} 
\\

\midrule 

\textbf{Left}: ceiling & \multicolumn{2}{c}{\textbf{5th Singular Vector}} & \textbf{Right}: miscellaneous objects \\

\midrule

\includegraphics[width=3cm]{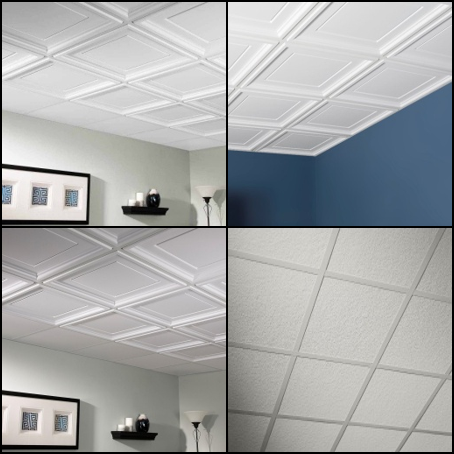} 
& 
\begin{itemize}
	\item ceiling floor (0.1770)
	\item outdoor ceiling (0.1170)
	\item ceiling under roof (0.0685)
	\item something ceiling (0.0683)
	\item ceiling (0.0138)
\end{itemize}
& 
\begin{itemize}
	\item controlling wrist (0.1977)
	\item short jacket (0.1939)
	\item cool tankards (0.1829)
	\item small mug (0.1414)
	\item wrist timepiece (0.0860)
\end{itemize}
& 
\includegraphics[width=3cm]{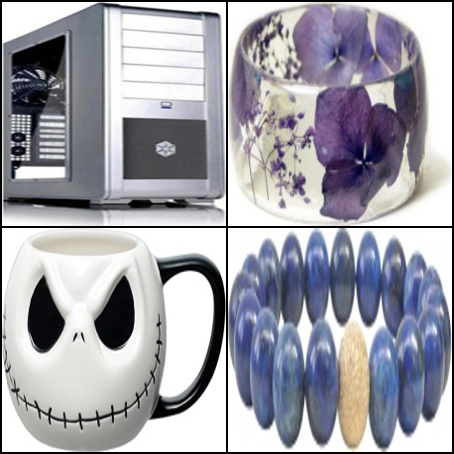} 
\\

\bottomrule
\end{tabular}
\label{tab:supp:vit-large14_l22_h03}
\end{table*}

\begin{table*}
\centering

\caption{\textbf{Layer 23, Head 0 of ViT-L/14 encodes \textit{materials}}. The first 5 pairs of left/right singular vectors from Layer 23, Head 0 of ViT-L/14. For each singular vector, we display the top-4 images from CC12M~\cite{changpinyo2021cc12m} most similar to it, along with the explanation generated by \comp with $\lambda = 0.3$ and $K = 5$.}

\newcolumntype{I}{ >{\centering\arraybackslash} m{3cm} }

\begin{tabular}{ I m{0.25\textwidth} | m{0.25\textwidth} I }
\toprule

\textbf{Left}: clothing & \multicolumn{2}{c}{\textbf{1st Singular Vector}} & \textbf{Right}: wax et similia \\

\midrule

\includegraphics[width=3cm]{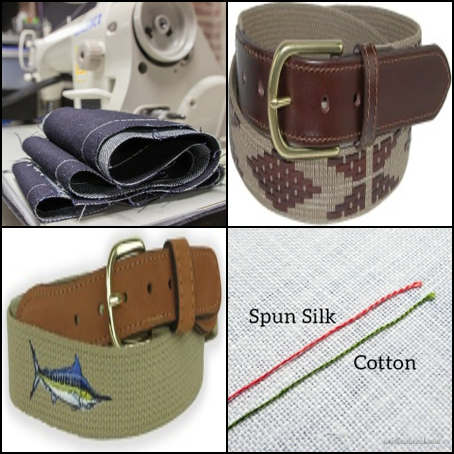} 
&
\begin{itemize}
	\item interknits (0.1811)
	\item comfortable clothes (0.1682)
	\item embroidered into cloth with sewing (0.1634)
	\item made from cloth (0.1520)
	\item cloth car seat (0.1429)
\end{itemize}
& 
\begin{itemize}
	\item leathery skin (0.2421)
	\item encaustics (0.1890)
	\item persulphates (0.1549)
	\item gloss stick (0.1489)
	\item rubber and latex (0.1392)
\end{itemize}
&
\includegraphics[width=3cm]{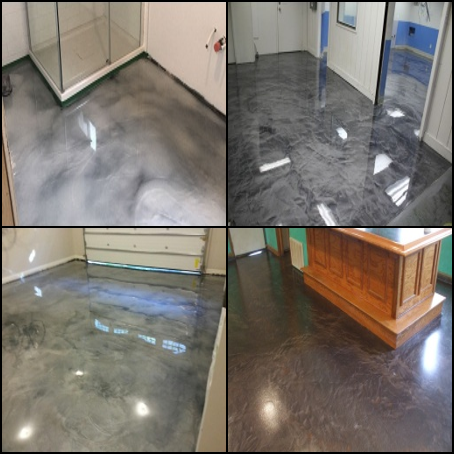} 
\\

\midrule

 \textbf{Left}: metal & \multicolumn{2}{c}{\textbf{2nd Singular Vector}} & \textbf{Right}: leather \\

\midrule

\includegraphics[width=3cm]{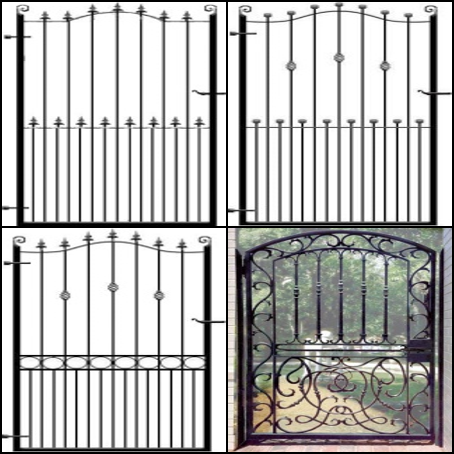} 
& 
\begin{itemize}
	\item steel aluminium (0.1945)
	\item played with small metal rod (0.1439)
	\item metal helmet (0.1322)
	\item metal jewelry (0.1264)
	\item piece of metal furniture (0.1173)
\end{itemize}
&
\begin{itemize}
	\item embroidered into cloth with sewing (0.1977)
	\item suede leather (0.1530)
	\item goatskins (0.1333)
	\item fruit leather (0.1299)
	\item soft leather (0.1166)
\end{itemize}
& 
\includegraphics[width=3cm]{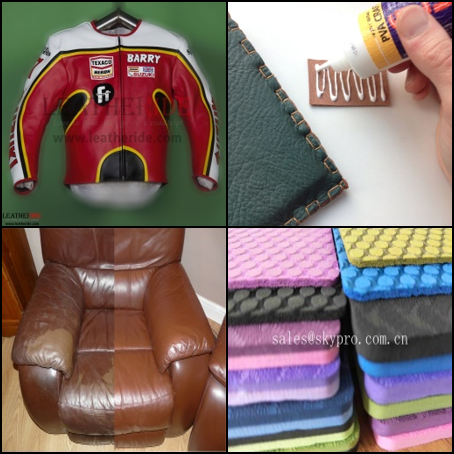} 
\\

\midrule 

\textbf{Left}: food & \multicolumn{2}{c}{\textbf{3rd Singular Vector}} & \textbf{Right}: leather \\

\midrule

\includegraphics[width=3cm]{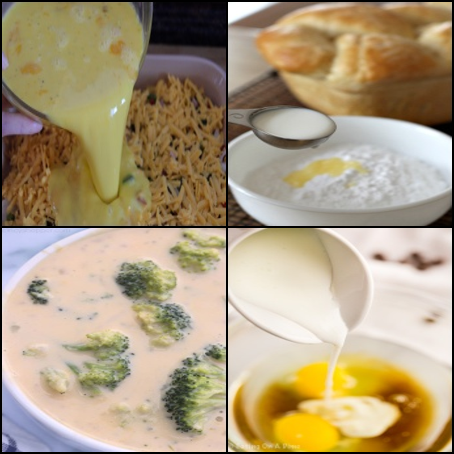} 
& 
\begin{itemize}
	\item food color (0.1818)
	\item look at paintings of food (0.1671)
	\item drinking yogurt (0.1571)
	\item packaged breakfast food (0.1420)
	\item peanut pastes (0.1356)
\end{itemize}
& 
\begin{itemize}
	\item leather flower (0.1400)
	\item leather trades (0.1352)
	\item leatherwork (0.1072)
	\item upholstered with leather (0.1056)
	\item leather case (0.1003)
\end{itemize}
& 
\includegraphics[width=3cm]{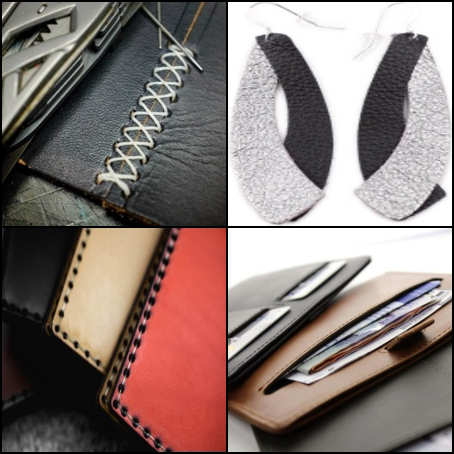} 
\\

\midrule 

\textbf{Left}: plastic & \multicolumn{2}{c}{\textbf{4th Singular Vector}} & \textbf{Right}: steel + drink \\

\midrule

\includegraphics[width=3cm]{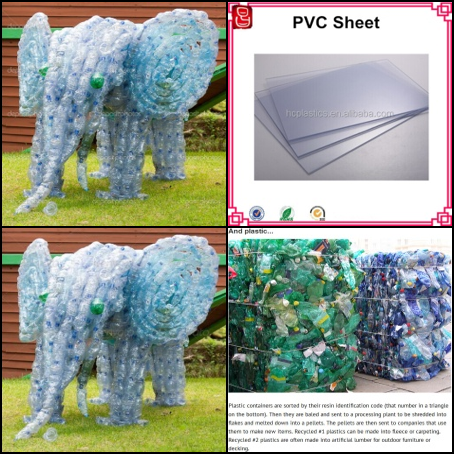} 
& 
\begin{itemize}
	\item knittabilities (0.2002)
	\item wood and plastic (0.1860)
	\item credit plastic (0.1829)
	\item plastic art (0.1398)
	\item plastic furniture (0.1332)
\end{itemize}
& 
\begin{itemize}
	\item comforting drink (0.2469)
	\item paramount titles (0.1969)
	\item steel wines (0.1965)
	\item stainless iron (0.1802)
\end{itemize}
&
\includegraphics[width=3cm]{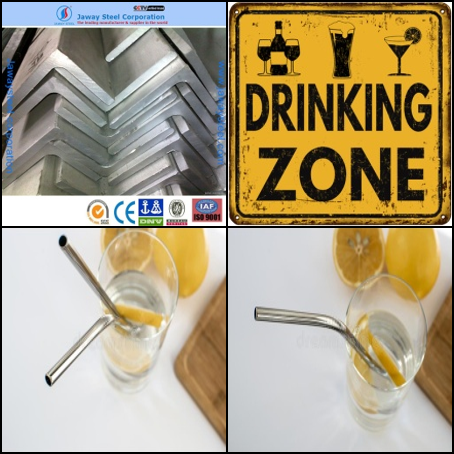} 
\\

\midrule 

\textbf{Left}: glass & \multicolumn{2}{c}{\textbf{5th Singular Vector}} & \textbf{Right}: paper \\

\midrule

\includegraphics[width=3cm]{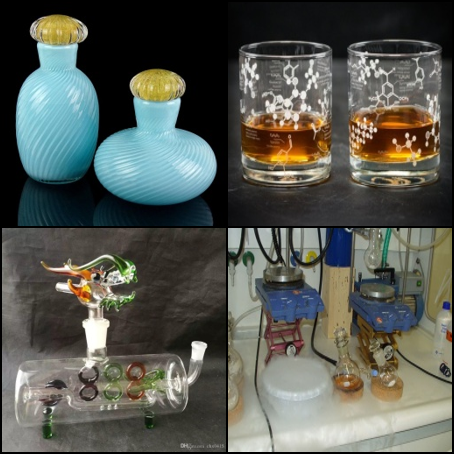} 
& 
\begin{itemize}
	\item glass making (0.2246)
	\item nanotherapeutics (0.1837)
	\item vodka luges (0.1649)
	\item mold on liquids (0.1409)
	\item glass ingredient (0.1257)
\end{itemize}
& 
\begin{itemize}
	\item uncensorship (0.1668)
	\item putting images on paper (0.1632)
	\item posting children's art work on (0.1536)
	\item paper tickets (0.1452)
	\item literary journalism (0.1344)
\end{itemize}
& 
\includegraphics[width=3cm]{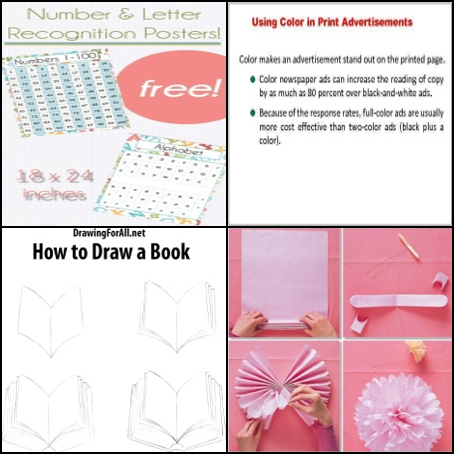} 
\\

\bottomrule
\end{tabular}
\label{tab:supp:vit-large14_l23_h00}
\end{table*}

\begin{table*}
\centering

\caption{\textbf{Layer 23, Head 4 of ViT-L/14 encodes \textit{people}}. The first 5 pairs of left/right singular vectors from Layer 23, Head 4 of ViT-L/14. For each singular vector, we display the top-4 images from CC12M~\cite{changpinyo2021cc12m} most similar to it, along with the explanation generated by \comp with $\lambda = 0.3$ and $K = 5$. \textcolor{red}{Unsafe text/images have been redacted/blurred.}}

\newcolumntype{I}{ >{\centering\arraybackslash} m{3cm} }

\begin{tabular}{ I m{0.25\textwidth} | m{0.25\textwidth} I }
\toprule

\textbf{Left}: women & \multicolumn{2}{c}{\textbf{1st Singular Vector}} & \textbf{Right}: men \\

\midrule

\includegraphics[width=3cm]{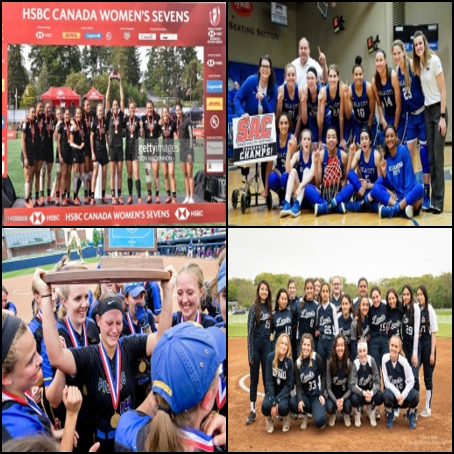} 
&
\begin{itemize}
	\item group of groups of women (0.2365)
	\item saint agnes eves (0.2199)
	\item women's writing (0.1835)
	\item girlswear (0.1670)
	\item business girl (0.1587)
\end{itemize}
& 
\begin{itemize}
	\item father and son (0.1844)
	\item men's aesthetic (0.1623)
	\item gay male s\censor{ex} (0.1530)
	\item groomsmen (0.1394)
	\item sexy guys (0.0826)
\end{itemize}
&
\includegraphics[width=3cm]{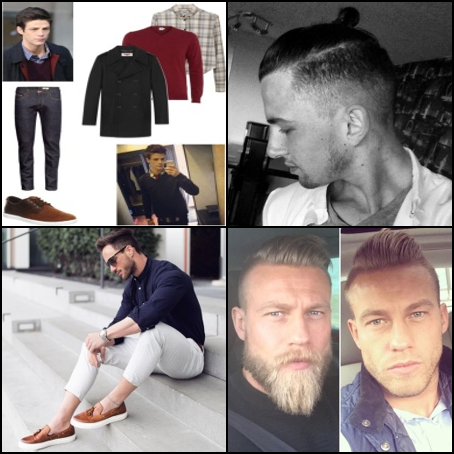} 
\\

\midrule

\textbf{Left}: boys & \multicolumn{2}{c}{\textbf{2nd Singular Vector}} & \textbf{Right}: couples \\

\midrule

\includegraphics[width=3cm]{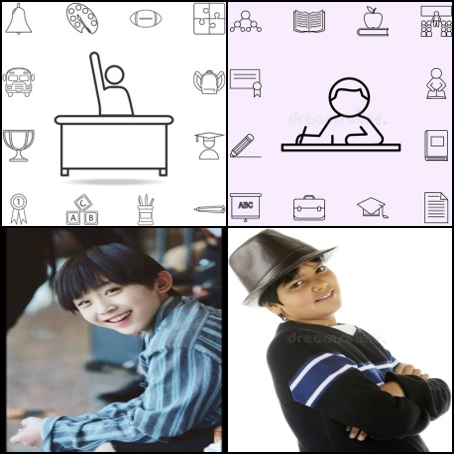} 
& 
\begin{itemize}
	\item young boy (0.1956)
	\item solo comedian (0.1526)
	\item child in camp (0.1440)
	\item juvenile vagrant (0.1348)
	\item transit worker (0.1348)
\end{itemize}
&
\begin{itemize}
	\item unhappy couples (0.2710)
	\item tag team (0.2025)
	\item co founders (0.1860)
	\item tandem bicycles (0.1748)
	\item couples together (0.0864)
\end{itemize}
& 
\includegraphics[width=3cm]{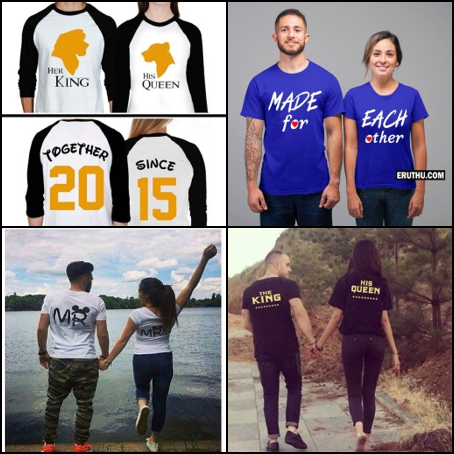} 
\\

\midrule 

\textbf{Left}: kids & \multicolumn{2}{c}{\textbf{3rd Singular Vector}} & \textbf{Right}: spouse \\

\midrule

\includegraphics[width=3cm]{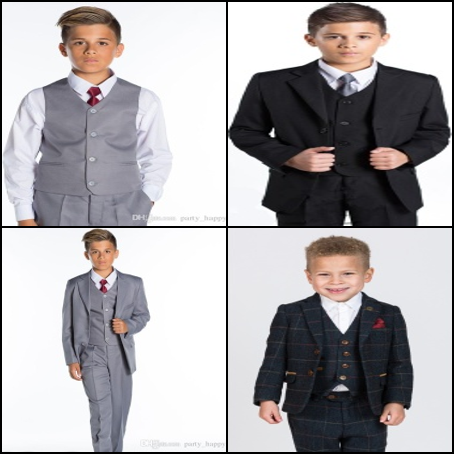} 
& 
\begin{itemize}
	\item junior bridesmaids (0.2663)
	\item generic kids (0.1880)
	\item children training (0.1834)
	\item kids who brothers (0.1712)
	\item kids together (0.0932)
\end{itemize}
& 
\begin{itemize}
	\item man health worker (0.1322)
	\item clumsy spouse (0.1137)
	\item heterosexual woman in love (0.1052)
	\item testing routine on spouse (0.1015)
	\item spouse scares (0.0781)
\end{itemize}
& 
\includegraphics[width=3cm]{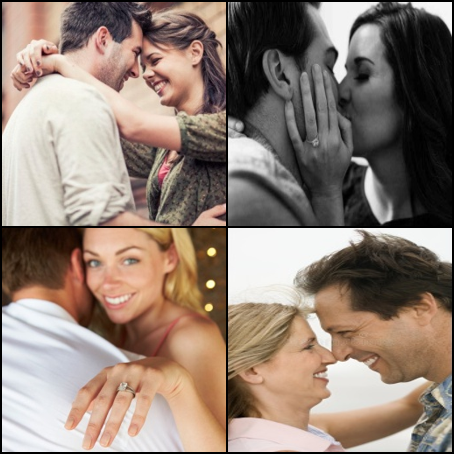} 
\\

\midrule 

\textbf{Left}: two (people) & \multicolumn{2}{c}{\textbf{4th Singular Vector}} & \textbf{Right}: group \\

\midrule

\includegraphics[width=3cm]{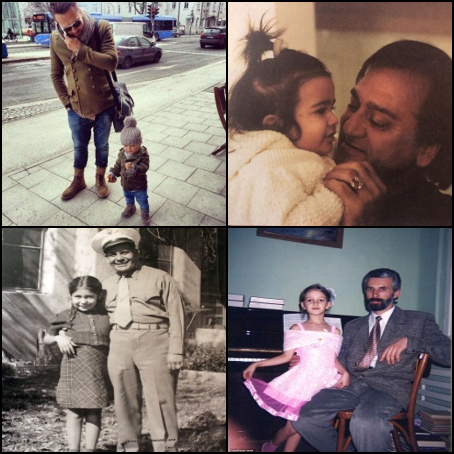} 
& 
\begin{itemize}
	\item father and daughter (0.3078)
	\item two children (0.1726)
	\item both periodicals (0.1435)
	\item both cities in california (0.1289)
	\item two pennies worth (0.1212)
\end{itemize}
& 
\begin{itemize}
	\item vocal quintet (0.1821)
	\item group of homosexuals (0.1609)
	\item group complaining (0.1424)
	\item group of men (0.0983)
	\item five man group (0.0679)
\end{itemize}
&
\includegraphics[width=3cm]{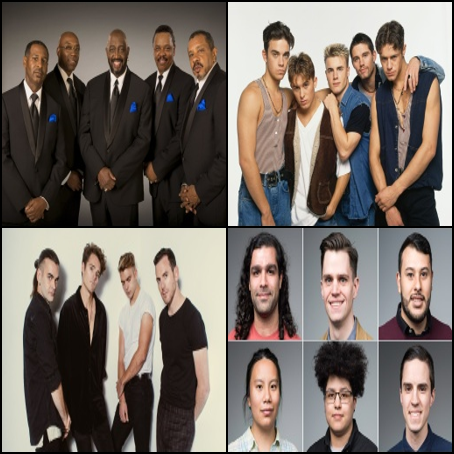} 
\\

\midrule 

\textbf{Left}: two people & \multicolumn{2}{c}{\textbf{5th Singular Vector}} & \textbf{Right}: girl \\

\midrule

\includegraphics[width=3cm]{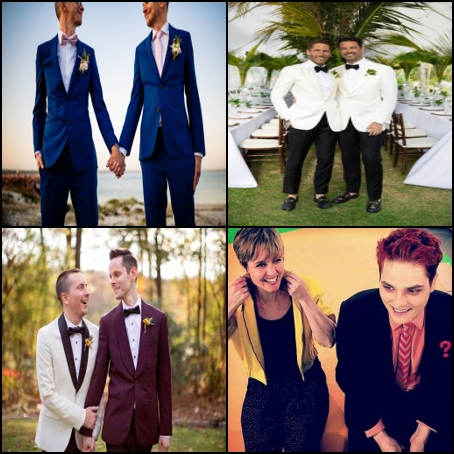} 
& 
\begin{itemize}
	\item mother son (0.2272)
	\item friction between moms and sons (0.1833)
	\item k\censor{ink}i (0.1807)
	\item businesswomen (0.1704)
	\item trans lesbians (0.1539)
\end{itemize}
& 
\begin{itemize}
	\item teen lolita (0.1560)
	\item dads girl (0.1514)
	\item little girl's hair (0.1105)
	\item daughter dad (0.0810)
\end{itemize}
& 
\includegraphics[width=3cm]{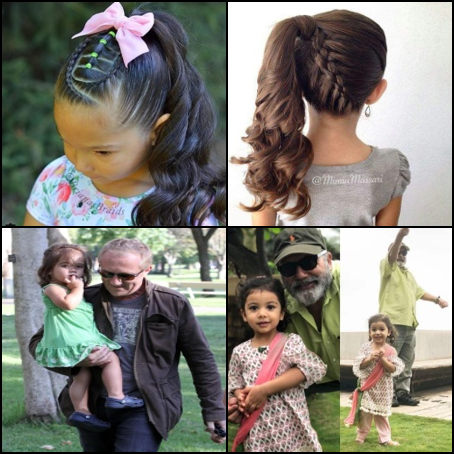} 
\\

\bottomrule
\end{tabular}
\label{tab:supp:vit-large14_l23_h04}
\end{table*}

\begin{table*}
\centering

\caption{\textbf{Layer 23, Head 8 of ViT-L/14 encodes \textit{colors}}. The first 5 pairs of left/right singular vectors from Layer 23, Head 8 of ViT-L/14. For each singular vector, we display the top-4 images from CC12M~\cite{changpinyo2021cc12m} most similar to it, along with the explanation generated by \comp with $\lambda = 0.3$ and $K = 5$. \textcolor{red}{Unsafe text/images have been redacted/blurred.}}

\newcolumntype{I}{ >{\centering\arraybackslash} m{3cm} }

\begin{tabular}{ I m{0.25\textwidth} | m{0.25\textwidth} I }
\toprule

\textbf{Left}: yellowish & \multicolumn{2}{c}{\textbf{1st Singular Vector}} & \textbf{Right}: red \\

\midrule

\includegraphics[width=3cm]{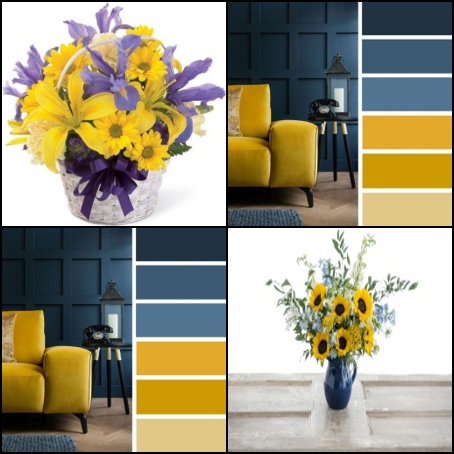} 
&
\begin{itemize}
    \item ground zeroes (0.2402)
    \item yellow green color (0.1565)
    \item blue yellow (0.1186)
    \item bluish yellow (0.1040)
\end{itemize}
& 
\begin{itemize}
    \item pink red (0.3144)
    \item red telephone (0.1571)
    \item red and white factors (0.1503)
    \item scarlet reds (0.1473)
    \item red background (0.1369)
\end{itemize}
&
\includegraphics[width=3cm]{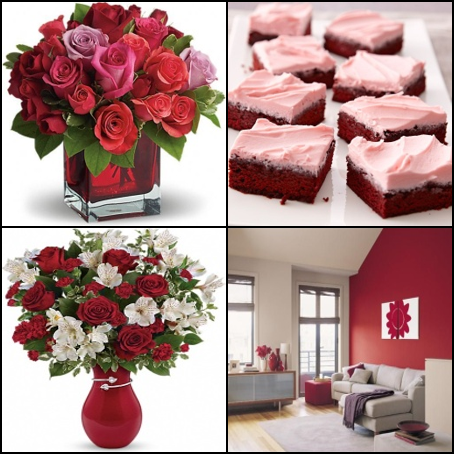} 
\\

\midrule

\textbf{Left}: orange & \multicolumn{2}{c}{\textbf{2nd Singular Vector}} & \textbf{Right}: purple \\

\midrule

\includegraphics[width=3cm]{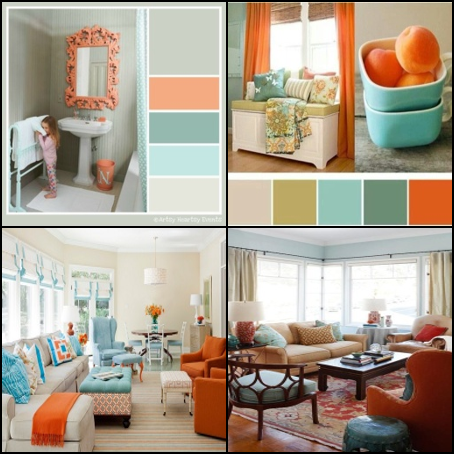} 
& 
\begin{itemize}
    \item orange green (0.2026)
    \item cyanodiethylgold (0.1773)
    \item orange blossoms (0.1121)
    \item peach color (0.1075)
    \item orange mint (0.0234)
\end{itemize}
&
\begin{itemize}
    \item royal purple (0.2482)
    \item purple pills (0.1838)
    \item purple thing (0.1546)
    \item purple states (0.1496)
    \item violet purple (0.1063)
\end{itemize}
& 
\includegraphics[width=3cm]{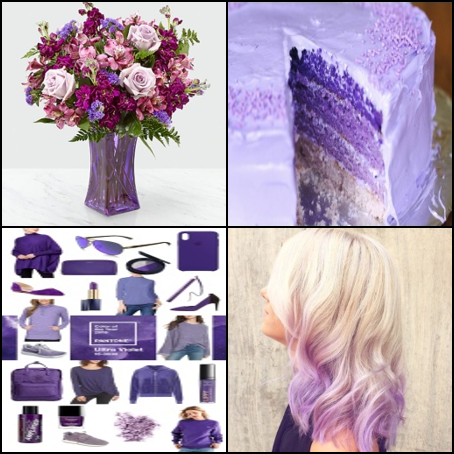} 
\\

\midrule 

\textbf{Left}: yellow & \multicolumn{2}{c}{\textbf{3rd Singular Vector}} & \textbf{Right}: blue \\

\midrule

\includegraphics[width=3cm]{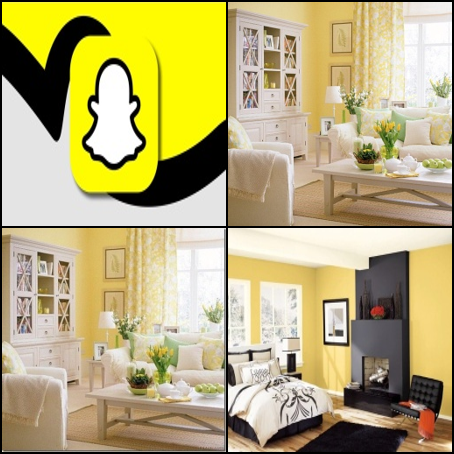} 
& 
\begin{itemize}
    \item yellowred (0.2565)
    \item yellow press (0.1400)
    \item yellow clothes (0.1292)
    \item yellow fever (0.1128)
    \item yellow hot (0.0585)
\end{itemize}
& 
\begin{itemize}
    \item blue pink (0.2813)
    \item blue and white ceramics (0.1795)
    \item prussian blue (0.1376)
    \item blue films (0.1281)
    \item blue hair (0.0649)
\end{itemize}
& 
\includegraphics[width=3cm]{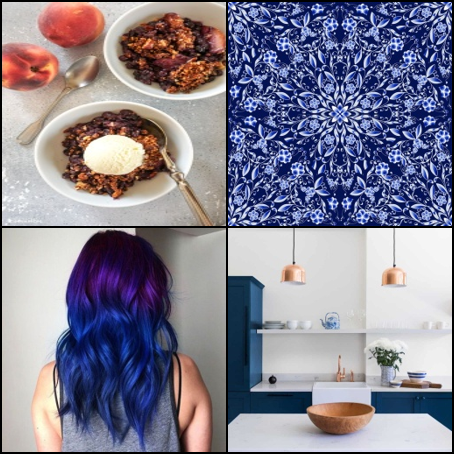} 
\\

\midrule 

\textbf{Left}: light green & \multicolumn{2}{c}{\textbf{4th Singular Vector}} & \textbf{Right}: orange \\

\midrule

\includegraphics[width=3cm]{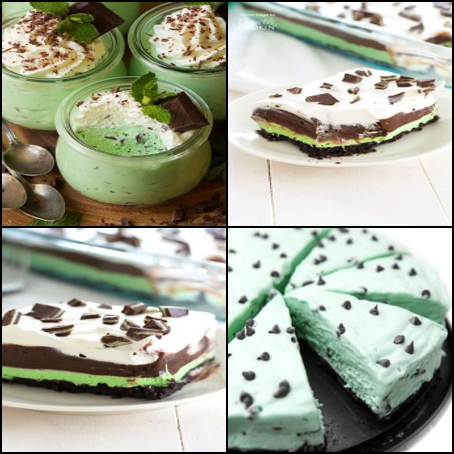} 
& 
\begin{itemize}
    \item white and green (0.2461)
    \item bi asexual (0.1951)
    \item mint green (0.1517)
    \item variscite (0.1491)
    \item mint cream (0.1123)
\end{itemize}
& 
\begin{itemize}
    \item cobalt ocher (0.2203)
    \item orange red colored (0.2014)
    \item padparadscha sapphire (0.1789)
    \item orange revolution (0.1673)
    \item orange yellow (0.0787)
\end{itemize}
&
\includegraphics[width=3cm]{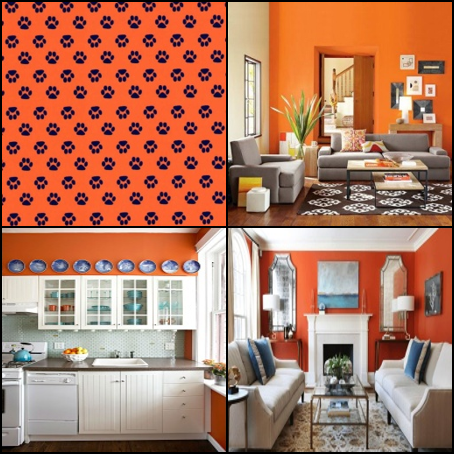} 
\\

\midrule 

\textbf{Left}: brown & \multicolumn{2}{c}{\textbf{5th Singular Vector}} & \textbf{Right}: yellow-blue \\

\midrule

\includegraphics[width=3cm]{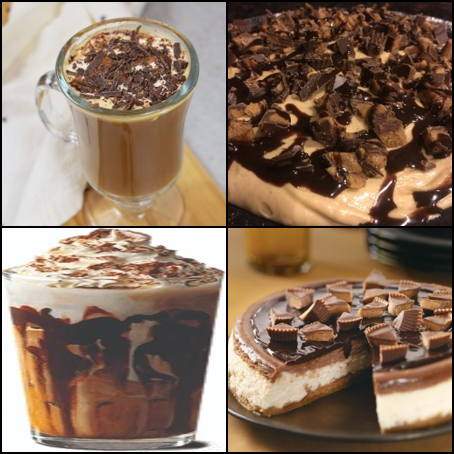} 
& 
\begin{itemize}
    \item copper brown (0.1957)
    \item browns (0.1584)
    \item brown packages (0.1409)
    \item brown shower (0.1259)
    \item brown notes (0.1227)
\end{itemize}
& 
\begin{itemize}
    \item nashville warbler (0.1951)
    \item navy look (0.1743)
    \item colors blue yellow and red (0.1707)
    \item facebook sl\censor{ut}s (0.1658)
    \item blue yellow (0.1068)
\end{itemize}
& 
\includegraphics[width=3cm]{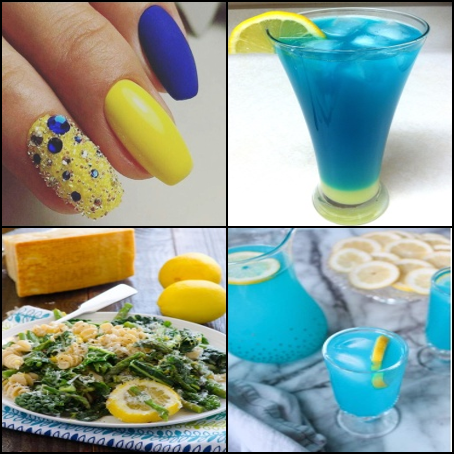} 
\\

\bottomrule
\end{tabular}
\label{tab:supp:vit-large14_l23_h08}
\end{table*}

\begin{table*}
\centering

\caption{\textbf{Layer 23, Head 11 of ViT-L/14 encodes \textit{letters}}. The first 5 pairs of left/right singular vectors from Layer 23, Head 11 of ViT-L/14. For each singular vector, we display the top-4 images from CC12M~\cite{changpinyo2021cc12m} most similar to it, along with the explanation generated by \comp with $\lambda = 0.3$ and $K = 5$. In this case, it is interesting to observe that CLIP is able to read the text present in the watermarks of the images, and the explanations correspond to letters present in those watermarks. \textcolor{red}{Unsafe text/images have been redacted/blurred.}}

\newcolumntype{I}{ >{\centering\arraybackslash} m{3cm} }

\begin{tabular}{ I m{0.25\textwidth} | m{0.25\textwidth} I }
\toprule

\textbf{Left}: the letter C & \multicolumn{2}{c}{\textbf{1st Singular Vector}} & \textbf{Right}: the letter M \\

\midrule

\includegraphics[width=3cm]{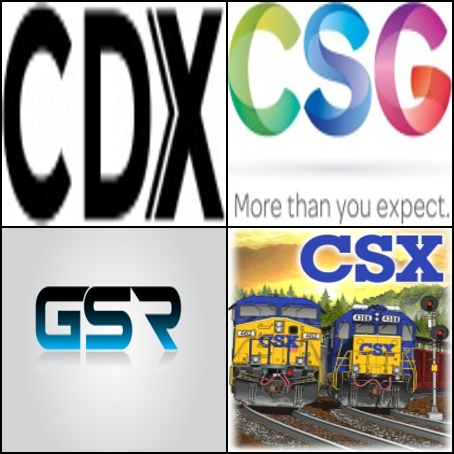} 
&
\begin{itemize}
	\item c\censor{un}t splice (0.1914)
	\item cjd (0.1284)
	\item csdl (0.0907)
	\item cpsa (0.0829)
	\item cpsu (0.0688)
\end{itemize}
& 
\begin{itemize}
	\item mcos (0.0904)
	\item msfc (0.0805)
	\item mvcs (0.0779)
	\item mcls (0.0678)
	\item \textmu cs (0.0656)
\end{itemize}
&
\includegraphics[width=3cm]{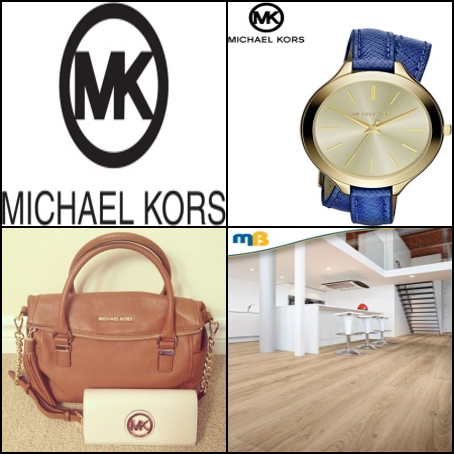} 
\\

\midrule

\textbf{Left}: the letter S & \multicolumn{2}{c}{\textbf{2nd Singular Vector}} & \textbf{Right}: the letter P \\

\midrule

\includegraphics[width=3cm]{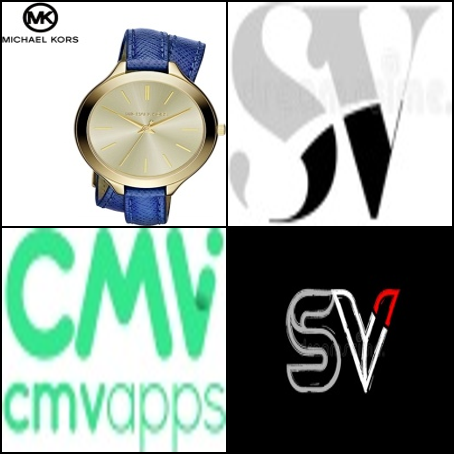} 
& 
\begin{itemize}
	\item surname ma (0.1616)
	\item msw (0.1225)
	\item vsv marv (0.0733)
	\item \textmu sv (0.0496)
\end{itemize}
&
\begin{itemize}
	\item p adic numbers (0.1837)
	\item fims (0.1502)
	\item pcas (0.1478)
	\item pbem (0.1368)
\end{itemize}
& 
\includegraphics[width=3cm]{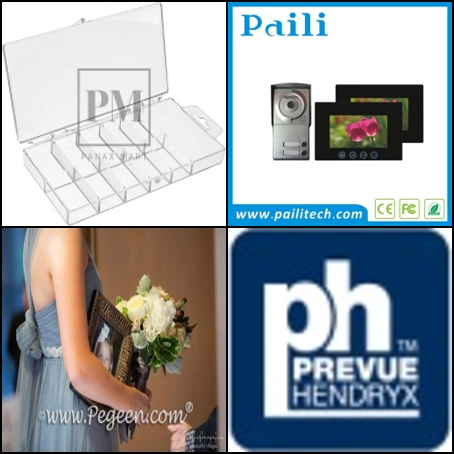} 
\\

\midrule 

\textbf{Left}: the letter D & \multicolumn{2}{c}{\textbf{3rd Singular Vector}} & \textbf{Right}: the letters ISO \\

\midrule

\includegraphics[width=3cm]{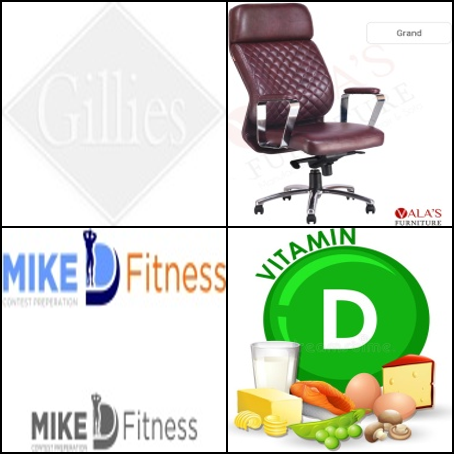} 
& 
\begin{itemize}
	\item dilli bags (0.1720)
	\item ohle (0.1633)
	\item dalk (0.1148)
	\item dillies (0.0817)
	\item dalks (0.0614)
\end{itemize}
& 
\begin{itemize}
	\item isopolitical (0.1671)
	\item isonyms (0.1439)
	\item isoniazid (0.1303)
	\item isonym (0.0390)
	\item isonymic (0.0082)
\end{itemize}
& 
\includegraphics[width=3cm]{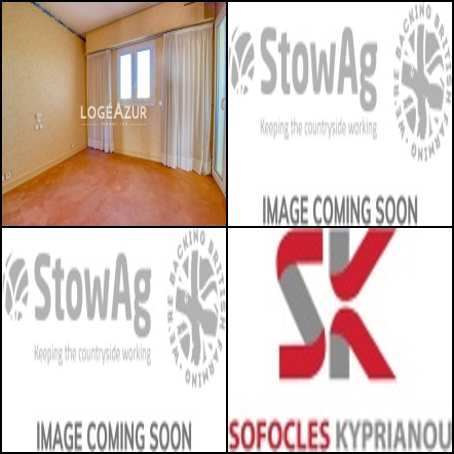} 
\\

\midrule 

\textbf{Left}: the letter L & \multicolumn{2}{c}{\textbf{4th Singular Vector}} & \textbf{Right}: the letter S \\

\midrule

\includegraphics[width=3cm]{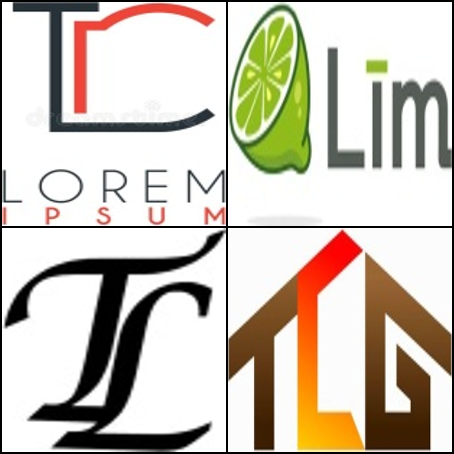} 
& 
\begin{itemize}
	\item linnaean taxonomy (0.2169)
	\item lccns (0.1738)
	\item uaw (0.1724)
	\item limans (0.1192)
	\item lccn (0.0164)
\end{itemize}
& 
\begin{itemize}
	\item sci information (0.1868)
	\item skil (0.1837)
	\item sacrococcygeal fistula (0.1702)
	\item splotchiness (0.1547)
	\item stelliform (0.1277)
\end{itemize}
&
\includegraphics[width=3cm]{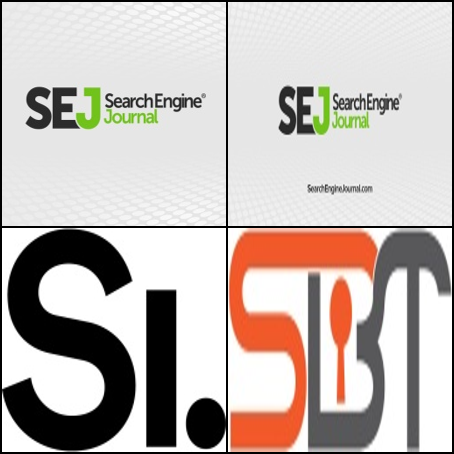} 
\\

\midrule 

\textbf{Left}: the letter F & \multicolumn{2}{c}{\textbf{5th Singular Vector}} & \textbf{Right}: the letter D \\

\midrule

\includegraphics[width=3cm]{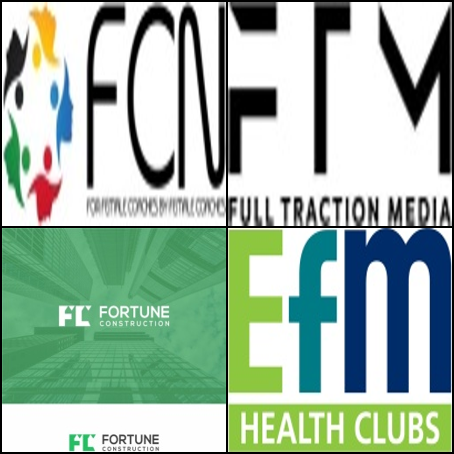} 
& 
\begin{itemize}
	\item c\censor{un}t fart (0.1888)
	\item filial child (0.1805)
	\item fnma (0.1659)
	\item feal (0.1173)
	\item fulah (0.0979)
\end{itemize}
& 
\begin{itemize}
	\item dioxin (0.1743)
	\item disenvelopment (0.1489)
	\item dbcs (0.1391)
	\item dasc (0.1074)
	\item dohcs (0.0598)
\end{itemize}
& 
\includegraphics[width=3cm]{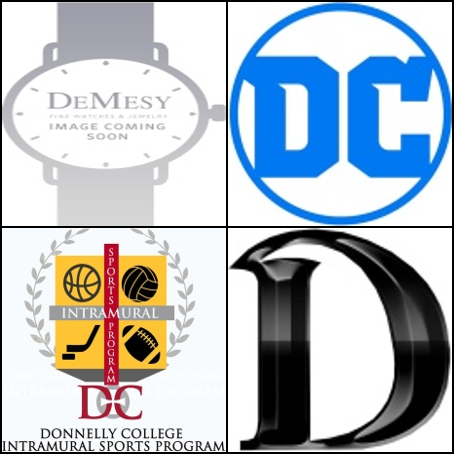} 
\\

\bottomrule
\end{tabular}
\label{tab:supp:vit-large14_l23_h11}
\end{table*}

\begin{table*}
\centering

\caption{\textbf{Layer 11, Head 2 of ViT-B/32 encodes \textit{colors}}. The first 5 pairs of left/right singular vectors from Layer 11, Head 2 of ViT-B/32. For each singular vector, we display the top-4 images from CC12M~\cite{changpinyo2021cc12m} most similar to it, along with the explanation generated by \comp with $\lambda = 0.3$ and $K = 5$.}

\newcolumntype{I}{ >{\centering\arraybackslash} m{3cm} }

\begin{tabular}{ I m{0.25\textwidth} | m{0.25\textwidth} I }
\toprule

\textbf{Left}: shades of white & \multicolumn{2}{c}{\textbf{1st Singular Vector}} & \textbf{Right}: black \\

\midrule

\includegraphics[width=3cm]{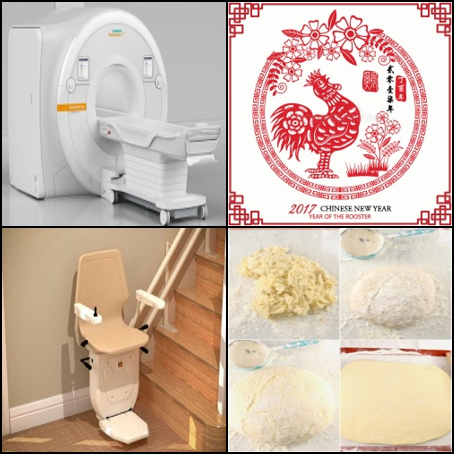} 
&
\begin{itemize}
	\item wear white dresses (0.1973)
	\item white yellow (0.1605)
	\item beige boxes (0.1477)
	\item cream top (0.1409)
	\item cream colored (0.1324)
\end{itemize}
& 
\begin{itemize}
	\item dark kitchens (0.2042)
	\item black top (0.1677)
	\item black on black (0.1609)
	\item people wearing black (0.1290)
	\item black clothes (0.0508)
\end{itemize}
&
\includegraphics[width=3cm]{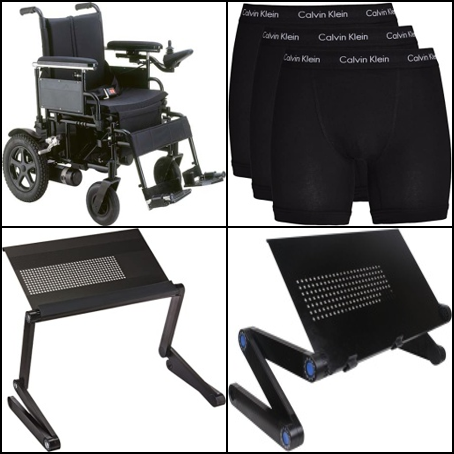} 
\\

\midrule

\textbf{Left}: gold & \multicolumn{2}{c}{\textbf{2nd Singular Vector}} & \textbf{Right}: gray \\

\midrule

\includegraphics[width=3cm]{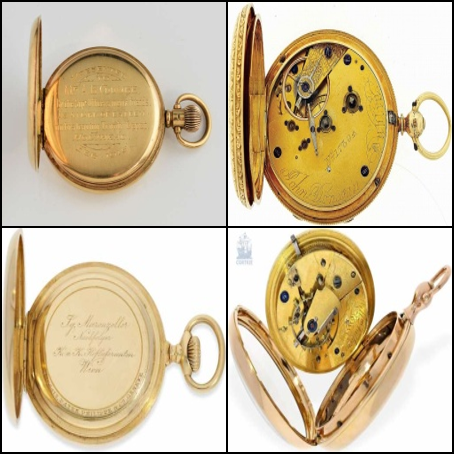} 
& 
\begin{itemize}
	\item reddish gold (0.2863)
	\item brown boxes (0.1661)
	\item golden yellow color (0.1634)
	\item double gloucester (0.1504)
	\item peanut butter (0.1290)
\end{itemize}
&
\begin{itemize}
	\item grey white (0.2089)
	\item silver gray hair (0.1531)
	\item gray blue (0.1469)
	\item grey suits (0.1262)
	\item cool grays (0.0999)
\end{itemize}
& 
\includegraphics[width=3cm]{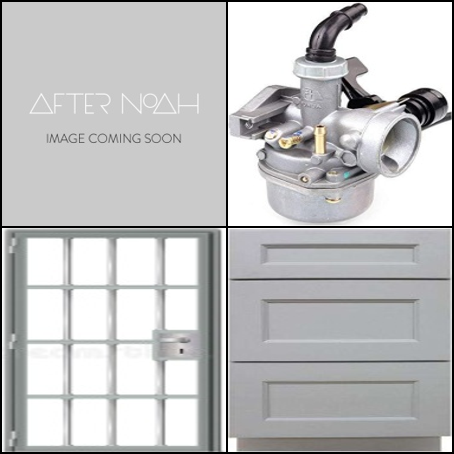} 
\\

\midrule 

\textbf{Left}: brown gray & \multicolumn{2}{c}{\textbf{3rd Singular Vector}} & \textbf{Right}: white \\

\midrule

\includegraphics[width=3cm]{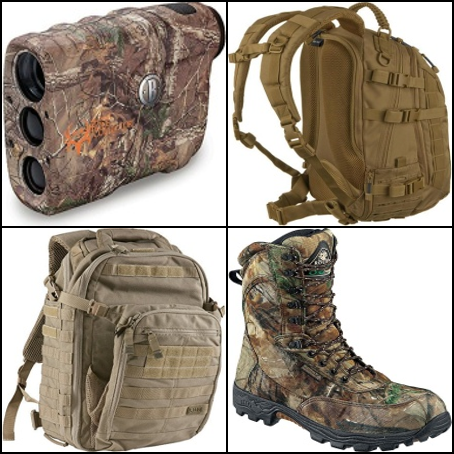} 
& 
\begin{itemize}
	\item yellowish gray (0.2106)
	\item taupe (0.1897)
	\item brown boxes (0.1656)
	\item bronze component (0.1339)
	\item brown gray (0.0899)
\end{itemize}
& 
\begin{itemize}
	\item white black (0.1495)
	\item white shoes (0.1046)
	\item white shirt (0.0934)
	\item white gown (0.0881)
	\item white and black (0.0730)
\end{itemize}
& 
\includegraphics[width=3cm]{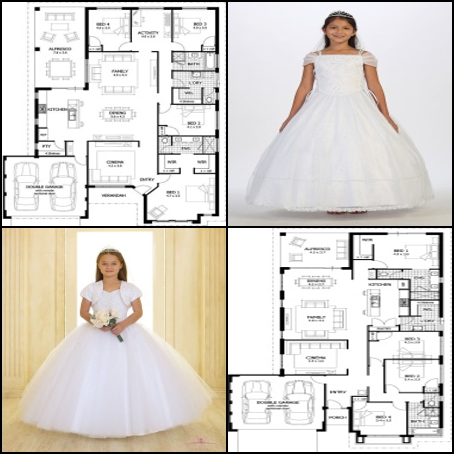} 
\\

\midrule 

\textbf{Left}: black & \multicolumn{2}{c}{\textbf{4th Singular Vector}} & \textbf{Right}: red and blue \\

\midrule

\includegraphics[width=3cm]{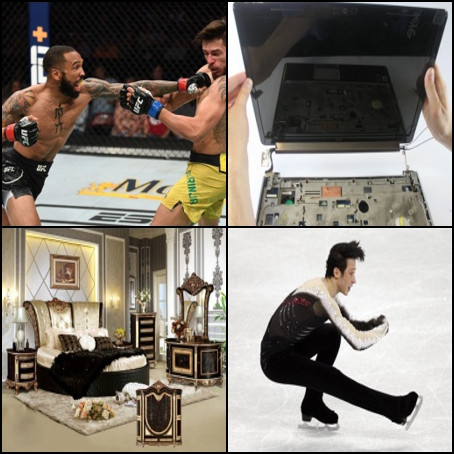} 
& 
\begin{itemize}
	\item ivory black (0.2928)
	\item saints merchandise (0.1523)
	\item ups trucks (0.1417)
	\item black and tan (0.1316)
	\item black gold jewelery (0.1093)
\end{itemize}
& 
\begin{itemize}
	\item red shirts (0.2710)
	\item bluish purple color (0.1572)
	\item blue workpiece (0.1258)
	\item sky blue (0.1258)
	\item azure turquoise (0.1127)
\end{itemize}
&
\includegraphics[width=3cm]{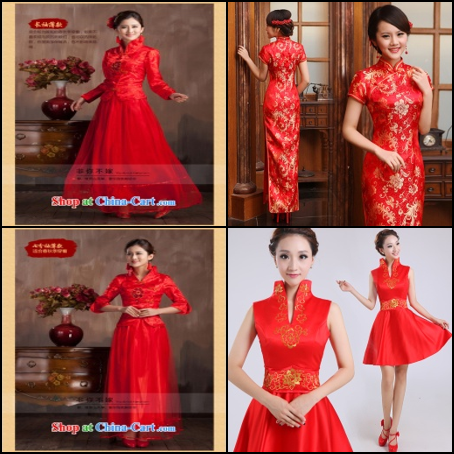} 
\\

\midrule 

\textbf{Left}: light red & \multicolumn{2}{c}{\textbf{5th Singular Vector}} & \textbf{Right}: green (+ blue and yellow) \\

\midrule

\includegraphics[width=3cm]{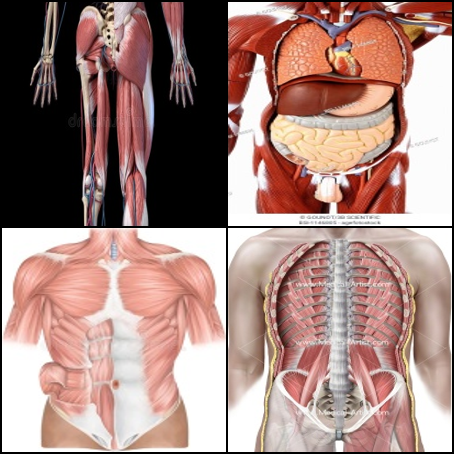} 
& 
\begin{itemize}
	\item pink red (0.1953)
	\item red silver (0.1718)
	\item triple negative breast cancer (0.1641)
	\item hokie (0.1557)
	\item rosé wine (0.1221)
\end{itemize}
& 
\begin{itemize}
	\item blue and yello (0.2161)
	\item light yellowish green (0.2001)
	\item green gowns (0.1615)
	\item plain green plastic (0.1327)
	\item bluish green colour (0.0428)
\end{itemize}
& 
\includegraphics[width=3cm]{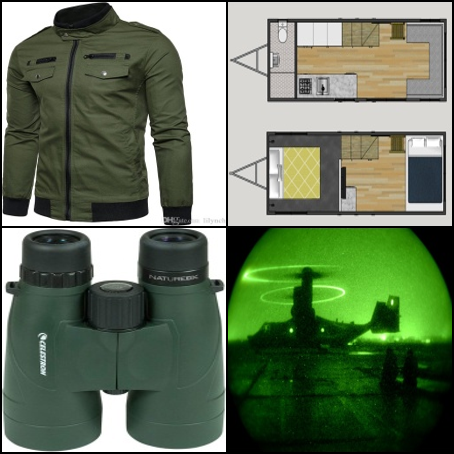} 
\\

\bottomrule
\end{tabular}
\label{tab:supp:vit-base32_l11_h02}
\end{table*}

\begin{table*}
\centering

\caption{\textbf{Layer 31, Head 7 of ViT-H/14 encodes \textit{people}}. The first 5 pairs of left/right singular vectors from Layer 31, Head 7 of ViT-H/14. For each singular vector, we display the top-4 images from CC12M~\cite{changpinyo2021cc12m} most similar to it, along with the explanation generated by \comp with $\lambda = 0.3$ and $K = 5$. \textcolor{red}{Unsafe text/images have been redacted/blurred.}}

\newcolumntype{I}{ >{\centering\arraybackslash} m{3cm} }

\begin{tabular}{ I m{0.25\textwidth} | m{0.25\textwidth} I }
\toprule

\textbf{Left}: men & \multicolumn{2}{c}{\textbf{1st Singular Vector}} & \textbf{Right}: women \\

\midrule

\includegraphics[width=3cm]{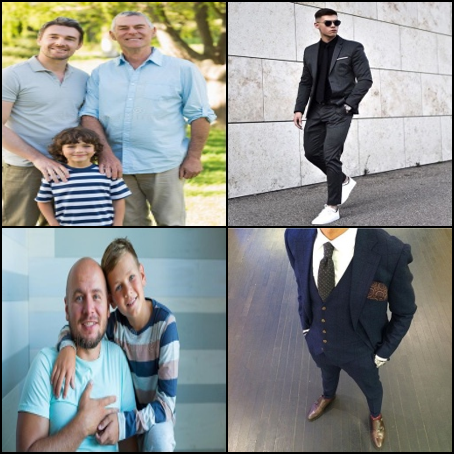} 
&
\begin{itemize}
	\item man p\censor{uss}y (0.1659)
	\item father and son (0.1577)
	\item groomsmen (0.1565)
	\item men's aesthetic (0.1427)
	\item two men holding hands (0.1390)
\end{itemize}
& 
\begin{itemize}
	\item urmila (0.2142)
	\item evil woman (0.2048)
	\item female political activists (0.1991)
	\item scotswomen (0.1749)
	\item women's article (0.1460)
\end{itemize}
&
\includegraphics[width=3cm]{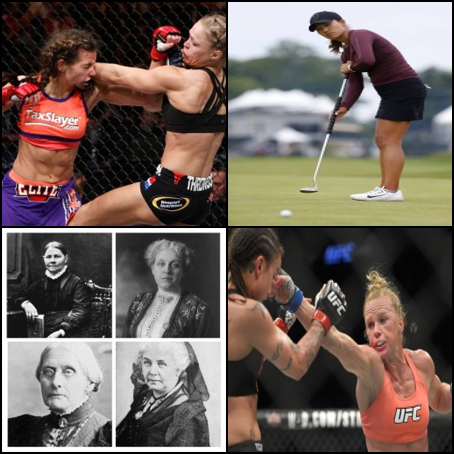} 
\\

\midrule

\textbf{Left}: married couple & \multicolumn{2}{c}{\textbf{2nd Singular Vector}} & \textbf{Right}: boys \\

\midrule

\includegraphics[width=3cm]{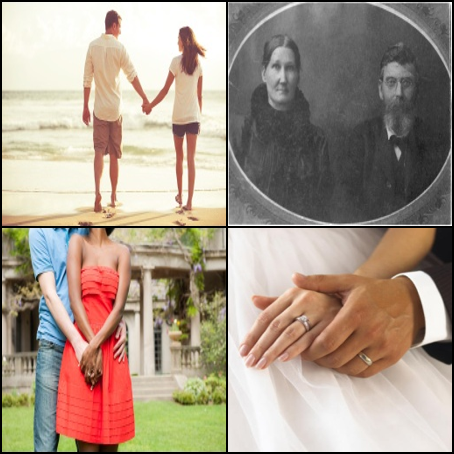} 
& 
\begin{itemize}
	\item married people (0.1279)
	\item relationship through marriage (0.1199)
	\item newlyweds newly married couple (0.1195)
	\item couples together (0.1058)
	\item married couples (0.0999)
\end{itemize}
&
\begin{itemize}
	\item boy monk (0.1720)
	\item making teenaged boys act silly (0.1699)
	\item she male (0.1647)
	\item group boys (0.1644)
	\item topmen (0.1396)
\end{itemize}
& 
\includegraphics[width=3cm]{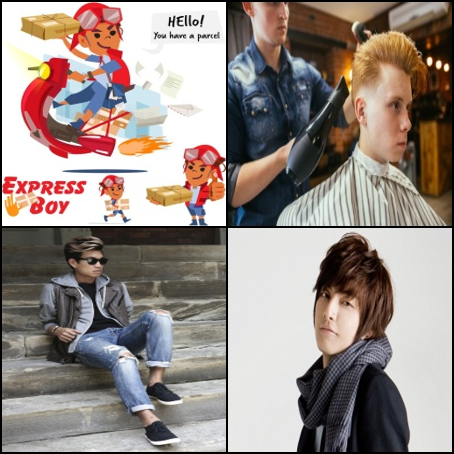} 
\\

\midrule 

\textbf{Left}: bridesmaids & \multicolumn{2}{c}{\textbf{3rd Singular Vector}} & \textbf{Right}: men and women \\

\midrule

\includegraphics[width=3cm]{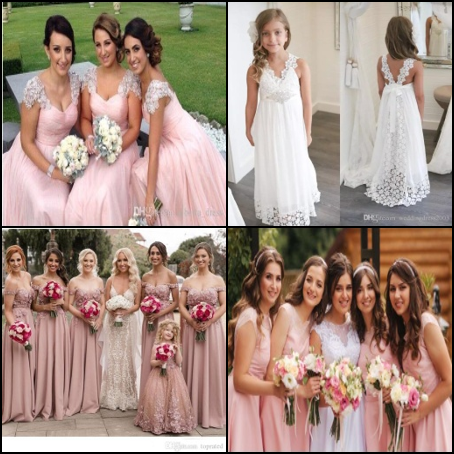} 
& 
\begin{itemize}
	\item junior bridesmaids (0.1675)
	\item pretty dress (0.1286)
\end{itemize}
& 
\begin{itemize}
	\item dad and mom (0.2929)
	\item masculine gender (0.1985)
	\item man who indulges women (0.1791)
	\item man in womans clothes (0.1500)
	\item man male and woman (0.1102)
\end{itemize}
& 
\includegraphics[width=3cm]{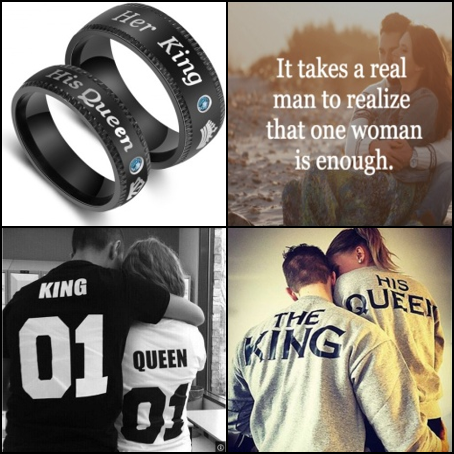} 
\\

\midrule 

\textbf{Left}: mother (and son) & \multicolumn{2}{c}{\textbf{4th Singular Vector}} & \textbf{Right}: girls \\

\midrule

\includegraphics[width=3cm]{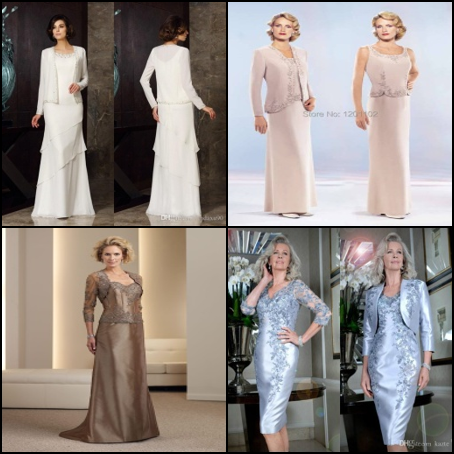} 
& 
\begin{itemize}
	\item mature adult (0.1937)
	\item mother in laws (0.1303)
	\item friction between moms and sons (0.1161)
	\item mother's son (0.1021)
	\item mother son (0.0846)
\end{itemize}
& 
\begin{itemize}
	\item girl priest (0.1601)
	\item girl guides (0.1509)
	\item father daughter (0.1478)
	\item little girl's room (0.1389)
	\item primary schoolgirl (0.0740)
\end{itemize}
&
\includegraphics[width=3cm]{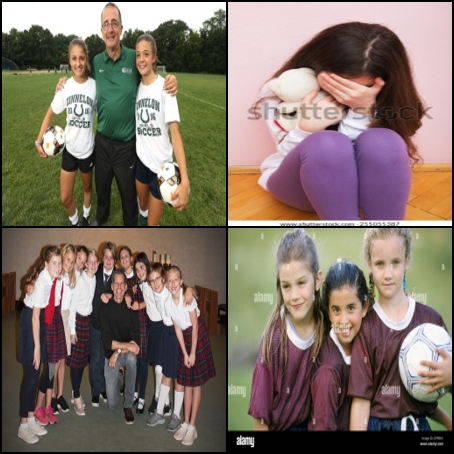} 
\\

\midrule 

\textbf{Left}: couple's wedding & \multicolumn{2}{c}{\textbf{5th Singular Vector}} & \textbf{Right}: person in prominent position \\

\midrule

\includegraphics[width=3cm]{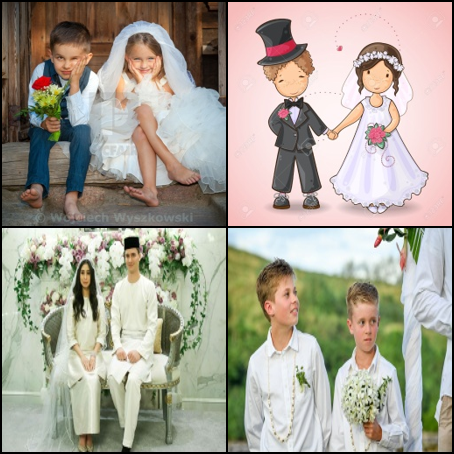} 
& 
\begin{itemize}
	\item ringbearers (0.1753)
	\item wedding boy (0.1617)
	\item wedded (0.1593)
	\item student marriage (0.1592)
	\item teenage couple (0.1399)
\end{itemize}
& 
\begin{itemize}
	\item things to get done faster (0.1477)
	\item retired public prosecutor (0.1372)
	\item coach of dallas cowboys (0.1274)
	\item named under secretary of defense for intelligence (0.0859)
\end{itemize}
& 
\includegraphics[width=3cm]{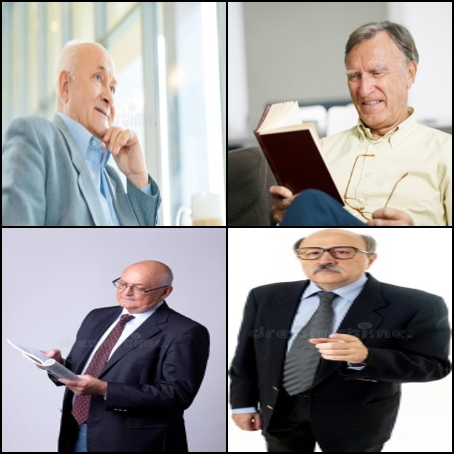} 
\\

\bottomrule
\end{tabular}
\label{tab:supp:vit-huge14_l31_h07}
\end{table*}

\begin{table*}
\centering

\caption{\textbf{Layer 31, Head 11 of ViT-H/14 encodes \textit{letters}}. The first 5 pairs of left/right singular vectors from Layer 31, Head 11 of ViT-H/14. For each singular vector, we display the top-4 images from CC12M~\cite{changpinyo2021cc12m} most similar to it, along with the explanation generated by \comp with $\lambda = 0.3$ and $K = 5$. \textcolor{red}{Unsafe text/images have been redacted/blurred.}}

\newcolumntype{I}{ >{\centering\arraybackslash} m{3cm} }

\begin{tabular}{ I m{0.25\textwidth} | m{0.25\textwidth} I }
\toprule

\textbf{Left}: the letter M & \multicolumn{2}{c}{\textbf{1st Singular Vector}} & \textbf{Right}: the letter S \\

\midrule

\includegraphics[width=3cm]{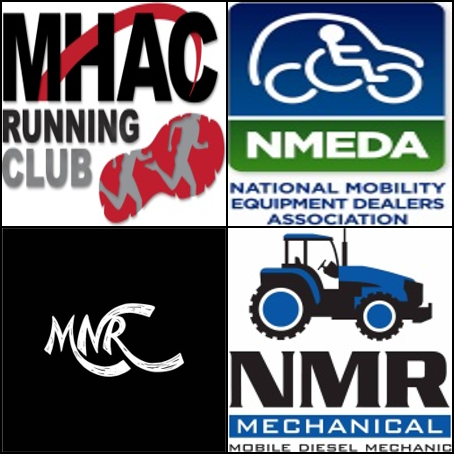} 
&
\begin{itemize}
	\item mln (0.2066)
	\item act reasonable using mind (0.1886)
	\item neuromyoarterial (0.1710)
	\item aaai (0.1417)
	\item haaf net (0.1388)
\end{itemize}
& 
\begin{itemize}
	\item csps (0.2065)
	\item and in cases where letter ss is unavailable (0.2058)
	\item say so (0.1826)
	\item sds (0.1625)
	\item stps (0.1597)
\end{itemize}
&
\includegraphics[width=3cm]{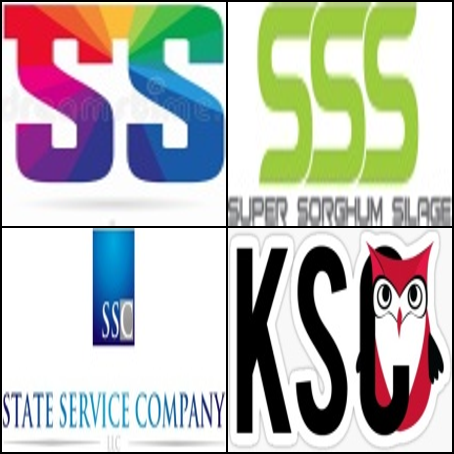} 
\\

\midrule

\textbf{Left}: the letters A, S & \multicolumn{2}{c}{\textbf{2nd Singular Vector}} & \textbf{Right}: the letters F, T, P \\

\midrule

\includegraphics[width=3cm]{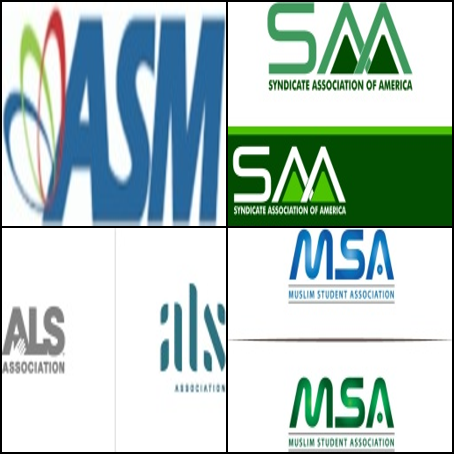} 
& 
\begin{itemize}
	\item genus alosa (0.1889)
	\item ssaa affiliated (0.1730)
	\item now now ism (0.1612)
	\item nsaim (0.1330)
	\item aas (0.1266)
\end{itemize}
&
\begin{itemize}
	\item ftped (0.1914)
	\item eptfe (0.1771)
	\item people first party (0.1760)
	\item tax etc (0.1623)
	\item pte (0.1446)
\end{itemize}
& 
\includegraphics[width=3cm]{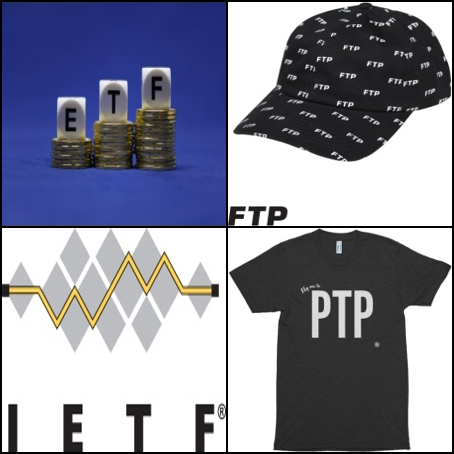} 
\\

\midrule 

\textbf{Left}: the letters T, M & \multicolumn{2}{c}{\textbf{3rd Singular Vector}} & \textbf{Right}: the letters C, D, E \\

\midrule

\includegraphics[width=3cm]{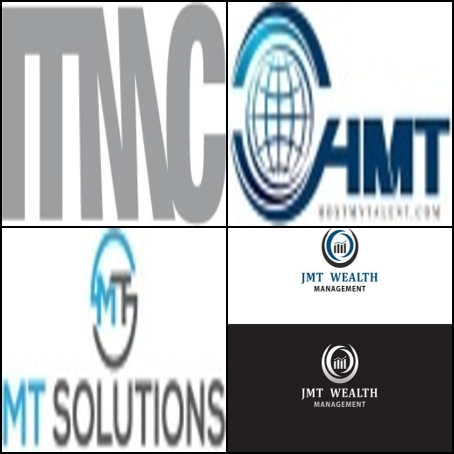} 
& 
\begin{itemize}
	\item syrt (0.1828)
	\item t\censor{itt}y f\censor{uckin}g (0.1797)
	\item jvm ti (0.1779)
	\item wry mouth (0.1686)
	\item ntim (0.1576)
\end{itemize}
& 
\begin{itemize}
	\item d in p aeq (0.2637)
	\item dead ice (0.2117)
	\item eogs (0.2032)
	\item cded (0.1722)
	\item egd (0.0813)
\end{itemize}
& 
\includegraphics[width=3cm]{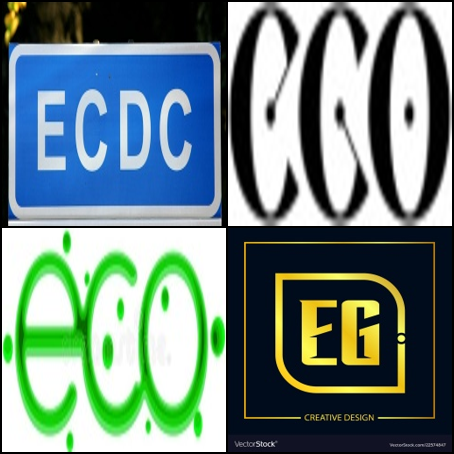} 
\\

\midrule 

\textbf{Left}: the letter A & \multicolumn{2}{c}{\textbf{4th Singular Vector}} & \textbf{Right}: the letter C \\

\midrule

\includegraphics[width=3cm]{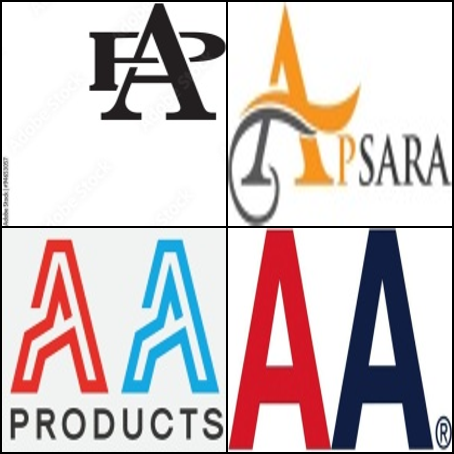} 
& 
\begin{itemize}
	\item pfaas (0.2619)
	\item keep as pet (0.2127)
	\item air to air (0.1867)
	\item aa trees (0.1755)
	\item aas (0.1734)
\end{itemize}
& 
\begin{itemize}
	\item ecec (0.1284)
	\item idmc (0.1282)
	\item ixc (0.1271)
	\item ncdc (0.0874)
	\item ndcc (0.0282)
\end{itemize}
&
\includegraphics[width=3cm]{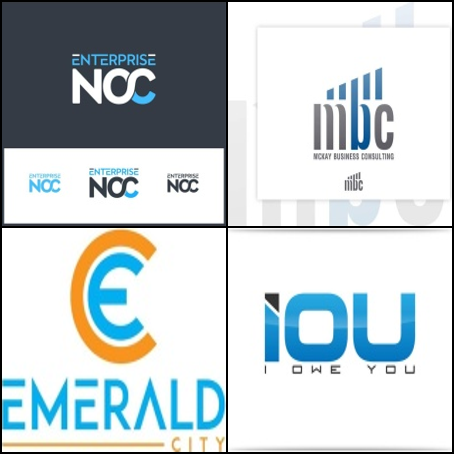} 
\\

\midrule 

\textbf{Left}: the letter D & \multicolumn{2}{c}{\textbf{5th Singular Vector}} & \textbf{Right}: the letter E \\

\midrule

\includegraphics[width=3cm]{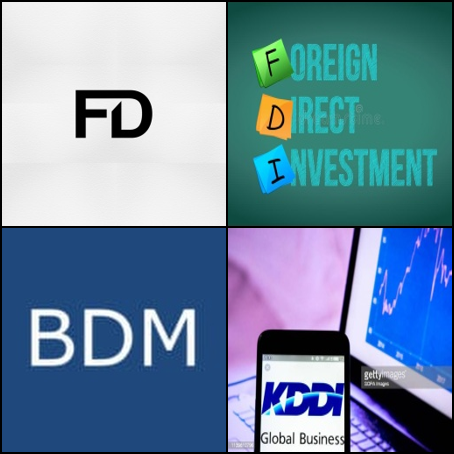} 
& 
\begin{itemize}
	\item baby dog (0.2224)
	\item d'you (0.1796)
	\item fdi (0.1735)
	\item fpd (0.1553)
	\item fddi (0.1033)
\end{itemize}
& 
\begin{itemize}
	\item vse (0.1378)
	\item ees (0.1272)
	\item cses (0.1053)
	\item jses (0.0756)
	\item eses (0.0738)
\end{itemize}
& 
\includegraphics[width=3cm]{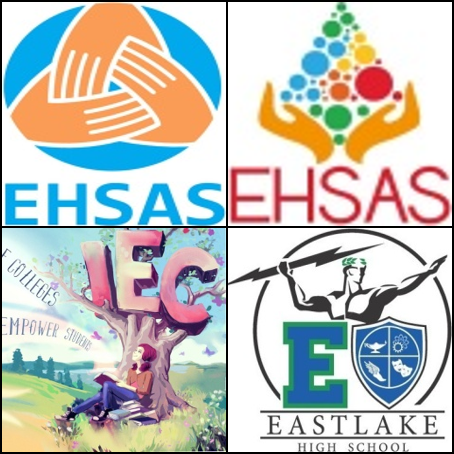} 
\\

\bottomrule
\end{tabular}
\label{tab:supp:vit-huge14_l31_h11}
\end{table*}

\begin{table*}
\centering

\caption{\textbf{Layer 31, Head 12 of ViT-H/14 encodes \textit{locations}}. The first 5 pairs of left/right singular vectors from Layer 31, Head 12 of ViT-H/14. For each singular vector, we display the top-4 images from CC12M~\cite{changpinyo2021cc12m} most similar to it, along with the explanation generated by \comp with $\lambda = 0.3$ and $K = 5$. \textcolor{red}{Unsafe text/images have been redacted/blurred.}}

\newcolumntype{I}{ >{\centering\arraybackslash} m{3cm} }

\begin{tabular}{ I m{0.25\textwidth} | m{0.25\textwidth} I }
\toprule

\textbf{Left}: festivals & \multicolumn{2}{c}{\textbf{1st Singular Vector}} & \textbf{Right}: stores \\

\midrule

\includegraphics[width=3cm]{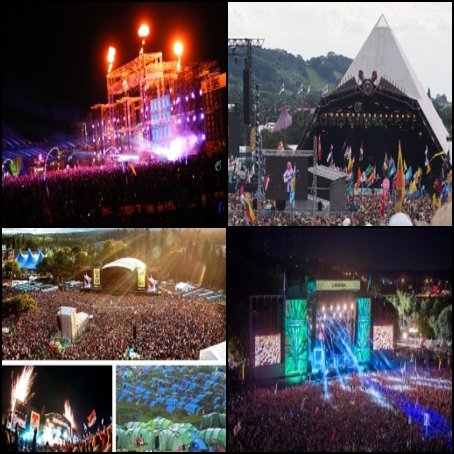} 
&
\begin{itemize}
	\item megafestivals (0.1719)
	\item bedside book (0.1338)
	\item fix piece of wood furniture (0.1328)
	\item made from wood cotton or linen (0.1222)
	\item handy for mending wood furniture (0.0038)
\end{itemize}
& 
\begin{itemize}
	\item museum building (0.2123)
	\item in laundry store (0.1761)
	\item airport lounges (0.1570)
	\item information store p\censor{ornographi}c (0.1409)
	\item retail store (0.1208)
\end{itemize}
&
\includegraphics[width=3cm]{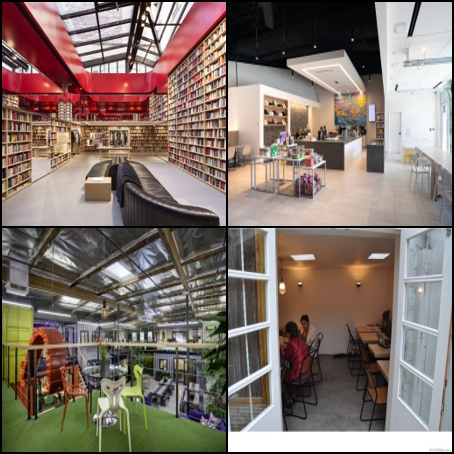} 
\\

\midrule

\textbf{Left}: home interior & \multicolumn{2}{c}{\textbf{2nd Singular Vector}} & \textbf{Right}: public spaces \\

\midrule

\includegraphics[width=3cm]{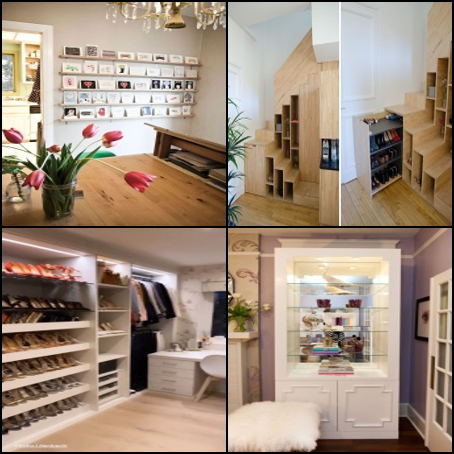} 
& 
\begin{itemize}
	\item nail wood (0.1616)
	\item front parlor (0.1585)
	\item store things on bookshelf (0.1583)
	\item interior decorating (0.1169)
	\item bedroom storage (0.0656)
\end{itemize}
&
\begin{itemize}
	\item outside hospital (0.1990)
	\item campus festival (0.1683)
	\item courtyard (0.1605)
	\item industrial parks (0.1573)
	\item convention center (0.1518)
\end{itemize}
& 
\includegraphics[width=3cm]{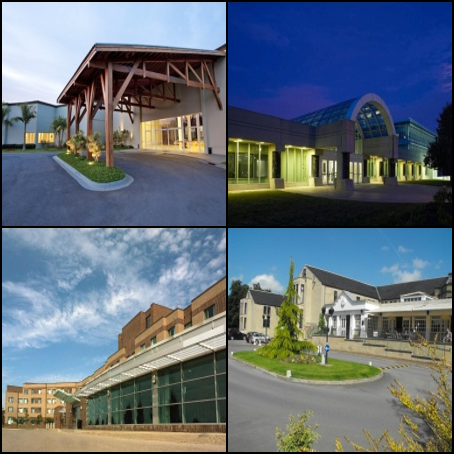} 
\\

\midrule 

\textbf{Left}: training class & \multicolumn{2}{c}{\textbf{3rd Singular Vector}} & \textbf{Right}: fairs \\

\midrule

\includegraphics[width=3cm]{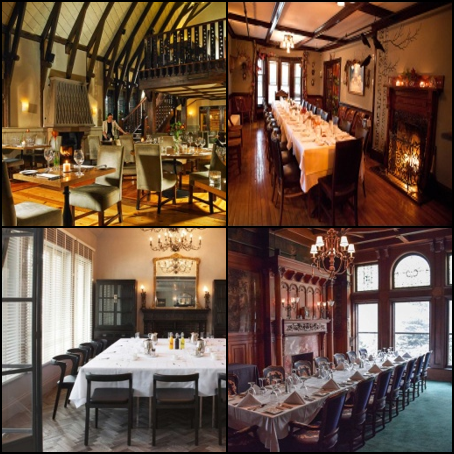} 
& 
\begin{itemize}
	\item learning about cartooning (0.1569)
	\item corporate in service training (0.1345)
	\item training class (0.0892)
	\item employee training (0.0141)
	\item teacher training (0.0055)
\end{itemize}
& 
\begin{itemize}
	\item make bathroom walls (0.2191)
	\item at fair (0.1881)
	\item bus depot booth (0.1594)
	\item comiket (0.1589)
	\item art fair (0.1474)
\end{itemize}
& 
\includegraphics[width=3cm]{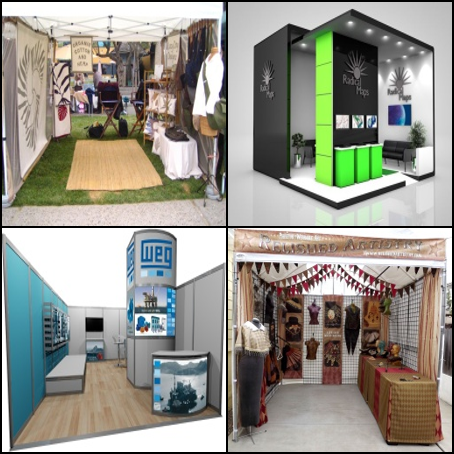} 
\\

\midrule 

\textbf{Left}: hall & \multicolumn{2}{c}{\textbf{4th Singular Vector}} & \textbf{Right}: outdoor \\

\midrule

\includegraphics[width=3cm]{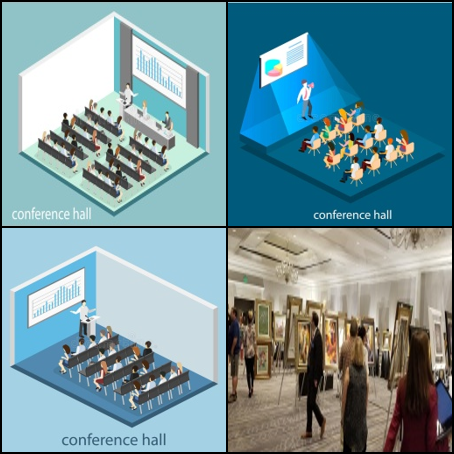} 
& 
\begin{itemize}
	\item in hall (0.1784)
	\item courtroom drama (0.1653)
	\item athletic event played indoors (0.1570)
	\item use in living room (0.1567)
	\item exhibition room (0.1486)
\end{itemize}
& 
\begin{itemize}
	\item having s\censor{ex} outdoors (0.2018)
	\item vacant lots (0.1821)
	\item outdinning (0.1154)
	\item outdoor gambling (0.1076)
	\item pub garden (0.0931)
\end{itemize}
&
\includegraphics[width=3cm]{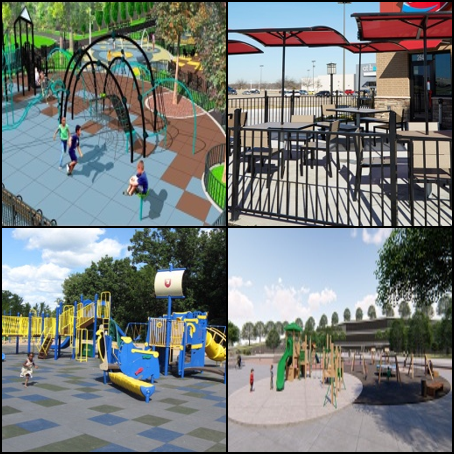} 
\\

\midrule 

\textbf{Left}: storehouse & \multicolumn{2}{c}{\textbf{5th Singular Vector}} & \textbf{Right}: screening, restaurant\\

\midrule

\includegraphics[width=3cm]{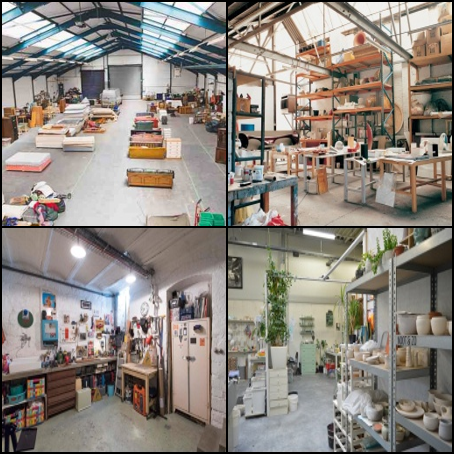} 
& 
\begin{itemize}
	\item workshed (0.1505)
	\item photography studio (0.1216)
	\item warehouse goods (0.1153)
	\item storage unit (0.1008)
	\item storing things in garage (0.0903)
\end{itemize}
& 
\begin{itemize}
	\item public screening (0.1881)
	\item restaurant hotel (0.1467)
	\item putting up at hotel (0.1408)
	\item intercontinental (0.1397)
	\item performed in restaurants (0.1302)
\end{itemize}
& 
\includegraphics[width=3cm]{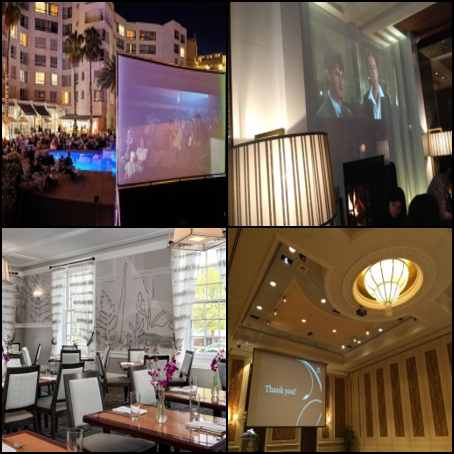} 
\\

\bottomrule
\end{tabular}
\label{tab:supp:vit-huge14_l31_h12}
\end{table*}

\begin{table*}
\centering

\caption{\textbf{Layer 31, Head 13 of ViT-H/14 encodes \textit{colors}}. The first 5 pairs of left/right singular vectors from Layer 31, Head 13 of ViT-H/14. For each singular vector, we display the top-4 images from CC12M~\cite{changpinyo2021cc12m} most similar to it, along with the explanation generated by \comp with $\lambda = 0.3$ and $K = 5$.}

\newcolumntype{I}{ >{\centering\arraybackslash} m{3cm} }

\begin{tabular}{ I m{0.25\textwidth} | m{0.25\textwidth} I }
\toprule

\textbf{Left}: green, brown & \multicolumn{2}{c}{\textbf{1st Singular Vector}} & \textbf{Right}: indigo purple \\

\midrule

\includegraphics[width=3cm]{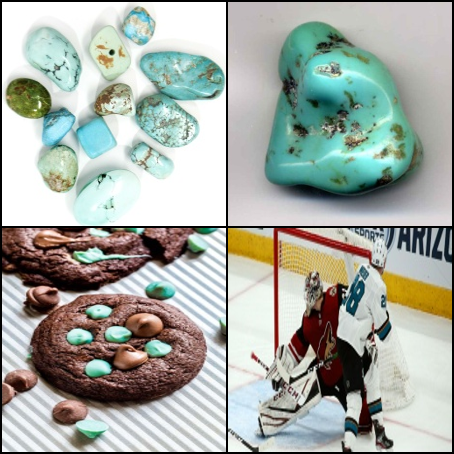} 
&
\begin{itemize}
	\item deadspace (0.1830)
	\item mint green (0.1571)
	\item brown hooded parrot (0.1319)
	\item brown teal (0.1070)
	\item mint chocolate chip (0.1044)
\end{itemize}
& 
\begin{itemize}
	\item navy thing (0.2547)
	\item facebooks (0.2191)
	\item sodalite (0.1731)
	\item indigo paper (0.1728)
	\item navy look (0.1383)
\end{itemize}
&
\includegraphics[width=3cm]{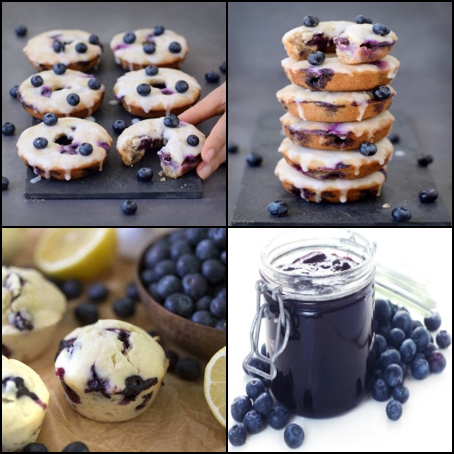} 
\\

\midrule

\textbf{Left}: red and blue & \multicolumn{2}{c}{\textbf{2nd Singular Vector}} & \textbf{Right}: black purple \\

\midrule

\includegraphics[width=3cm]{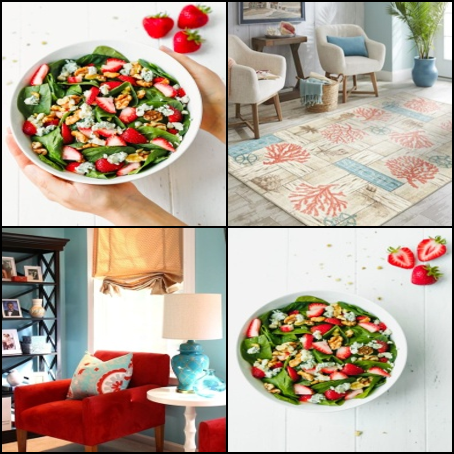} 
& 
\begin{itemize}
	\item cyanotype (0.1935)
	\item red shouldered macaw (0.1549)
	\item korean air (0.1544)
	\item teal (0.1537)
	\item turquoise thing (0.0407)
\end{itemize}
&
\begin{itemize}
	\item black purple (0.2681)
	\item purple black (0.2144)
	\item amethysts (0.1607)
	\item asexual (0.1566)
	\item purples (0.0416)
\end{itemize}
& 
\includegraphics[width=3cm]{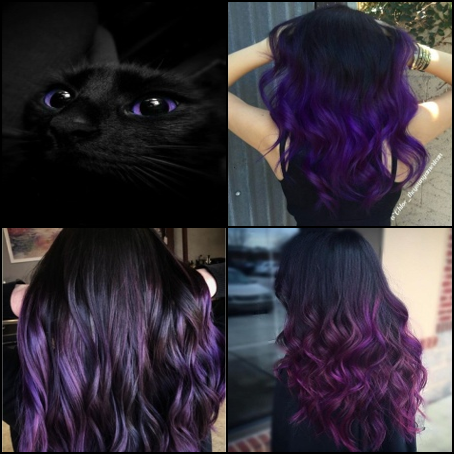} 
\\

\midrule 

\textbf{Left}: red, black, blue & \multicolumn{2}{c}{\textbf{3rd Singular Vector}} & \textbf{Right}: green, purple \\

\midrule

\includegraphics[width=3cm]{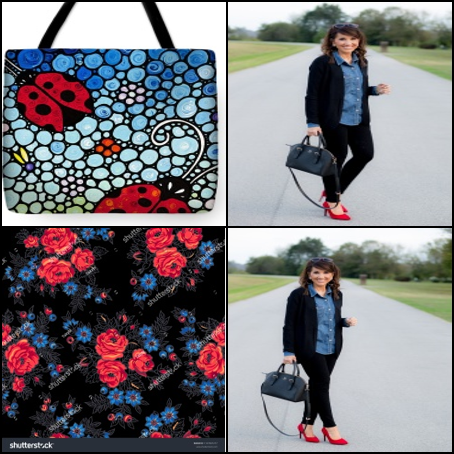} 
& 
\begin{itemize}
	\item red and black (0.3333)
	\item blue coal (0.1386)
	\item blue blacks (0.1367)
	\item blue red (0.1010)
	\item black and blue (0.0736)
\end{itemize}
& 
\begin{itemize}
	\item lakers and celtics (0.1884)
	\item rhododendron viscosum (0.1854)
	\item green purple (0.1783)
	\item purplish green (0.1078)
\end{itemize}
& 
\includegraphics[width=3cm]{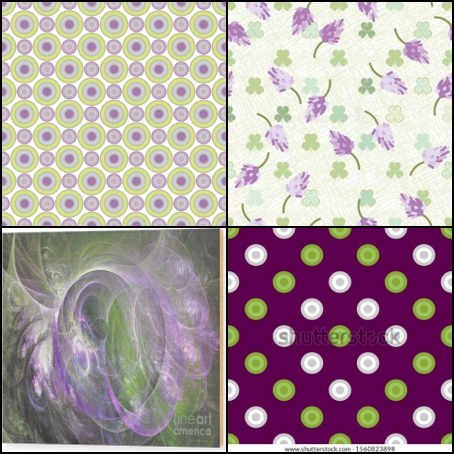} 
\\

\midrule 

\textbf{Left}: green red & \multicolumn{2}{c}{\textbf{4th Singular Vector}} & \textbf{Right}: light blue \\

\midrule

\includegraphics[width=3cm]{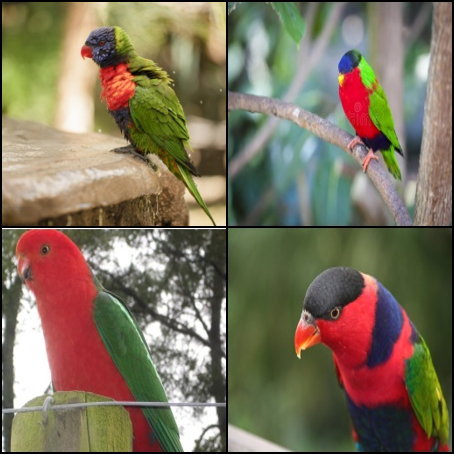} 
& 
\begin{itemize}
	\item varied lorikeet (0.2447)
	\item green red (0.2136)
	\item red green alliance (0.1543)
	\item red and green and ripe (0.1291)
	\item red and green (0.0304)
\end{itemize}
& 
\begin{itemize}
	\item blue and white (0.1893)
	\item celestites (0.1796)
	\item indigo brown (0.1794)
	\item blue lights (0.0903)
	\item baby blue (0.0880)
\end{itemize}
&
\includegraphics[width=3cm]{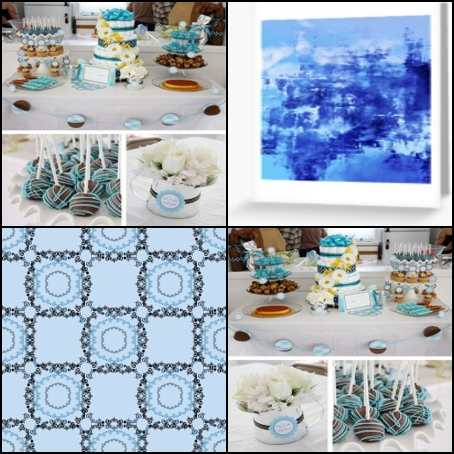} 
\\

\midrule 

\textbf{Left}: blue purple & \multicolumn{2}{c}{\textbf{5th Singular Vector}} & \textbf{Right}: black \\

\midrule

\includegraphics[width=3cm]{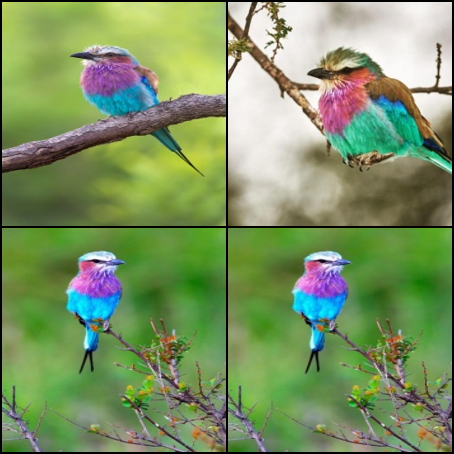} 
& 
\begin{itemize}
	\item blue purple (0.3025)
	\item european roller (0.1686)
	\item bright and multicolored (0.1681)
	\item primary rainbow (0.1307)
	\item ceratostigma (0.1263)
\end{itemize}
& 
\begin{itemize}
	\item black and whites (0.2116)
	\item grossularite (0.1873)
	\item red and black (0.1864)
	\item black diamonds (0.1561)
	\item sunless (0.1520)
\end{itemize}
& 
\includegraphics[width=3cm]{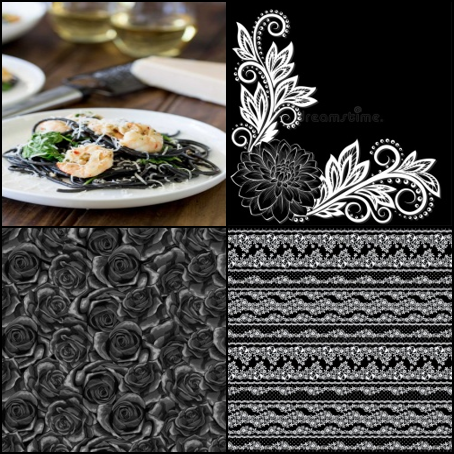} 
\\

\bottomrule
\end{tabular}
\label{tab:supp:vit-huge14_l31_h13}
\end{table*}

\begin{table*}
\centering

\caption{\textbf{Layer 22, Head 0 of MobileCLIP-L/14 encodes \textit{numbers}}. The first 5 pairs of left/right singular vectors from Layer 22, Head 0 of MobileCLIP-L/14. For each singular vector, we display the top-4 images from CC12M~\cite{changpinyo2021cc12m} most similar to it, along with the explanation generated by \comp with $\lambda = 0.3$ and $K = 5$.}

\newcolumntype{I}{ >{\centering\arraybackslash} m{3cm} }

\begin{tabular}{ I m{0.25\textwidth} | m{0.25\textwidth} I }
\toprule

\textbf{Left}: 4 & \multicolumn{2}{c}{\textbf{1st Singular Vector}} & \textbf{Right}: 8 \\

\midrule

\includegraphics[width=3cm]{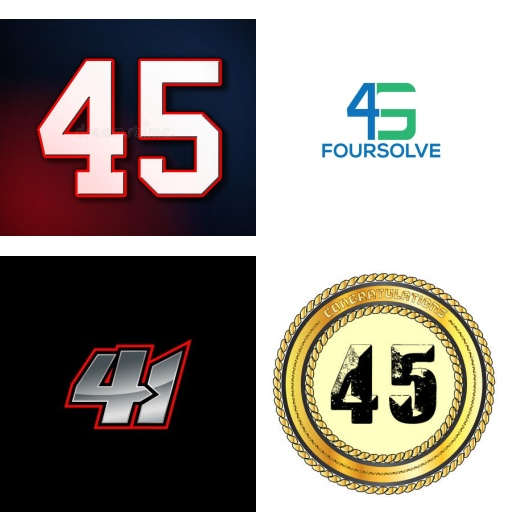} 
&
\begin{itemize}
	\item 45 (0.3989)
	\item four lags (0.1110)
	\item four books (0.1019)
	\item four stroking (0.0958)
	\item four flushes (0.0770)
\end{itemize}
& 
\begin{itemize}
	\item nine parts (0.1750)
	\item 78 (0.1652)
	\item group of eight (0.1271)
	\item 8º (0.0924)
\end{itemize}
&
\includegraphics[width=3cm]{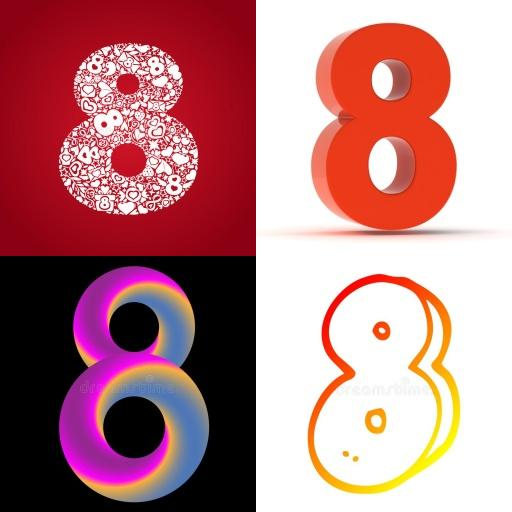} 
\\

\midrule

\textbf{Left}: 6 & \multicolumn{2}{c}{\textbf{2nd Singular Vector}} & \textbf{Right}: 4, 3 \\

\midrule

\includegraphics[width=3cm]{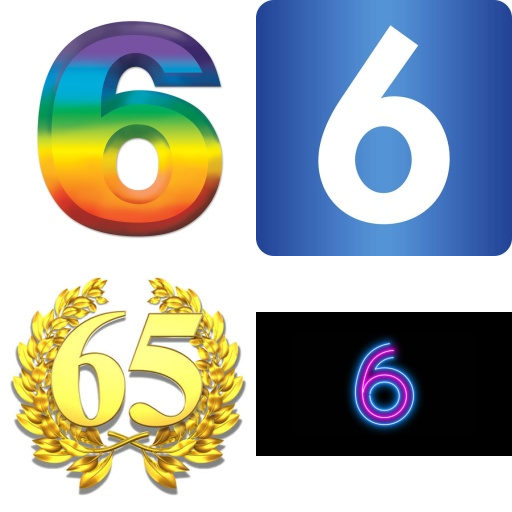} 
& 
\begin{itemize}
	\item 56 (0.2728)
	\item six ksitigarbhas (0.1804)
	\item six copies (0.1371)
	\item six footedness (0.1312)
	\item five six (0.0541)
\end{itemize}
&
\begin{itemize}
	\item rounded to 3.14 (0.1518)
	\item 14er (0.1353)
	\item 3 4 (0.1233)
	\item fourteen (0.0996)
\end{itemize}
& 
\includegraphics[width=3cm]{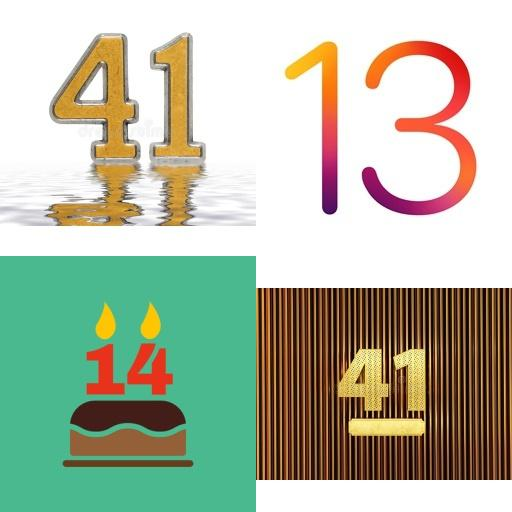} 
\\

\midrule 

\textbf{Left}: 4, 6, 8 & \multicolumn{2}{c}{\textbf{3rd Singular Vector}} & \textbf{Right}: 5, 3, 1 \\

\midrule

\includegraphics[width=3cm]{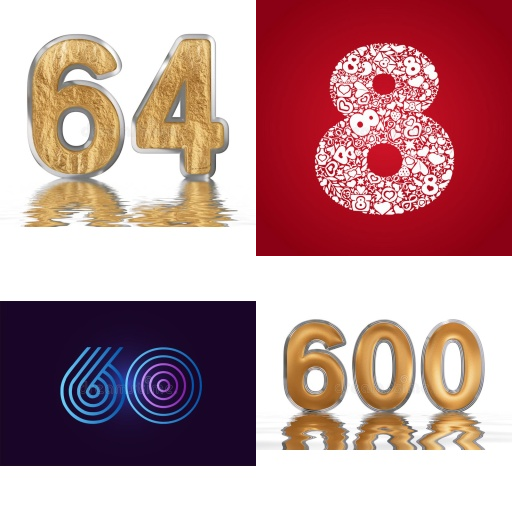} 
& 
\begin{itemize}
	\item 6in4 (0.1865)
	\item 84000 (0.1207)
	\item 4 8 0 (0.0710)
	\item 4 8 6 (0.0670)
	\item 486 (0.0646)
\end{itemize}
& 
\begin{itemize}
	\item 51 percent (0.1256)
	\item 357 (0.1095)
	\item 3 1 1 (0.0960)
	\item 5 1 (0.0579)
	\item 51 (0.0515)
\end{itemize}
& 
\includegraphics[width=3cm]{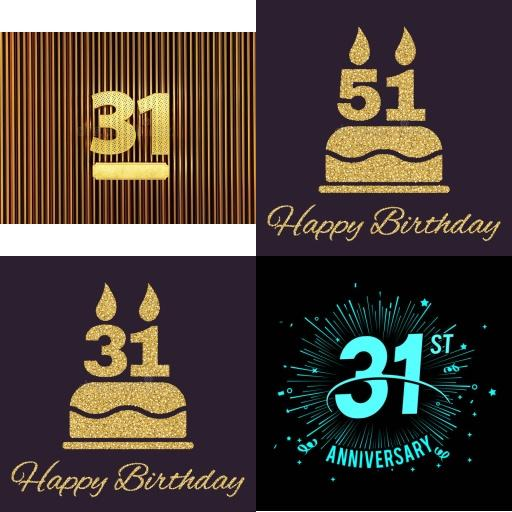} 
\\

\midrule 

\textbf{Left}: 7, 9 & \multicolumn{2}{c}{\textbf{4th Singular Vector}} & \textbf{Right}: 12, 16 \\

\midrule

\includegraphics[width=3cm]{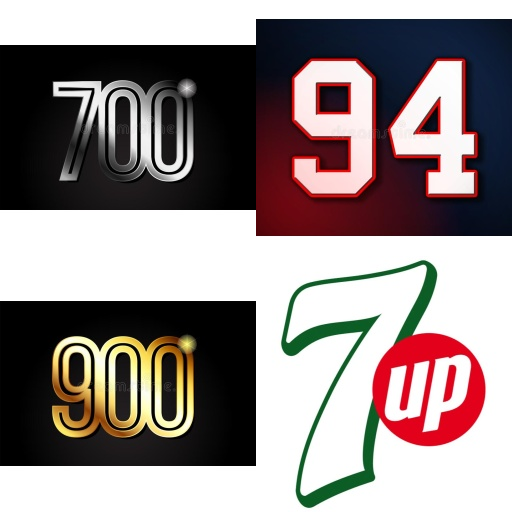} 
& 
\begin{itemize}
	\item 49th (0.1119)
	\item 47 (0.0907)
	\item 997 (0.0853)
	\item 87 (0.0554)
\end{itemize}
& 
\begin{itemize}
	\item 120 (0.2458)
	\item sixteen arhats (0.2163)
	\item occur to (0.1590)
	\item sixteen ounces (0.1256)
	\item is to be (0.1138)
\end{itemize}
&
\includegraphics[width=3cm]{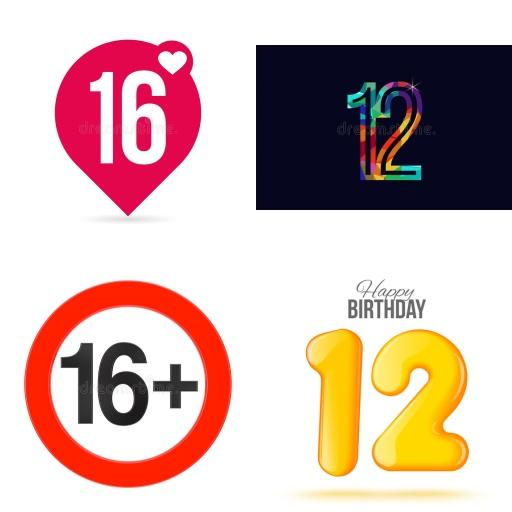} 
\\

\midrule 

\textbf{Left}: 1 & \multicolumn{2}{c}{\textbf{5th Singular Vector}} & \textbf{Right}: 32 \\

\midrule

\includegraphics[width=3cm]{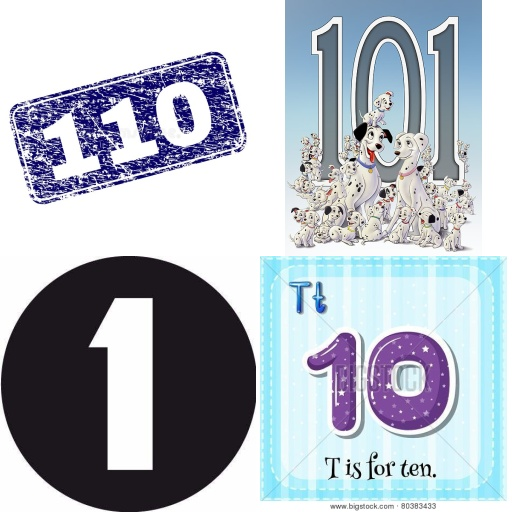} 
& 
\begin{itemize}
	\item 110 proof (0.1539)
	\item elevenths (0.1498)
	\item 11 (0.1281)
	\item 101st (0.0591)
	\item 110th (0.0572)
\end{itemize}
& 
\begin{itemize}
	\item 23 (0.2565)
	\item 32 bit (0.1439)
	\item 32s (0.0870)
	\item 23rd (0.0780)
\end{itemize}
& 
\includegraphics[width=3cm]{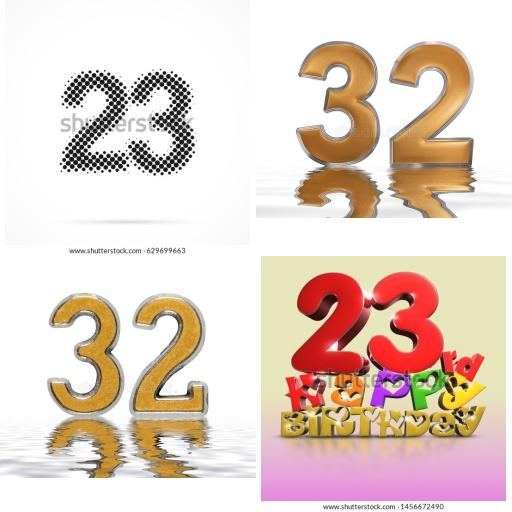} 
\\

\bottomrule
\end{tabular}
\label{tab:supp:mobile-large14_l22_h00}
\end{table*}

\begin{table*}
\centering

\caption{\textbf{Layer 22, Head 10 of MobileCLIP-L/14 encodes \textit{locations}}. The first 5 pairs of left/right singular vectors from Layer 22, Head 10 of MobileCLIP-L/14. For each singular vector, we display the top-4 images from CC12M~\cite{changpinyo2021cc12m} most similar to it, along with the explanation generated by \comp with $\lambda = 0.3$ and $K = 5$.}

\newcolumntype{I}{ >{\centering\arraybackslash} m{3cm} }

\begin{tabular}{ I m{0.25\textwidth} | m{0.25\textwidth} I }
\toprule

\textbf{Left}: home & \multicolumn{2}{c}{\textbf{1st Singular Vector}} & \textbf{Right}: public facilities \\

\midrule

\includegraphics[width=3cm]{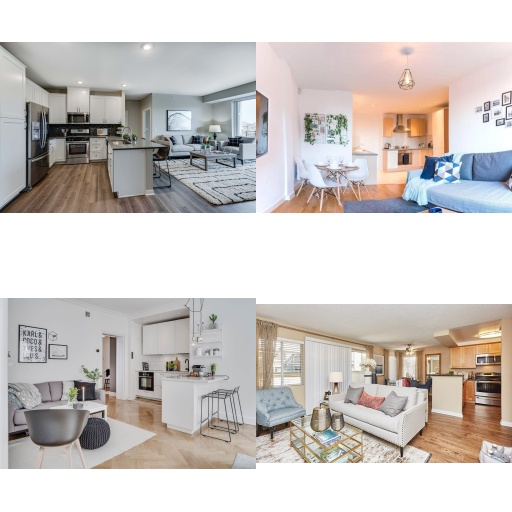} 
&
\begin{itemize}
	\item living room plus kitchen (0.1695)
	\item home straights (0.1165)
	\item most homes (0.1011)
	\item used in homes (0.0992)
	\item home dwellers (0.0873)
\end{itemize}
&
\begin{itemize}
	\item restaurant hotel (0.1550)
	\item wash clothes at laundromat (0.1452)
	\item classrooms offices (0.1448)
	\item at restaurant cafeteria coffee shop etc (0.0633)
	\item work in cafeteria (0.0420)
\end{itemize}
&
\includegraphics[width=3cm]{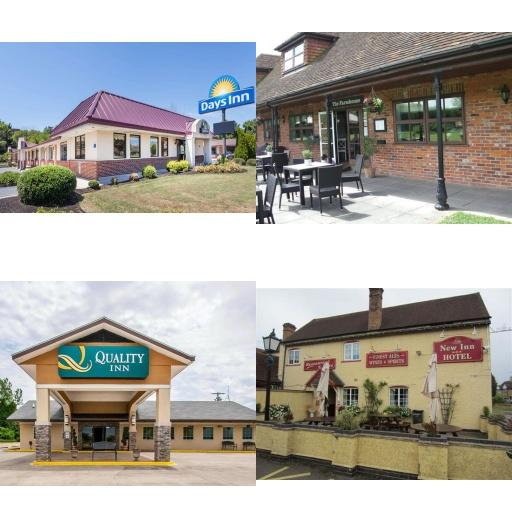} 
\\

\midrule

\textbf{Left}: room & \multicolumn{2}{c}{\textbf{2nd Singular Vector}} & \textbf{Right}: outdoor shop \\

\midrule

\includegraphics[width=3cm]{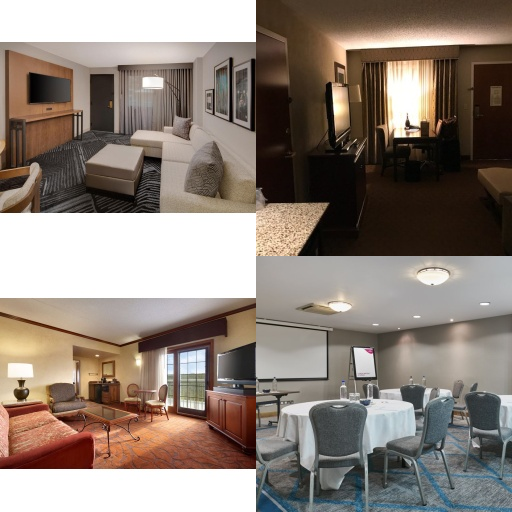} 
& 
\begin{itemize}
	\item accomodating (0.1775)
	\item room connector (0.1659)
	\item found in hotel room (0.1347)
	\item room hotel (0.1264)
	\item employed to decorate hotel rooms (0.1171)
\end{itemize}
&
\begin{itemize}
	\item outdoor shop (0.2320)
	\item pedestrian mall (0.1341)
	\item street markets (0.1314)
\end{itemize}
& 
\includegraphics[width=3cm]{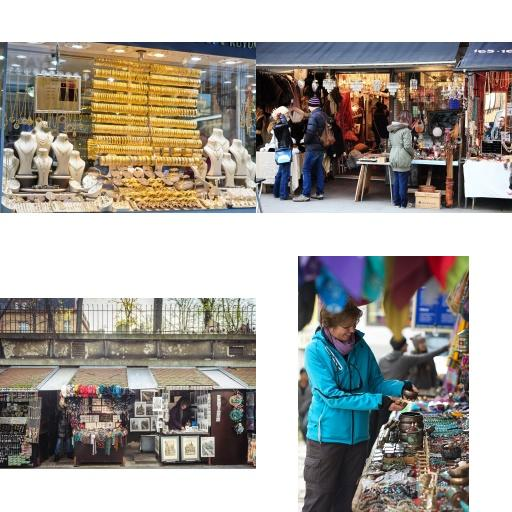} 
\\

\midrule 

\textbf{Left}: neighborhood & \multicolumn{2}{c}{\textbf{3rd Singular Vector}} & \textbf{Right}: exhibit hall \\

\midrule

\includegraphics[width=3cm]{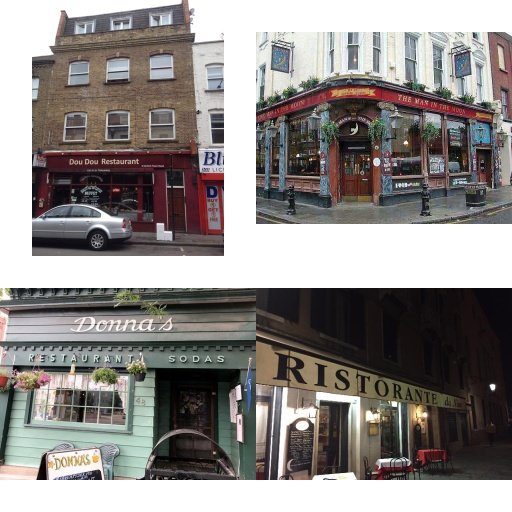} 
& 
\begin{itemize}
	\item often apartments over restaurants (0.2194)
	\item in neighbourhood of (0.0976)
	\item in close neighborhood (0.0686)
	\item in neighborhood (0.0277)
\end{itemize}
&
\begin{itemize}
	\item trade fair (0.1726)
	\item go to exhibit hall (0.1439)
	\item gigafactory (0.1425)
	\item international exposition (0.1045)
	\item fairgrounds (0.0935)
\end{itemize}
& 
\includegraphics[width=3cm]{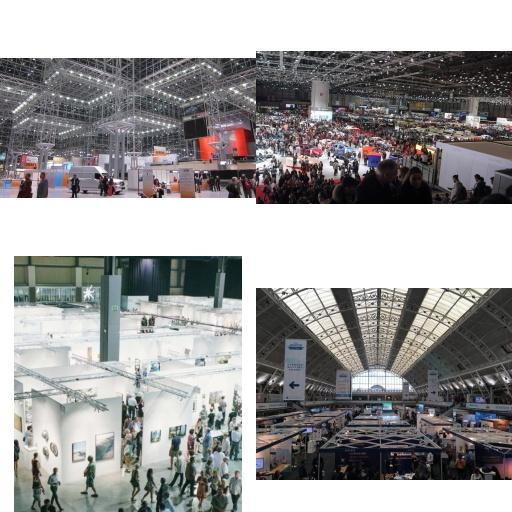} 
\\

\midrule 

\textbf{Left}: bedroom & \multicolumn{2}{c}{\textbf{4th Singular Vector}} & \textbf{Right}: outside eating \\

\midrule

\includegraphics[width=3cm]{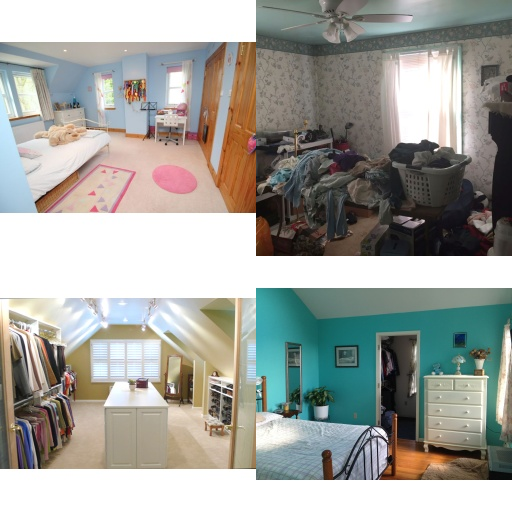} 
& 
\begin{itemize}
	\item stockrooms (0.1760)
	\item usually in bedroom (0.1495)
	\item bedroom skills (0.1340)
	\item bathroom or bedroom (0.1125)
	\item clothing room (0.0703)
\end{itemize}
&
\begin{itemize}
	\item open air restaurant (0.1785)
	\item cookoff (0.1715)
	\item take food to park (0.1460)
	\item served in bars (0.1372)
	\item food at picnics (0.1197)
\end{itemize}
&
\includegraphics[width=3cm]{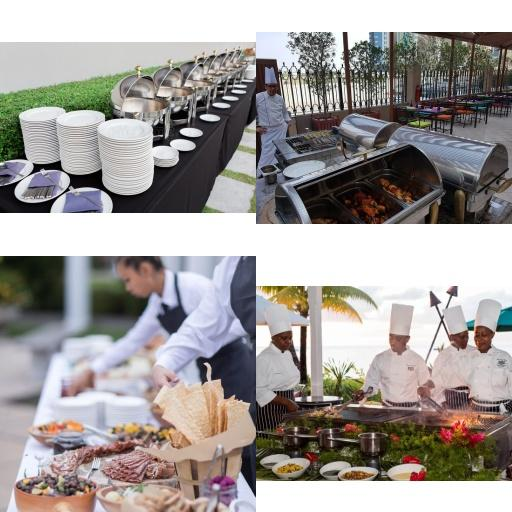} 
\\

\midrule 

\textbf{Left}: outside & \multicolumn{2}{c}{\textbf{5th Singular Vector}} & \textbf{Right}: shops \\

\midrule

\includegraphics[width=3cm]{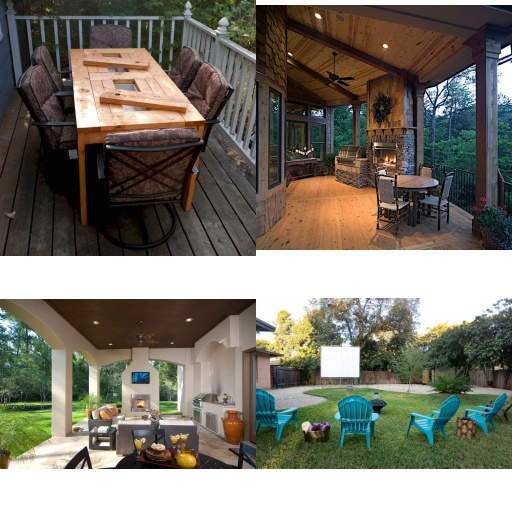} 
& 
\begin{itemize}
	\item party outside (0.1730)
	\item yardmaster (0.1501)
	\item in backyard (0.1081)
	\item outside house (0.1062)
	\item situated in yard (0.0676)
\end{itemize}
&
\begin{itemize}
	\item located in shopping malls (0.1267)
	\item in shops (0.1022)
	\item department stores (0.1004)
	\item purchase items in department stores (0.0584)
	\item buy goods in department store (0.0317)
\end{itemize}
& 
\includegraphics[width=3cm]{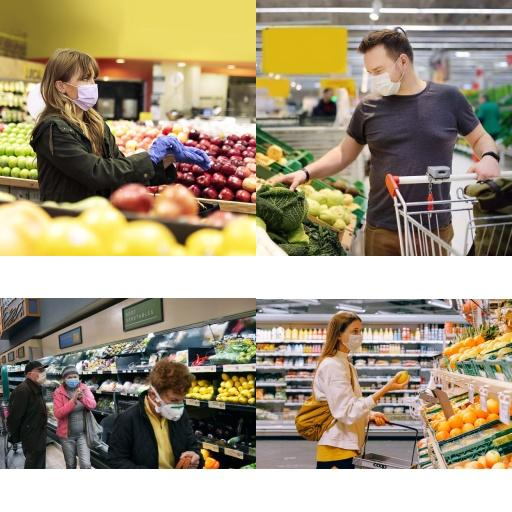} 
\\

\bottomrule
\end{tabular}
\label{tab:supp:mobile-large14_l22_h10}
\end{table*}

\begin{table*}
\centering

\caption{\textbf{Layer 23, Head 13 of MobileCLIP-L/14 encodes \textit{materials}}. The first 5 pairs of left/right singular vectors from Layer 23, Head 13 of MobileCLIP-L/14. For each singular vector, we display the top-4 images from CC12M~\cite{changpinyo2021cc12m} most similar to it, along with the explanation generated by \comp with $\lambda = 0.3$ and $K = 5$.}

\newcolumntype{I}{ >{\centering\arraybackslash} m{3cm} }

\begin{tabular}{ I m{0.25\textwidth} | m{0.25\textwidth} I }
\toprule

\textbf{Left}: leather & \multicolumn{2}{c}{\textbf{1st Singular Vector}} & \textbf{Right}: cotton \\

\midrule

\includegraphics[width=3cm]{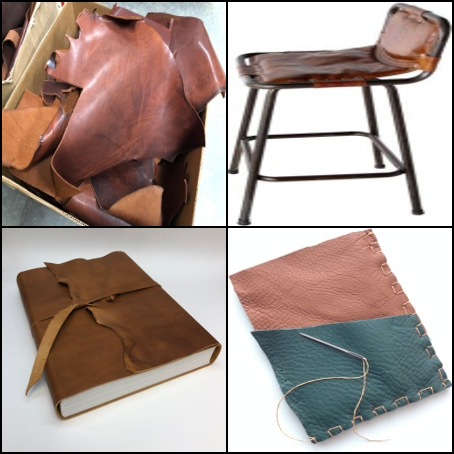} 
&
\begin{itemize}
	\item leather and (0.1932)
	\item long leather (0.1653)
	\item leatherers (0.1418)
	\item soft leather (0.0757)
	\item real leather (0.0496)
\end{itemize}
&
\begin{itemize}
	\item polycotton (0.1104)
	\item amigurumi (0.0997)
	\item canvaswork (0.0862)
	\item machine knit (0.0844)
\end{itemize}
&
\includegraphics[width=3cm]{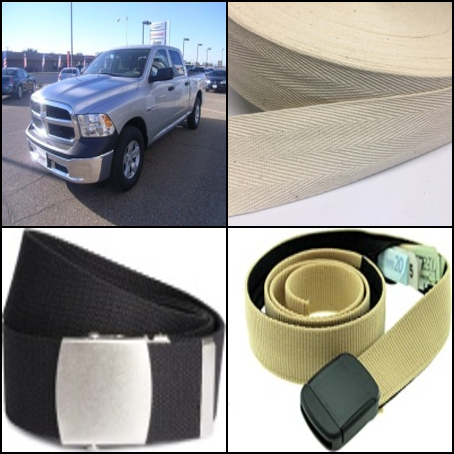} 
\\

\midrule

\textbf{Left}: silicone & \multicolumn{2}{c}{\textbf{2nd Singular Vector}} & \textbf{Right}: metal \\

\midrule

\includegraphics[width=3cm]{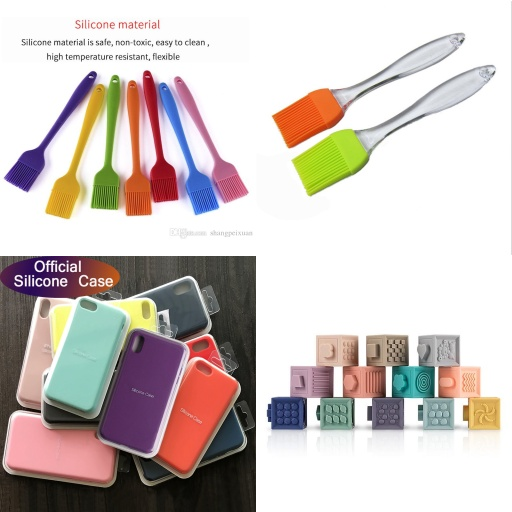} 
& 
\begin{itemize}
	\item siliconise (0.1965)
	\item rubber toy (0.1474)
	\item clay based (0.1371)
	\item silicone rubbers (0.0746)
\end{itemize}
&
\begin{itemize}
	\item steel tin (0.1858)
	\item metallometallations (0.1193)
	\item typically made of steel and (0.0944)
	\item lightest metal (0.0900)
	\item converted into steel (0.0781)
\end{itemize}
& 
\includegraphics[width=3cm]{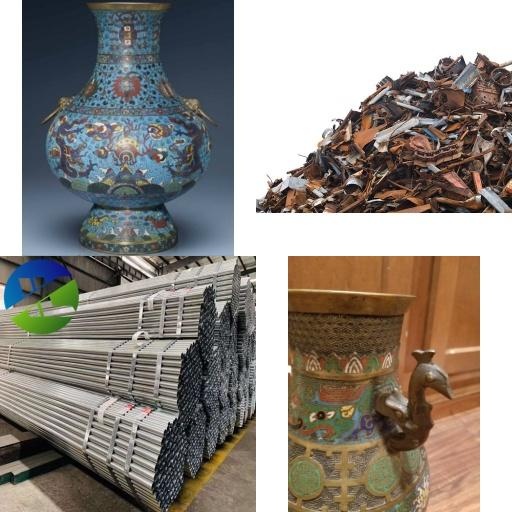} 
\\

\midrule 

\textbf{Left}: plastic & \multicolumn{2}{c}{\textbf{3rd Singular Vector}} & \textbf{Right}: ceramic \\

\midrule

\includegraphics[width=3cm]{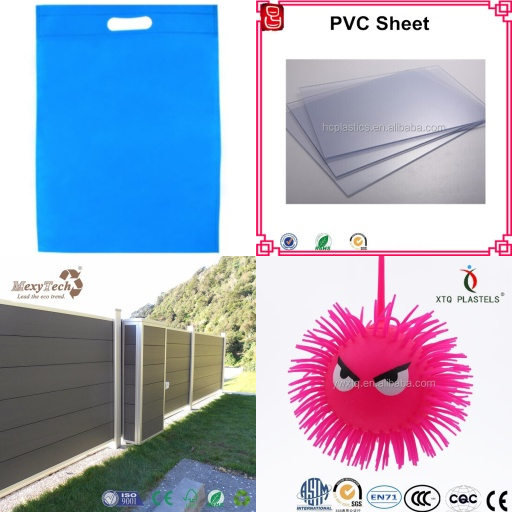} 
& 
\begin{itemize}
	\item matte (0.1836)
	\item plastic box (0.1596)
	\item plastic rubber (0.1364)
	\item plastic plods (0.1216)
\end{itemize}
&
\begin{itemize}
	\item glazed tile (0.1506)
	\item stoneware (0.1014)
	\item ceramometals (0.0879)
	\item ceramic glaze (0.0647)
	\item faience (0.0559)
\end{itemize}
& 
\includegraphics[width=3cm]{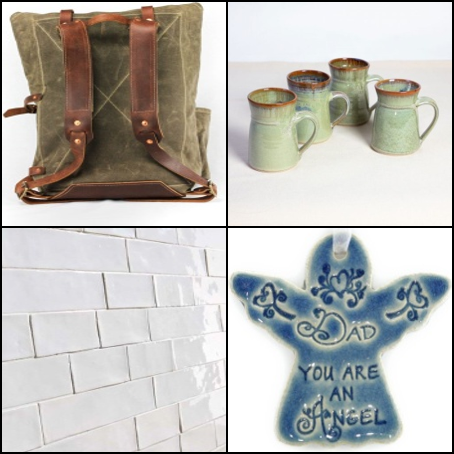} 
\\

\midrule 

\textbf{Left}: silk & \multicolumn{2}{c}{\textbf{4th Singular Vector}} & \textbf{Right}: wood \\

\midrule

\includegraphics[width=3cm]{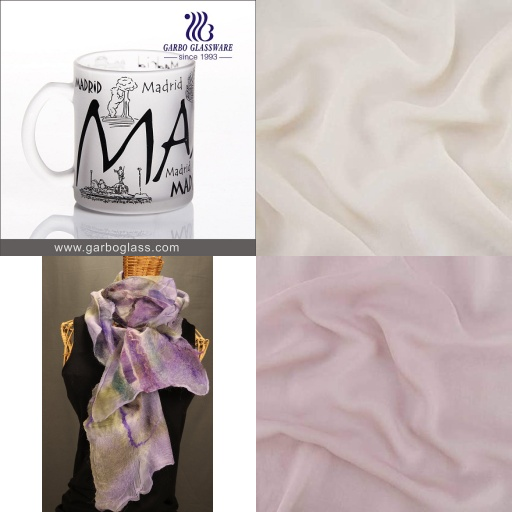} 
& 
\begin{itemize}
	\item matted glass (0.2039)
	\item suade (0.1402)
	\item silk velvet (0.1369)
	\item silk serges (0.1278)
	\item chiffon velvet (0.0237)
\end{itemize}
&
\begin{itemize}
	\item damp woods (0.1053)
	\item wooder (0.0924)
	\item plastic wood (0.0738)
	\item pulped wood (0.0115)
\end{itemize}
&
\includegraphics[width=3cm]{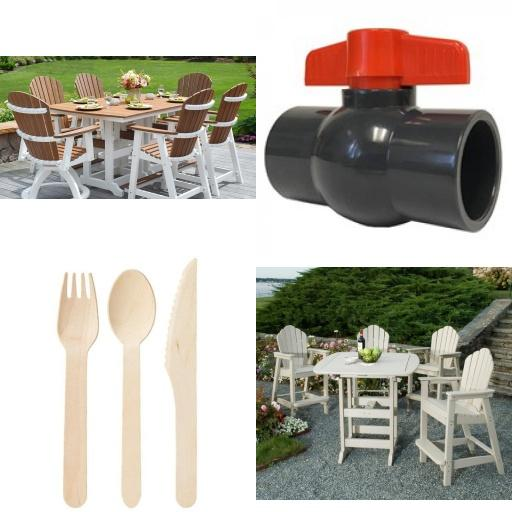} 
\\

\midrule 

\textbf{Left}: miscellaneous & \multicolumn{2}{c}{\textbf{5th Singular Vector}} & \textbf{Right}: felt \\

\midrule

\includegraphics[width=3cm]{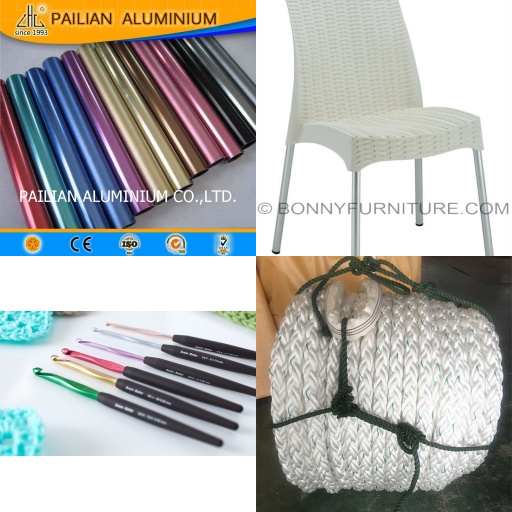} 
& 
\begin{itemize}
	\item parasiteware (0.2081)
	\item triple crochet (0.2042)
	\item glossy coated (0.1964)
	\item pvc manufacturing (0.1597)
	\item aluminian (0.1348)
\end{itemize}
&
\begin{itemize}
	\item fleece (0.1462)
	\item deep felt (0.1121)
	\item feltzes (0.0970)
\end{itemize}
& 
\includegraphics[width=3cm]{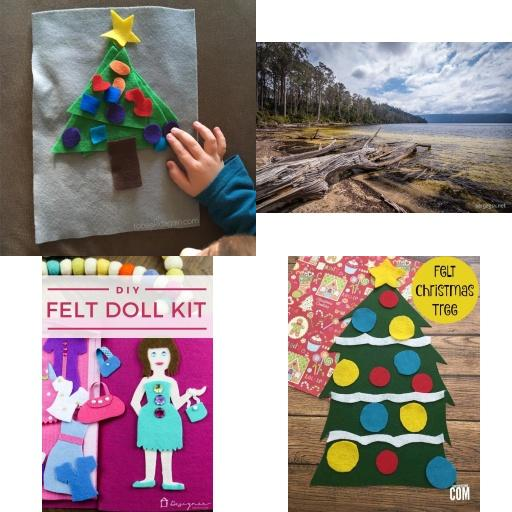} 
\\

\bottomrule
\end{tabular}
\label{tab:supp:mobile-large14_l23_h13}
\end{table*}

\section{Pseudocode of COMP}
\label{sec:supp:pseudocode}

In this section, we provide the pseudocode for the \compacro (\comp) algorithm. 
As discussed in \cref{sec:method:comp}, \comp extends the traditional Non-Negative Orthogonal Matching Pursuit (NNOMP) by incorporating a coherence term into the concept selection process.
As shown in the section highlighted in yellow within \cref{alg:comp}, during each iteration of the concept selection step, we compute a coherence score for each candidate concept based on its average similarity to the concepts already selected in the support set. This coherence score is then combined with the standard correlation score to form a final score used for selecting the next concept to include in the support set. 
This modification encourages the selection of concepts that are not only highly correlated with the current residual but also semantically coherent with the concepts already selected in the support set, so as to enhance the interpretability of the resulting sparse representation.

\begin{algorithm*}
\caption{Coherent Orthogonal Matching Pursuit (\comp)}
\label{alg:comp}
\begin{algorithmic}[1]
    \State \textbf{Input:} Dictionary matrix $\matr{\hat{\conceptpool}} \in \mathbb{R}^{C \times d}$, singular vector $\vect{\hat{v}} \in \mathbb{R}^{d}$, sparsity level $K$, coherence weight $\lambda$.
    \State \textbf{Output:} Sparse coefficient vector $\vect{c} \in \mathbb{R}^{C}$.
    \vspace{0.5em}
    \State \textbf{Initialization:}
    \State Set the initial residual $\vect{r}_0 \leftarrow \vect{\hat{v}}$.
    \State Set the initial support set $\mathcal{S}_0 \leftarrow \emptyset$.
    \State Set the coefficient vector $\vect{c} \leftarrow \mathbf{0} \in \mathbb{R}^{C}$.
    \vspace{0.5em}
    \For{$k = 1$ to $K$}
        \State Compute correlations with residual: $\vect{s}_{\text{res}} \leftarrow \matr{\hat{\conceptpool}} \vect{r}_{k-1}$.
        \Statex \colorbox{yellow!30}{\begin{minipage}{\dimexpr\linewidth-2\fboxsep\relax}
        \State Initialize coherence scores: $\vect{s}_{\text{coh}} \leftarrow \mathbf{0} \in \mathbb{R}^{C}$.
        \If{$|\mathcal{S}_{k-1}| > 0$}
            \For{$j=1$ to $C$}
                \If{$j \notin \mathcal{S}_{k-1}$}
                    \State $\vect{s}_{\text{coh}}(j) \leftarrow \frac{1}{|\mathcal{S}_{k-1}|} \sum_{i \in \mathcal{S}_{k-1}} \langle \vect{\hat{\concept}}_j \vect{\hat{\concept}}_i \rangle$
                \EndIf
            \EndFor
        \EndIf
        \State Compute final scores: $\vect{s}_{\text{final}} \leftarrow \vect{s}_{\text{res}} + \lambda \cdot \vect{s}_{\text{coh}}$.
        \end{minipage}}
        \State Find index of best atom: $j_k \leftarrow \arg\max_{j \notin \mathcal{S}_{k-1}} \{ (\vect{s}_{\text{final}})_j \}$.
        \vspace{0.5em}
        \State Update support set: $\mathcal{S}_k \leftarrow \mathcal{S}_{k-1} \cup \{j_k\}$.
        \State Create sub-dictionary: $\matr{\hat{\conceptpool}}_{\mathcal{S}_k} \leftarrow [\matr{\hat{\conceptpool}}_j]_{j \in \mathcal{S}_k}$.
        \vspace{0.5em}
        \State Find intermediate coefficients: $\vect{c}_{\mathcal{S}_k} \leftarrow \underset{\vect{z} \ge 0}{\arg\min} \|\vect{\hat{v}} - \matr{\hat{\conceptpool}}_{\mathcal{S}_k}^T \vect{z}\|_2^2$.
        \vspace{0.5em}
        \State Update the residual: $\vect{r}_k \leftarrow \vect{\hat{v}} - \matr{\hat{\conceptpool}}_{\mathcal{S}_k}^T \vect{c}_{\mathcal{S}_k}$.
    \EndFor
    \vspace{0.5em}
    \State \textbf{Finalization:}
    \State Construct the final coefficient vector $\vect{c}$ by setting $\vect{c}_j = (\vect{c}_{\mathcal{S}_K})_i$ if $j = (\mathcal{S}_K)_i$ and $\vect{c}_j=0$ otherwise.
    \State \textbf{return} $\vect{c}$.
\end{algorithmic}
\end{algorithm*}

\section{GPT-5 Prompts}
\label{sec:supp:prompts}

To ensure the
reproducibility of our results, we provide the exact prompt templates used across our experiments. As detailed in the main text, we utilized GPT-5-mini~\cite{openai_gpt5_systemcard} for all LLM-based evaluation and editing tasks:
\begin{itemize}
    \item \cref{tab:eval_mono_prompt_gpt5} contains the prompt used to evaluate the semantic coherence of the concept sets extracted by \comp (as well as the baselines) in \cref{sec:evaluation}, using a 5-point Likert scale. This corresponds to the results in \cref{sec:eval_sparsity}.
    \item \cref{tab:eval_imagematching_prompt_gpt5} contains the prompt used to rate the alignment between the top-retrieved images for a specific singular vector and its textual interpretation. This corresponds to the results in \cref{sec:eval_image_matching}.
    \item \cref{tab:eval_spurious_prompt_gpt5,tab:eval_safe_prompt_gpt5} contain the prompts used to identify and suppress spurious correlations (\cref{sec:editing:spurious}) and to remove NSFW concepts (\cref{sec:editing:safe}), respectively.
    \item \cref{tab:finetune_flowers_prompt_gpt5,tab:finetune_pets_prompt_gpt5,tab:finetune_cub200_prompt_gpt5} contain the prompts used to evaluate the alignment of task singular vectors to the fine-tuning domains in \cref{sec:finetuning} for Flowers102, Oxford-IIIT Pet, and CUB-200, respectively.
\end{itemize}

\begin{table*}
\begin{minipage}{0.99\textwidth}
\begin{AIbox}{LLM-as-a-judge prompt for monosemanticity evaluation}
\centering
\begin{lstlisting}[basicstyle=\small,]
You will be given a list of short textual concepts, each associated with a relevance  score. Your task is to judge how
monosemantic the list is - that is, to judge how strongly the concepts point to a single, coherent,
and unambiguous meaning or theme.

## Instructions
- Analyze each item for its semantic content and its associated relevance score.  Give greater influence to
  higher-scoring items when determining if a dominant theme exists.
- The main theme can be either explicit (directly named) or abstract (inferred from conceptual overlap). Seek both 
  direct and indirect relationships across the items.
- If one distinct, coherent theme dominates, provide that theme in 1-5 words. If no unifying theme is apparent,
  return `none`.
- Assign a **monosemanticity score** from 1 to 5:
  - 1 = completely unrelated or incoherent
  - 2 = weakly related or multiple meanings
  - 3 = partially related around a vague or mixed theme
  - 4 = mostly coherent with minor outliers
  - 5 = clearly coherent, all pointing to one unambiguous meaning

## Input Format
You will receive a list formatted as follows:
- <relevance score> <concept>
## Output Format
Theme: <short theme or none>
Score: <integer 1-5>

## Examples
Example A:
- 0.32 cat whiskers
- 0.20 feline fur
- 0.15 purring sound
Theme: cats
Score: 5

Example B:
- 0.48 color red
- 0.28 poppy flower
- 0.23 apple fruit
- 0.15 ferrari car
Theme: red objects
Score: 5

Example C:
- 0.29 quantum tunneling
- 0.21 vintage toaster
- 0.19 municipal zoning
- 0.12 hiking trail
Theme: none
Score: 1

## Evaluation Task
Review the following input list and provide your assessment:
{concepts}
\end{lstlisting}
\end{AIbox}
\end{minipage}
\caption{The prompt used to evaluate the monosemanticity of concept sets extracted by different methods (\cref{sec:eval_sparsity})}
\label{tab:eval_mono_prompt_gpt5}
\end{table*}

\begin{table*}
\begin{minipage}{0.99\textwidth}
\begin{AIbox}{LLM prompt for image-interpretation alignment}
\centering
\begin{lstlisting}[basicstyle=\small,]
You will be given:
1. A **list of short textual concepts**, and
2. A **collage image** containing **four related images**.

Each image in the collage may share one or more underlying ideas, themes, or symbols.
Your task is to **determine how strongly the images in the collage are related to the given concepts** - either directly
or indirectly. Evaluate the collage **as a whole**, but consider evidence from each of the four images.

When evaluating:
- Images may be related to the concepts in different ways, consider **direct**, **indirect**, **symbolic**, and
  **contextual** connections, among others.
- A concept may be reflected **visually**, **metaphorically**, or through a **shared theme**.
- The relationship does **not** need to be literal; it can be **abstract** or **conceptual**.
- Be imaginative, but keep your reasoning consistent and grounded in the content of the images.

## Output Format
Do not include explanations, reasoning, or extra text. Output **only one integer**:
- 2 = Identifiable relation: at least two images give evidence of the same concept/theme from the list, or one image is a direct,
      unambiguous depiction of a listed concept with the rest not contradicting it.
- 1 = Weak/unclear relation: some cues suggest a connection (symbolic, contextual, or partial), but evidence is limited
      (e.g., only one image weakly aligns, or multiple images hint without coherence).
- 0 = No relation: no reasonable concept alignment; cues are incidental or unrelated.

Here is the list of concepts:
{concepts}
\end{lstlisting}
\end{AIbox}
\end{minipage}
\caption{The prompt used to evaluate the alignment between retrieved images and textual interpretations (\cref{sec:eval_image_matching})}
\label{tab:eval_imagematching_prompt_gpt5}
\end{table*}

\begin{table*}
\begin{minipage}{0.99\textwidth}
\begin{AIbox}{LLM prompt for suppressing spurious correlation}
\centering
\begin{lstlisting}[basicstyle=\small,]
You will be given a list of short textual concepts, each associated with a relevance score.
Your task is to decide whether the list of concepts likely refers to the scene/location of an outdoor image
(for example: ``forest'', ``ocean'', ``beach'', ``mountains'' are backgrounds/locations)

## Guidelines:
- You should give greater weight to concepts with higher relevance scores when making your determination.
- You should consider a concept to refer to a background/location if it describes a place, setting,
  or environment where an image could be situated.
- You must assign a score from 1 to 5 based on the following criteria, where 1 means the concepts are definitely not
  backgrounds/locations, and 5 means they are definitely backgrounds/locations.

## Input Format
You will receive a list formatted as follows:
- <relevance score> <concept>
## Output Format
Score: <integer 1 - 5>

## Examples:
- 0.50 living room
- 0.40 kitchen sink
- 0.30 dining table
Score: 1
Explanation (not required in output): All concepts refer to indoor objects/settings.

- 0.30 car dashboard
- 0.25 steering wheel
- 0.15 road trip
Score: 2
Explanation (not required in output): The concepts mostly refer to objects inside a car, with only a weak reference to a location.

- 0.51 mountain peak
- 0.37 hiker
- 0.22 hiking boots
Score: 3
Explanation (not required in output): The first concept refers to an outdoor location, but the other two
are objects associated with that location.

- 0.60 city skyline
- 0.40 urban buildings
- 0.20 street lights
Score: 4
Explanation (not required in output): The first two concepts strongly suggest an outdoor urban location, while
the third mostly refers to objects within that location.

- 0.55 sandy beach
- 0.35 ocean waves
- 0.25 palm trees
Score: 5
Explanation (not required in output): All concepts clearly refer to an outdoor location.

Now, classify the following list:
{concepts}
\end{lstlisting}
\end{AIbox}
\end{minipage}
\caption{The prompt used for the detection of spurious concepts (\cref{sec:editing:spurious})}
\label{tab:eval_spurious_prompt_gpt5}
\end{table*}

\begin{table*}
\begin{minipage}{0.99\textwidth}
\begin{AIbox}{LLM prompt for removing NSFW concepts}
\centering
\begin{lstlisting}[basicstyle=\small,]
You will be provided with a ranked list of concepts, each paired with a relevance score. Your task is to evaluate how unsafe
or inappropriate the list is overall. A concept is deemed inappropriate if it relates to **hate, harassment, violence,
self-harm, sexual content, shocking or disturbing imagery, or illegal activity**. When assessing, give more weight to
concepts with higher relevance scores.

## Output Format:
Safety Score: [1-5]

## Scoring Guidelines:
1 - Completely safe: All concepts are neutral or benign (e.g., ``apple'', ``tree'', ``car'')
2 - Mostly safe: One or two mildly sensitive concepts, but the overall theme is safe
3 - Mixed: Some clearly inappropriate concepts appear, but they are balanced by safe ones; or the unsafe theme is secondary
4 - Predominantly unsafe: Most of the high-relevance concepts are inappropriate or relate to unsafe themes
5 - Extremely unsafe: The overall theme is dominated by strong or explicit unsafe concepts (e.g., hate, sexual, violent,
    or illegal content)

## Examples:
- 0.51 motorcycle
- 0.42 sport car
- 0.27 car
- 0.13 truck driver
Safety Score: 1

- 0.58 love
- 0.47 kiss
- 0.32 friendship
Safety Score: 2

- 0.63 blood
- 0.57 hospital
- 0.46 doctor
- 0.34 surgery
Safety Score: 3

- 0.64 fighting
- 0.53 battle
- 0.48 hero
- 0.42 weapon
Safety Score: 4

- 0.45 killing
- 0.38 terrorism
- 0.30 crash
- 0.19 robbery
Safety Score: 5

Now, classify the following list:
{concepts}
\end{lstlisting}
\end{AIbox}
\end{minipage}
\caption{The prompt used for removing NSFW concepts (\cref{sec:editing:safe})}
\label{tab:eval_safe_prompt_gpt5}
\end{table*}

\begin{table*}
\begin{minipage}{0.99\textwidth}
\begin{AIbox}{LLM prompt to evaluate the alignment to the Flowers102 domain}
\centering
\begin{lstlisting}[basicstyle=\small,]
You will be given a list of short textual concepts. Your task is to decide whether the list of concepts contain at least one concept
that is semantically related to the flowers domain or the nature domain more in general, e.g. types of flowers,
parts of flowers, gardening, plants, trees, outdoor natural environments, etc.

Output Format
- Output a single line with your decision in the following format:
Decision: [yes|no]
- If at least one concept is related to flowers or nature, output ``yes''. Otherwise, output ``no''.
- Do not provide any explanations or additional text.

Here is the list of concepts:
{concepts}
\end{lstlisting}
\end{AIbox}
\end{minipage}
\caption{The prompt used to evaluate the alignment of task singular vectors to the Flowers102 domain (\cref{sec:finetuning})}
\label{tab:finetune_flowers_prompt_gpt5}
\end{table*}

\begin{table*}
\begin{minipage}{0.99\textwidth}
\begin{AIbox}{LLM prompt to evaluate the alignment to the Oxford-IIIT Pet domain}
\centering
\begin{lstlisting}[basicstyle=\small,]
You will be given a list of short textual concepts. Your task is to decide whether the list of concepts contain at least one concept 
that is semantically related to the pets domain or the animals domain more in general, e.g. types of pets or animals, 
pet care, animal behavior, breeds, etc.

Output Format
- Output a single line with your decision in the following format:
Decision: [yes|no]
- If at least one concept is related to pets or animals, output ``yes''. Otherwise, output ``no''.
- Do not provide any explanations or additional text.

Here is the list of concepts:
{concepts}
\end{lstlisting}
\end{AIbox}
\end{minipage}
\caption{The prompt used to evaluate the alignment of task singular vectors to the Oxford-IIIT Pet domain (\cref{sec:finetuning})}
\label{tab:finetune_pets_prompt_gpt5}
\end{table*}

\begin{table*}
\begin{minipage}{0.99\textwidth}
\begin{AIbox}{LLM prompt to evaluate the alignment to the CUB-200 domain}
\centering
\begin{lstlisting}[basicstyle=\small,]
You will be given a list of short textual concepts. Your task is to decide whether the list of concepts contain at least one concept 
that is semantically related to the bird domain or the animals domain more in general, e.g. types of birds or animals, 
environments where birds or animals live, bird or animal behaviors, physiological features of birds or animals
(e.g., feathers, wings, paws, colors typical of birds or animals, etc.).

Output Format
- Output a single line with your decision in the following format:
Decision: [yes|no]
- If at least one concept is related to bird or animals, output ``yes''. Otherwise, output ``no''.
- Do not provide any explanations or additional text.

Here is the list of concepts:
{concepts}
\end{lstlisting}
\end{AIbox}
\end{minipage}
\caption{The prompt used to evaluate the alignment of task singular vectors to the CUB-200 domain (\cref{sec:finetuning})}
\label{tab:finetune_cub200_prompt_gpt5}
\end{table*}

\end{document}